# Zero-Shot Object Removal via Attention Masking, Latent Anchoring, and Refinement

Arman Taghizadeh* Ulf Krumnack Kai-Uwe Kühnberger

Institute of Cognitive Science, Osnabrück University, Osnabrück, Germany

*staghizadehm@uni-osnabrueck.de

**Abstract**

Removing an object from a real image requires more than synthesizing plausible content within a mask: the method must suppress residual object features, preserve the unedited scene, and generate replacement content that is consistent with the surrounding background. This paper approaches object removal from a stage-based perspective and proposes a zero-shot framework for constrained latent inpainting with a frozen pretrained Stable Diffusion model, requiring no task-specific training or model fine-tuning. The method integrates SAM-based mask construction, BLIP image-caption conditioning, DDIM inversion, background-weighted masked null-text optimization, decoder self-attention masking, hard outside-mask latent anchoring, and localized renoise–denoise refinement into a unified pipeline. The method is evaluated through qualitative examples, quantitative local-consistency metrics, and ablation studies. The results demonstrate effective object removal and context-consistent replacement content. The ablations indicate that background-weighted masked NTI is particularly beneficial for structurally complex backgrounds, whereas the no-NTI variant is sufficient in other evaluated examples. Repeated refinement further reduces object remnants and boundary artifacts remaining after the primary editing pass.

## 1. Introduction

Pretrained latent diffusion models such as **Stable Diffusion** (Rombach et al. 2022) have made high-quality text-guided image generation widely accessible. A natural next step is to use these models for *image editing* without task-specific fine-tuning. The central premise is that modern diffusion models provide strong internal "control handles" through their architecture and denoising dynamics. In particular, their generative priors, latent trajectories, text embeddings, and internal attention representations can be manipulated to support controlled editing at inference time.

Object removal is a specialized form of image inpainting in which a user-specified object must be eliminated and the resulting region must be filled with realistic and context-consistent background content. The method must suppress residual object features, preserve the unedited scene, recover content that may have been occluded by the object, and maintain consistency across the mask boundary. These requirements make object removal a useful setting for investigating how the internal representations and denoising dynamics of pretrained diffusion models can be controlled during inference. From the extensive body of literature, we prioritize zero-shot methods (i.e., requiring no task-specific model fine-tuning) and examine their mechanisms and design choices in relation to recurring challenges associated with this approach in diffusion-based image editing, including inversion and trajectory recovery, background anchoring and context preservation, mask-based localized edit control, attention-driven editing control, and boundary/seam refinement. The review focuses primarily on approaches involving attention manipulation, embedding optimization, sampling-time control, and the distinct functional roles of early, middle, and late denoising timesteps, as these mechanisms constitute the main technical pillars of the object-removal pipeline developed in this work. The reviewed methods are organized according to their primary technical mechanisms.

A first line of training-free diffusion inpainting and local-editing methods modifies the sampling process to condition generation on the visible image content through known-region replacement, sampling-time constraints, guidance objectives, or latent optimization. **RePaint** uses a pretrained unconditional DDPM and repeatedly combines known regions sampled from the input image with unknown regions generated by the model during reverse diffusion (Lugmayr et al. 2022). Its back-and-forth resampling procedure gives the model repeated opportunities to harmonize the generated region with the preserved context, providing an important basis for boundary and context negotiation without mask-specific training. **Blended Latent Diffusion** transfers this known-region replacement principle to latent diffusion by repeatedly blending generated masked-region latents with source-derived background latents during denoising (Avrahami et al. 2023). **DiffEdit** combines deterministic inversion with automatically derived mask guidance to perform zero-shot semantic editing while limiting changes to the identified edit region (Couairon et al. 2023). **GradPaint** instead applies gradient guidance from a masked reconstruction loss and a boundary-alignment loss to the denoised image estimate at each sampling step (Grechka et al. 2023). By backpropagating these losses through the frozen diffusion model, it directly steers the denoising trajectory toward improved context harmonization and boundary consistency without modifying the model parameters. **CoPaint** formulates inpainting with a fixed diffusion model through a Bayesian framework that jointly modifies the revealed and unrevealed regions rather than relying only on direct known-region replacement (G. Zhang et al. 2023). Its approximation of the intermediate posterior is designed so that the inpainting-constraint error decreases throughout denoising, addressing the incoherence that can arise between generated and preserved regions. **PILOT** performs training-free multimodal image inpainting through latent-space optimization, using semantic-centralization and background-preservation losses to search for latents that satisfy the user-provided condition while maintaining coherence with the visible background (Pan et al. 2024).

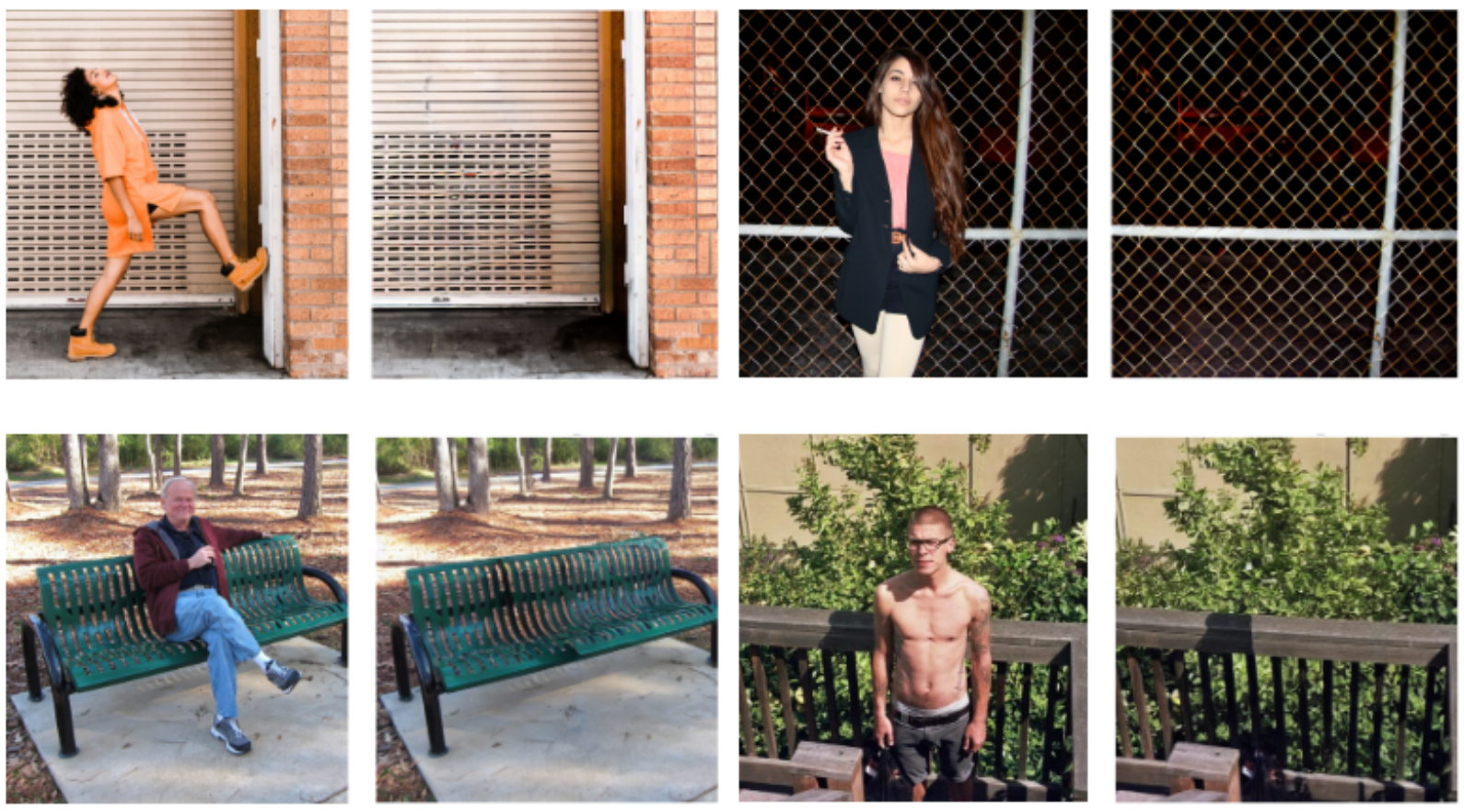

**Figure 1.** Qualitative examples of the proposed object-removal framework.

Its optimization-based formulation demonstrates how the generative prior of a frozen diffusion model can be exploited for an individual inpainting problem without modifying the model parameters. More recently, **LanPaint** formulates inpainting as partial conditional sampling and introduces a training-free, asymptotically exact method for ODE-based diffusion samplers and rectified-flow models (Zheng et al. 2025). Its Bidirectional Guided Score enables mutual adaptation between the inpainted and observed regions, while Fast Langevin Dynamics accelerates convergence through backpropagation-free sampling with only a small number of inner iterations.

A second line of work uses textual conditions, embedding manipulation, and internal attention representations to identify, suppress, regenerate, or spatially localize image content. **MagicRemover** is a tuning-free text-guided object-removal method that uses null-text inversion and interactions between reconstruction and editing branches to guide generation through cross-attention and self-attention (Yang, L. Zhang, et al. 2023). Its attention guidance strategy suppresses the instructed object while preserving the remaining image, and a classifier optimization procedure improves erasure stability within fewer sampling steps. **Impact of Negative Prompts** analyzes when and how negative prompts influence diffusion generation, showing that their effect is delayed and that concept deletion occurs through neutralization in latent space (Ban et al. 2024). By examining cross-attention maps across denoising timestep intervals, the study identifies a critical window after the unwanted concept has begun to form: applying the negative prompt too early may disrupt the image layout, whereas introducing it around the middle of the reverse-diffusion process provides the best balance between object suppression and background preservation. Relatedly, **SuppressEOT** approaches unwanted-content suppression at the conditioning level by manipulating text embeddings through soft-weighted regularization and inference-time embedding optimization (S. Li, Weijer, et al. 2024). Although it primarily addresses content suppression during text-to-image generation rather than masked editing of an arbitrary real image, it demonstrates that undesired concepts can be weakened by modifying the conditioning representation while keeping the diffusion model fixed. **HD-Painter** addresses prompt neglect in text-guided inpainting through Prompt-Aware Introverted Attention, which reweights self-attention according to the textual condition without model fine-tuning (Manukyan et al. 2025). It further introduces Reweighting Attention Score Guidance to steer DDIM sampling toward prompt-aligned latents while reducing out-of-distribution shifts caused by conventional post-hoc gradient guidance. **Uni-paint** combines textual conditioning with background blending and masked cross-attention and self-attention to support unconditional, text-, stroke-, exemplar-, and mixed-guided inpainting, demonstrating how multiple conditioning modalities can be incorporated while restricting generated content to the unknown region (Yang, Chen, et al. 2023). Unlike the zero-shot methods emphasized here, however, it is a few-shot approach because it performs masked fine-tuning on the individual input image. **ZONE** instead performs zero-shot instruction-guided local editing by converting the user instruction into spatial edit regions, extracting the corresponding image layers with an off-the-shelf segmentation model, and applying FFT-based edge smoothing when recombining the edited and preserved regions (S. Li, Zeng, et al. 2024). Although designed for general local editing rather than object removal alone, it illustrates how textual editing instructions can be mapped to spatial edit regions and combined with outside-region preservation and explicit boundary processing without task-specific training.

Related training-free guidance methods demonstrate that self-attention can also be used to construct guidance directions during sampling. **Self-Attention Guidance (SAG)** uses intermediate self-attention maps to identify salient regions, selectively degrades these regions through blurring, and guides denoising away from the degraded prediction to improve sample quality and reduce artifacts (Hong et al. 2023). **Perturbed-Attention Guidance (PAG)** instead replaces selected self-attention maps with identity matrices to produce structurally degraded predictions and guides sampling away from them (Ahn et al. 2024). These methods are not designed specifically for object removal, but they establish the broader principle of deriving training-free sampling guidance from differences between original predictions and predictions obtained through attention-informed perturbations. More recent spatial-editing and object-removal methods focus directly on controlling self-attention so that the masked region receives background information rather than features belonging to the target object or misleading neighboring instances. **DesignEdit** formulates training-free spatial editing through multi-layered latent decomposition and fusion, using a key-masking self-attention scheme that prevents queries from retrieving masked-region features and propagates surrounding context into the removed region (Jia et al. 2025). Its artifact-suppression strategy further improves the quality of the reconstructed background. **Attentive Eraser** targets object removal by redirecting self-attention from foreground to background information through Attention Activation and Suppression (Sun et al. 2025). Its Self-Attention Redirection Guidance uses the difference between the original and redirected attention predictions to steer sampling toward effective removal and plausible background generation. **PANDORA** introduces Pixel-wise Attention Dissolution, which disconnects masked query pixels from their most strongly correlated self-attention keys and thereby reduces the propagation of object information (Vo et al. 2026). Localized Attentional Disentanglement Guidance complements this operation by reshaping the latent denoising trajectory to suppress residual artifacts and support prompt-free, single-pass, and multi-object removal. **AdaEraser** observes that indiscriminate self-attention suppression can remove the target object while also disrupting background generation, since the vacated region still requires the normal generative capability of the pretrained model (Liu 2026). It therefore estimates the remaining presence of the target object from token-wise self-attention-map similarities and adaptively adjusts the suppression strength throughout denoising. **ClickRemoval** reduces the interaction requirements of mask- and text-guided methods by converting user clicks into object and background semantic maps derived from the internal self-attention of Stable Diffusion (L. Zhang et al. 2026). It combines staged self-attention redirection with output-level guidance to localize the target, suppress its features, and reconstruct the background without additional training, manually drawn masks, or text descriptions. More recently, **DORS** addresses object removal in dense scenes, where visually similar neighboring instances can provide misleading contextual information. It combines Instance-Filtered Attention with Context-Guided Routing to suppress information from similar objects while retaining useful background context (Tang et al. 2026).

Taken together, these methods demonstrate that pretrained diffusion models can be adapted to inpainting and object removal through sampling-time conditioning, gradient or posterior guidance, embedding optimization, and internal attention control. However, these approaches emphasize different subsets of the editing process: known-region replacement and resampling improve preservation but may still produce inconsistencies between generated and preserved regions; gradient guidance improves harmonization at additional inference-time cost; text-based suppression provides limited explicit spatial control; and strong self-attention suppression can eliminate the target while weakening the generation of plausible background content. *Object removal therefore remains a multi-stage problem in which localization, real-image reconstruction, background preservation, object-feature suppression, replacement-content generation, and boundary refinement must be handled jointly.*

Based on this stage-based perspective, we propose a zero-shot object-removal framework that performs constrained latent inpainting with a frozen pretrained Stable Diffusion model and requires no task-specific training or model fine-tuning. The method integrates SAM-based mask construction, BLIP image-caption conditioning, DDIM inversion, background-weighted masked null-text optimization, decoder self-attention masking, hard outside-mask latent anchoring, and localized renoise–denoise refinement into a unified pipeline. Each component has a distinct role: the mask localizes synthesis, DDIM inversion and masked NTI establish a context-consistent background anchor, self-attention masking suppresses target-object feature leakage, hard latent anchoring preserves the unedited region, and refinement improves the consistency of the generated replacement content with the surrounding background while reducing boundary artifacts. The proposed framework is evaluated using qualitative examples, quantitative local-consistency metrics, and ablation studies. The results demonstrate effective object removal with context-consistent replacement content, preservation of the surrounding image, and reduced boundary artifacts. Additional qualitative results, together with their quantitative analyses, are provided in **Appendix**, while the ablation analysis clarifies the contribution of each component to the final edited result.

The implementation of the proposed object-removal pipeline is available in the accompanying GitHub repository: `zero-shot-diffusion-object-removal`.

## 2. Preliminaries

### 2.1. Latent Diffusion and DDIM Inversion

Latent diffusion models perform diffusion in the latent space of a pretrained variational autoencoder (Rombach et al. 2022). Given an input image $I$, the encoder and decoder produce

$$z_{\text{src}} = E(I), \qquad I_{\text{rec}} = D(z_{\text{src}}).$$

At timestep $t$, a clean latent $z_{\text{src}}$ can be noised according to

$$z_t = \sqrt{\bar{\alpha}_t}\, z_{\text{src}} + \sqrt{1-\bar{\alpha}_t}\,\epsilon, \qquad \epsilon \sim \mathcal{N}(0, \mathbf{I}),$$

where $\bar{\alpha}_t$ is the cumulative noise-schedule coefficient.

A conditional U-Net $\epsilon_\theta(z_t, t; c)$ predicts the noise under text condition $c$.

DDIM provides a deterministic mapping between diffusion timesteps (Song et al. 2021). In the remainder of this paper,

$$\text{Step}(z_k, t_k, \epsilon_{\text{pred}})$$

denotes one deterministic scheduler update from $z_k$ to the next latent in the denoising trajectory.

DDIM inversion applies this deterministic process in the opposite direction to map the source latent to a noisy latent. For scheduler timesteps $t_0 > t_1 > \cdots > t_{N-1}$, inversion produces the time-aligned trajectory

$$\{z_k^{\text{inv}}\}_{k=0}^{N},$$

where $z_0^{\text{inv}}$ is the inverted noisy latent and $z_N^{\text{inv}} = z_{\text{src}}$. Its intermediate states provide source-image references at the corresponding noise levels.

### 2.2. Classifier-Free Guidance and Null-Text Inversion

A prompt $p$ is encoded as

$$c = \text{TextEnc}(p),$$

while the empty prompt provides the initial unconditional embedding

$$\varnothing_{\text{init}} = \text{TextEnc}(\text{“”}).$$

Classifier-free guidance combines the conditional and unconditional noise predictions:

$$\epsilon_{\text{cfg}}(z_k, t_k; \varnothing, c, w) = \epsilon_\theta(z_k, t_k; \varnothing) + w\,[\epsilon_\theta(z_k, t_k; c) - \epsilon_\theta(z_k, t_k; \varnothing)],$$

where $w$ is the guidance scale (Ho and Salimans 2022). Strong classifier-free guidance may reduce reconstruction accuracy when the guided denoising path is not fully compatible with the inversion trajectory. Null-Text Inversion addresses this mismatch by optimizing a timestep-specific unconditional embedding while keeping the conditional prompt and diffusion model fixed (Mokady et al. 2023):

$$\varnothing_k^\star = \arg\min_{\varnothing_k} \left\| \text{Step}\left(z_k, t_k, \epsilon_{\text{cfg}}(z_k, t_k; \varnothing_k, c, w)\right) - z_{k+1}^{\text{inv}} \right\|_2^2.$$

The resulting embeddings $\{\varnothing_k^\star\}_{k=0}^{N-1}$ improve the compatibility between classifier-free-guided denoising and the real-image inversion trajectory. Standard NTI applies this reconstruction objective to the complete latent.

### 2.3. Self-Attention and Denoising-Time Dynamics

Self-attention allows each spatial U-Net token to aggregate information from other image locations (Vaswani et al. 2017). Query tokens determine where information is requested, key tokens determine which spatial locations act as contextual sources, and value tokens provide the information transferred to the output. Consequently, object information can propagate from object-region keys to the features used to synthesize the masked region. The role of denoising also changes over time. Early high-noise timesteps mainly establish coarse layout and global structure, whereas later timesteps increasingly determine local structure, texture, and boundaries. Similarly, higher-resolution decoder features have a stronger influence on local visual details (Tumanyan et al. 2022). These properties motivate the timestep-dependent decoder self-attention masking and intermediate-timestep refinement later introduced in the proposed method.

## 3. Problem Definition and Challenges

Given a real RGB image $I \in [0,1]^{H\times W\times 3}$ and a binary object mask $M_{\text{img}} \in \{0,1\}^{H\times W}$, the goal is to generate an edited image $\hat{I}$ in which the selected object is removed. The mask follows the convention

$$M_{\text{img}}(p) = \begin{cases} 1, & \text{if pixel } p \text{ belongs to the object region,} \\ 0, & \text{otherwise.} \end{cases}$$

The synthesized content inside the mask should be plausible and consistent with the surrounding scene, while the region outside the mask should remain unchanged.

This task involves several coupled challenges. Object information may persist in latent features and self-attention, causing remnants or ghost structures, while global denoising can unintentionally modify the surrounding image. The occluded background must also be inferred from the visible context and the pretrained generative prior. Real-image inversion introduces a further trade-off: accurate reconstruction preserves the source image, but may also encourage the masked object to reappear. The editing process must therefore preserve the background while leaving sufficient freedom for the object region to be resynthesized. Finally, mask quality and boundary consistency are critical, since incomplete masks may leave fragments, oversized masks increase the region to be regenerated, and local mismatches can produce seams, halos, or discontinuities.

*The problem is therefore formulated as constrained latent inpainting that must jointly balance object suppression, outside-mask preservation, context-consistent synthesis, and boundary refinement.*

## 4. Methodology

### 4.1. Pipeline Overview

The proposed method performs zero-shot object removal as constrained latent inpainting with a frozen pretrained Stable Diffusion model. Given an input image $I$ and object mask $M_{\text{img}}$, the pipeline constructs a time-aligned inversion trajectory of the source image and resynthesizes only the masked region while preserving the remaining scene. The method consists of several components: mask construction, image-aligned text conditioning, DDIM inversion, background-weighted masked null-text optimization, decoder self-attention masking with hard background anchoring, and localized renoise–denoise refinement.

The mask determines where synthesis is allowed, the inversion trajectory provides a reference for the source image, masked NTI improves background-aware reconstruction under classifier-free guidance, self-attention masking suppresses the propagation of object features, hard anchoring preserves the outside region, and refinement improves local consistency between the synthesized content and its surrounding context.

### 4.2. Mask Construction and Text Conditioning

An initial object mask is obtained from user-provided positive clicks using the Segment Anything Model (SAM) (Kirillov et al. 2023). The mask can be corrected through a brush interface and is subsequently postprocessed by retaining its largest connected component, filling internal holes, and applying a small dilation. Dilation slightly enlarges the editable region so that residual foreground edges and mixed foreground–background pixels are not preserved by the later anchoring operation. The final image-space mask is downsampled to the latent resolution:

$$M_{\text{in}} = \text{Resize}(M_{\text{img}}), \qquad M_{\text{out}} = 1 - M_{\text{in}},$$

where $M_{\text{in}}$ denotes the editable object region and $M_{\text{out}}$ denotes the preserved region. The masks are broadcast across latent channels when applied to latent tensors. For decoder self-attention, resolution-specific token masks are constructed as

$$m^{(r)} = \text{vec}\left(\text{Resize}_{r\times r}(M_{\text{in}})\right),$$

where $r$ corresponds to the spatial resolution of the respective attention layer.

BLIP (J. Li et al. 2022) generates an image-aligned caption $p$, which is encoded as $c = \text{TextEnc}(p)$. The same embedding is used during inversion, masked NTI, primary editing, and refinement. The method does not require a target editing prompt, negative prompt, or manually specified object token. Instead, the caption provides a general description of the input image, while the mask and inference-time constraints control the removal operation.

### 4.3. Background-Weighted Masked NTI

The input image is first encoded into the latent space of Stable Diffusion, producing $z_{\text{src}}$. DDIM inversion is then applied to obtain the time-aligned trajectory $\left\{z_k^{\text{inv}}\right\}_{k=0}^{N}$, where $z_0^{\text{inv}}$ is the inverted noisy latent and $z_N^{\text{inv}} = z_{\text{src}}$. The trajectory is used both as the initialization for editing and as a reference for preserving the outside-mask region and providing its contextual information throughout denoising. Standard Null-Text Inversion optimizes timestep-specific unconditional embeddings to reconstruct the complete input image (Mokady et al. 2023). For object removal, however, strongly reconstructing the masked region may encourage the original object to persist. We therefore introduce a background-weighted masked NTI objective that prioritizes reconstruction outside the object mask. This extends the role of NTI from reconstruction only to a background-aware anchor for later generation inside the mask.

Let $z_k^{\text{rec}}$ be the current reconstruction latent and $\varnothing_k$ the unconditional embedding optimized at timestep $t_k$. The classifier-free-guided prediction is

$$\epsilon_{\text{cfg},k} = \epsilon_{\text{cfg}}\left(z_k^{\text{rec}}, t_k; \varnothing_k, c, w_{\text{train}}\right),$$

and one deterministic scheduler step gives

$$\hat{z}_{k+1}^{\text{rec}}(\varnothing_k) = \text{Step}\left(z_k^{\text{rec}}, t_k, \epsilon_{\text{cfg},k}\right).$$

The deviation from the corresponding inversion pivot is

$$\Delta_{k+1}(\varnothing_k) = \hat{z}_{k+1}^{\text{rec}}(\varnothing_k) - z_{k+1}^{\text{inv}}.$$

The optimized unconditional embedding is obtained from

$$\varnothing_k^{\star} = \arg\min_{\varnothing_k}\left[\|\Delta_{k+1}(\varnothing_k)\odot M_{\text{out}}\|_2^2 + \lambda_{\text{in}}\,\|\Delta_{k+1}(\varnothing_k)\odot M_{\text{in}}\|_2^2\right],$$

where $\lambda_{\text{in}} \ll 1$. The first term aligns guided denoising with the source trajectory in the preserved region, while the weak second term prevents unconstrained instability without forcing accurate reconstruction of the target object.

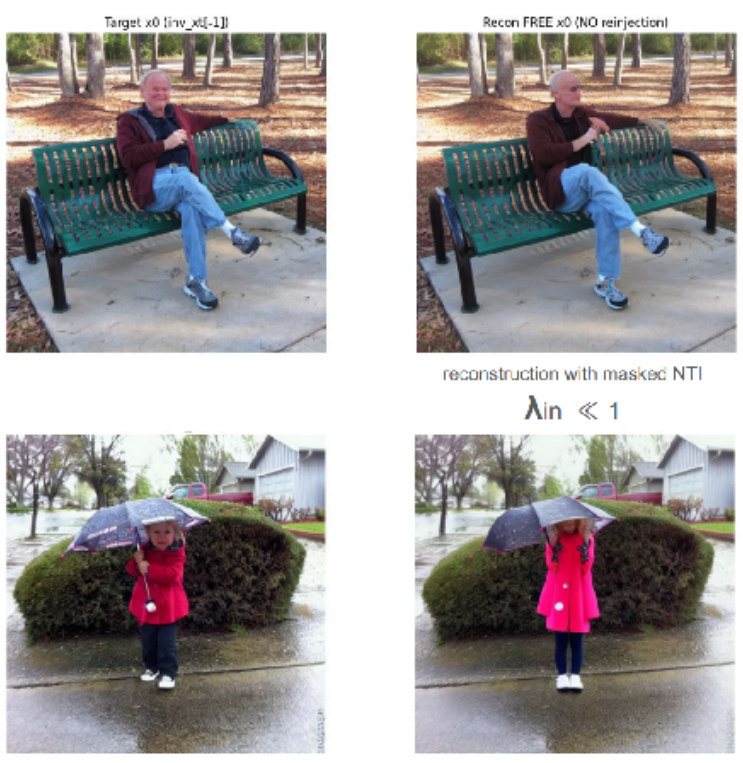

**Figure 2.** Masked NTI background reconstruction ($\lambda_{\text{in}} \ll 1$).

Optimization proceeds sequentially over the denoising trajectory. Each $\varnothing_k$ is initialized from the previously optimized embedding, and the reconstruction path is advanced after each optimization:

$$\varnothing_k \leftarrow \begin{cases} \varnothing_{\text{init}}, & k = 0, \\ \varnothing_{k-1}^{\star}, & k > 0, \end{cases}$$

$$z_{k+1}^{\text{rec}} \leftarrow \hat{z}_{k+1}^{\text{rec}}(\varnothing_k^{\star}).$$

The resulting sequence $\{\varnothing_k^{\star}\}_{k=0}^{N-1}$ is reused during primary editing and refinement. In the no-NTI variant, the fixed empty-prompt embedding $\varnothing_{\text{init}}$ is used at all timesteps.

### 4.4. Decoder Self-Attention Masking

Although the edit is spatially localized, information belonging to the target object can remain in the U-Net features and propagate through self-attention. To reduce this leakage, the proposed method prevents object-region tokens from acting as contextual sources in decoder self-attention.

For a self-attention layer, let $Q_k^{(r)}$, $K_k^{(r)}$, and $V_k^{(r)}$ be the query, key, and value tensors. The standard scaled attention logits and output at denoising step $k$ are

$$S_k^{(r)} = \frac{Q_k^{(r)}\left(K_k^{(r)}\right)^{\top}}{\sqrt{d_h}}, \qquad O_{\text{orig},k}^{(r)} = \text{softmax}\left(S_k^{(r)}\right)V_k^{(r)},$$

where $d_h$ denotes the dimensionality of each attention head.

Let $m^{(r)}$ be the object token mask at the layer resolution, and let $m_j^{(r)}$ denote its entry at key-token position $j$. Here, $m_j^{(r)} = 1$ indicates that key token $j$ belongs to the object region. The implementation uses two regimes. The soft attention-masking regime is applied during the early high-noise denoising steps, when the coarse scene layout and global structure are still being formed. In this regime, all query tokens use softened background-only logits:

$$S_{\text{soft},k}^{(r)}(i,j) = \begin{cases} -\infty, & m_j^{(r)} = 1, \\ \gamma_{\text{soft}} S_k^{(r)}(i,j), & m_j^{(r)} = 0, \end{cases} \qquad \gamma_{\text{soft}} = 0.3.$$

The corresponding output is

$$O_{\text{soft},k}^{(r)} = \text{softmax}\left(S_{\text{soft},k}^{(r)}\right)V_k^{(r)}.$$

Scaling the logits makes the attention distribution less peaked for all query tokens, allowing object-region queries in particular to aggregate information from a broader set of background locations.

At later denoising steps, after the initial global layout has been established, the method switches to a hard attention-masking regime. The background-key-only logits are

$$S^{(r)}_{\text{hard},k}(i,j) = \begin{cases} -\infty, & m^{(r)}_j = 1, \\ S^{(r)}_k(i,j), & m^{(r)}_j = 0. \end{cases}$$

The corresponding output is

$$O^{(r)}_{\text{hard},k} = \text{softmax}\left(S^{(r)}_{\text{hard},k}\right) V^{(r)}_k.$$

Under both masking regimes, no query token can retrieve information from object-region keys. Importantly, object-region query tokens are still allowed to update; they can attend to background keys and therefore synthesize content based on the surrounding scene.

Let $k_{\text{soft}}$ denote the last denoising step at which the soft attention-masking regime is used. The regime-specific modified output at denoising step $k$ is

$$O^{(r)}_{\text{mod},k} = \begin{cases} O^{(r)}_{\text{soft},k}, & k \le k_{\text{soft}}, \\ O^{(r)}_{\text{hard},k}, & k > k_{\text{soft}}. \end{cases}$$

The final attention output at denoising step $k$ is then blended with the original attention output:

$$O^{(r)}_k = (1-\lambda_{\text{attn},k}) O^{(r)}_{\text{orig},k} + \lambda_{\text{attn},k} O^{(r)}_{\text{mod},k}, \qquad \lambda_{\text{attn},k} \in [0,1].$$

The strength $\lambda_{\text{attn},k}$ follows a cosine ramp over denoising steps. This avoids applying a strong attention restriction too early and increases anti-leakage control later, when object remnants and local textures are more likely to appear.

The modification is applied only to decoder self-attention layers in the U-Net upsampling blocks and only to the conditional branch of classifier-free guidance. The unconditional branch is kept unchanged as a stable reference, while the conditional prediction is guided to draw spatial context from background-region keys rather than object-region keys.

### 4.5. Primary Editing with Hard Background Anchoring

Primary editing starts from the inverted noisy latent $z^{\text{edit}}_0 = z^{\text{inv}}_0$. At timestep $t_k$, the guided prediction is computed using the caption embedding and either the optimized unconditional embedding $\varnothing^\star_k$ or the fixed empty-prompt embedding:

$$\bar{\varnothing}_k = \begin{cases} \varnothing^\star_k, & \text{masked-NTI variant}, \\ \varnothing_{\text{init}}, & \text{no-NTI variant}, \end{cases}$$

$$\epsilon^{\text{edit}}_{\text{cfg},k} = \epsilon_{\text{cfg}}\left(z^{\text{edit}}_k, t_k; \bar{\varnothing}_k, c, w_{\text{edit}}\right).$$

Decoder self-attention masking is active during the conditional U-Net prediction. A provisional denoising update is then computed as

$$z^{\text{raw,edit}}_{k+1} = \text{Step}\left(z^{\text{edit}}_k, t_k, \epsilon^{\text{edit}}_{\text{cfg},k}\right).$$

To prevent drift outside the editable region, the provisional latent is projected onto the corresponding source-image inversion pivot. For trajectory index $j$, define

$$\Pi_j(z) = z \odot M_{\text{in}} + z^{\text{inv}}_j \odot M_{\text{out}}.$$

The final update at each editing step is therefore

$$z^{\text{edit}}_{k+1} = \Pi_{k+1}\left(z^{\text{raw,edit}}_{k+1}\right).$$

This hard projection allows the diffusion model to synthesize the masked region while restoring the outside-mask latent directly from the source trajectory at every timestep.

### 4.6. Localized Renoise–Denoise Refinement

The primary editing pass may leave object remnants, texture inconsistencies, weak continuation of background patterns, or boundary artifacts inside the mask. Moreover, a single denoising pass may not be sufficient for the generated content inside the mask to become fully compatible with the fixed surrounding context, especially when the removed object is large or overlaps important background structures. To address these local errors, the method applies repeated localized renoise–denoise refinement while preserving the same outside-mask constraint. The refinement stage partially destroys the currently generated content inside the mask by adding noise and then allows the diffusion model to synthesize it again. During this process, the outside region remains locked to the original inversion trajectory and acts as a context-aware anchor for generation inside the mask. Refinement therefore provides repeated opportunities for the editable region to adapt to the fixed surrounding background.

Let $R$ denote the number of refinement rounds, let $\ell = 1, \ldots, R$ index the refinement rounds, and let $z^{(\ell-1)}_N$ denote the result of the current editing or refinement round with $z^{(0)}_N = z^{\text{edit}}_N$. An intermediate refinement index is selected as

$$k_{\text{ref}} = \lfloor \rho(N-1) \rfloor, \qquad \rho \in [0,1].$$

A smaller value of $\rho$ returns to an earlier and noisier state, providing more freedom to revise the generated content at the cost of additional denoising and reduced stability. A larger value starts closer to the final image and mainly supports minor corrections. An intermediate value therefore provides sufficient noise for revising local structure and texture without restarting generation from pure noise.

Fresh Gaussian noise $\eta^{(\ell)} \sim \mathcal{N}(0, \mathbf{I})$ is added at the corresponding timestep $t_{\text{ref}} = t_{k_{\text{ref}}}$:

$$z^{(\ell)}_{\text{noisy}} = \sqrt{\bar{\alpha}_{t_{\text{ref}}}}\, z^{(\ell-1)}_N + \sqrt{1-\bar{\alpha}_{t_{\text{ref}}}}\, \eta^{(\ell)}.$$

The outside region is then immediately restored from the time-aligned inversion trajectory:

$$z^{(\ell)}_{k_{\text{ref}}} = z^{(\ell)}_{\text{noisy}} \odot M_{\text{in}} + z^{\text{inv}}_{k_{\text{ref}}} \odot M_{\text{out}}.$$

The U-Net therefore receives a full latent in which the masked region contains newly added noise, while the outside region contains the corresponding background anchor. The preserved background can provide spatial context for the resynthesis, while hard projection prevents it from being modified.

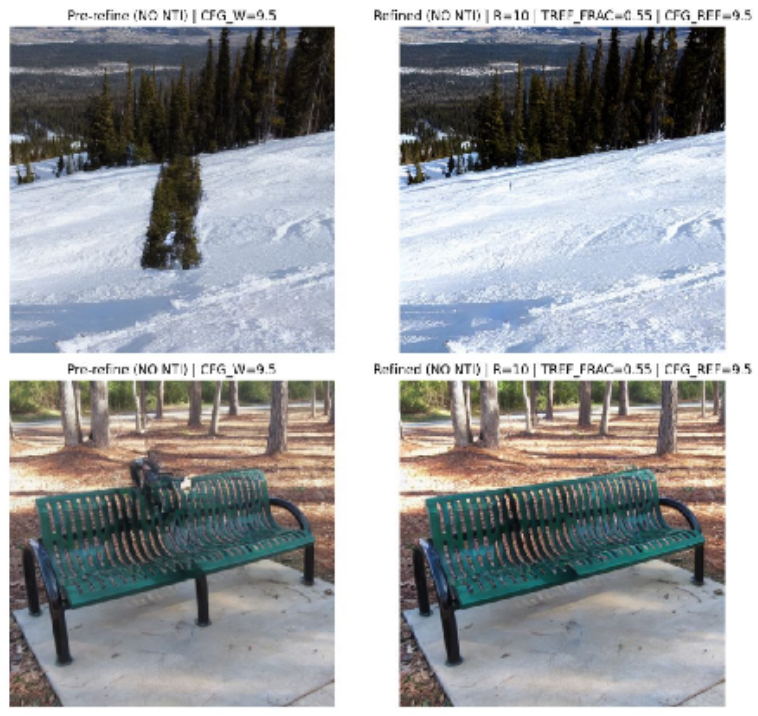

**Figure 3.** Impact of the refinement stage on reducing residual artifacts and improving boundary consistency.

Denoising then resumes from $k_{\text{ref}}$ using the same caption conditioning, timestep-specific unconditional embeddings, decoder self-attention masking, and hard outside projection:

$$z_{k+1}^{\text{raw},(\ell)} = \text{Step}\left(z_k^{(\ell)}, t_k, \epsilon_{\text{cfg}}\left(z_k^{(\ell)}, t_k; \bar{\varnothing}_k, c, w_{\text{ref}}\right)\right),$$

$$z_{k+1}^{(\ell)} = \Pi_{k+1}\left(z_{k+1}^{\text{raw},(\ell)}\right), \qquad k = k_{\text{ref}}, \ldots, N-1.$$

Thus, every refinement step can update the masked region, while the outside region is repeatedly projected back to the inversion trajectory. Each refinement round uses a new noise sample, allowing the model to explore a different local completion while keeping the visible background fixed. This repeated renoise–denoise process acts as a negotiation between the editable region and its preserved context, giving the synthesized content multiple opportunities to align its color, texture, structure, and boundary with the surrounding scene. It consequently helps reduce seams, halos, object remnants, and local inconsistencies remaining after the primary editing pass. After the final refinement round, the resulting latent is decoded to obtain the edited image $\hat{I} = D\left(z_N^{(R)}\right)$.

# 5. Experiments

## 5.1. Implementation Details

The method is implemented using Stable Diffusion v1.4 (Rombach et al. 2022) with a DDIM scheduler (Song et al. 2021) and $N = 50$ denoising steps. Input images are resized to $512 \times 512$. SAM (Kirillov et al. 2023) generates the initial object mask from user-provided clicks, after which a brush interface allows manual correction. The final mask is postprocessed through largest-component selection, hole filling, and $9 \times 9$ dilation. BLIP (J. Li et al. 2022) generates the base caption, which is encoded by the Stable Diffusion text encoder and used as the conditional prompt during inversion, masked NTI, primary editing, and refinement. The empty prompt provides the initial unconditional embedding $\varnothing_{\text{init}}$. VAE encoding and decoding are performed in FP32 to reduce reconstruction artifacts. For the masked-NTI variant, the optimization uses guidance scale $w_{\text{train}} = 7.5$, Adam with learning rate $10^{-2}$, at most 5 optimization iterations per timestep, and inside-mask weight $\lambda_{\text{in}} = 0.001$. The no-NTI variant skips this optimization and uses the fixed empty-prompt embedding $\varnothing_{\text{init}}$ at every timestep. Primary editing uses CFG scale $w_{\text{edit}} = 9.5$. Decoder self-attention masking is applied only in the U-Net upsampling blocks and only to the conditional branch of classifier-free guidance.

The attention strength $\lambda_{\text{attn},k}$ follows a cosine ramp, remaining weak during the early denoising stage and reaching full strength at later timesteps. The refinement stage uses $R = 10$ rounds, starts from the middle of the denoising trajectory with $\rho = 0.5$, uses guidance scale $w_{\text{ref}} = 9.5$, and applies the same hard outside-mask anchoring as the primary editing pass. The implementation of the proposed object-removal pipeline is available in the accompanying GitHub repository: `zero-shot-diffusion-object-removal`.

## 5.2. Evaluation Protocol

We evaluate the method on a diverse set of real-image object-removal examples. Each includes the input image, object mask, output before refinement and final refined result. The experiments compare the complete masked-NTI pipeline with a no-NTI variant in which DDIM inversion, decoder self-attention masking, hard latent anchoring, and localized refinement are retained, but the optimized timestep-specific unconditional embeddings are omitted. Object removal generally has no unique ground-truth completion because the content hidden behind the removed object is unknown. The evaluation therefore focuses on three properties: preservation of the surrounding image, local compatibility between the synthesized region and its context, and visual continuity across the mask boundary. Boundary gradient mismatch and boundary color jump measure edge and color discontinuities across the mask boundary. Outside-ring RGB and gradient differences detect halo-like changes near the edited region. Patch–context representation alignment compares CLIP vision embeddings of the synthesized region and its surrounding context (Radford et al. 2021), while local feature distance compares normalized pooled ResNet features (He et al. 2016). Ring-focused SSIM and LPIPS measure perceptual changes around the seam by constructing a composite image in which only the boundary band is replaced by the edited result (Wang et al. 2004; R. Zhang et al. 2018). As these metrics primarily target local failure modes rather than complete object-removal success, the quantitative values are interpreted together with qualitative inspection. The complete set of examples and metric values is reported in **Appendix**.

## 5.3. Qualitative Results and Ablation Analysis

The qualitative results show that the proposed pipeline can remove selected objects while preserving most of the surrounding image. Because the outside-mask latent is projected back to the corresponding DDIM inversion state after every denoising step, background structures remain largely aligned with the input. This behavior is visible in scenes containing hedges, wooden floor patterns, fences, waves, vegetation, and architectural elements, where the main changes remain concentrated within the object mask. Decoder self-attention masking reduces the propagation of object-related features into the generated region. Nevertheless, the initial editing pass can occasionally leave weak object traces, texture inconsistencies, or boundary artifacts, which are subsequently alleviated by the refinement stage. The ablation analysis further demonstrates the contribution of key components to the final edited result. In particular, we examine the effects of background-weighted masked NTI and localized renoise–denoise refinement.

The no-NTI variant shows that DDIM inversion, decoder self-attention masking, hard anchoring, and localized refinement alone are often sufficient when the surrounding background is homogeneous or contains regular textures and repetitive patterns. In such cases, the model can infer a plausible completion without additional per-timestep optimization.

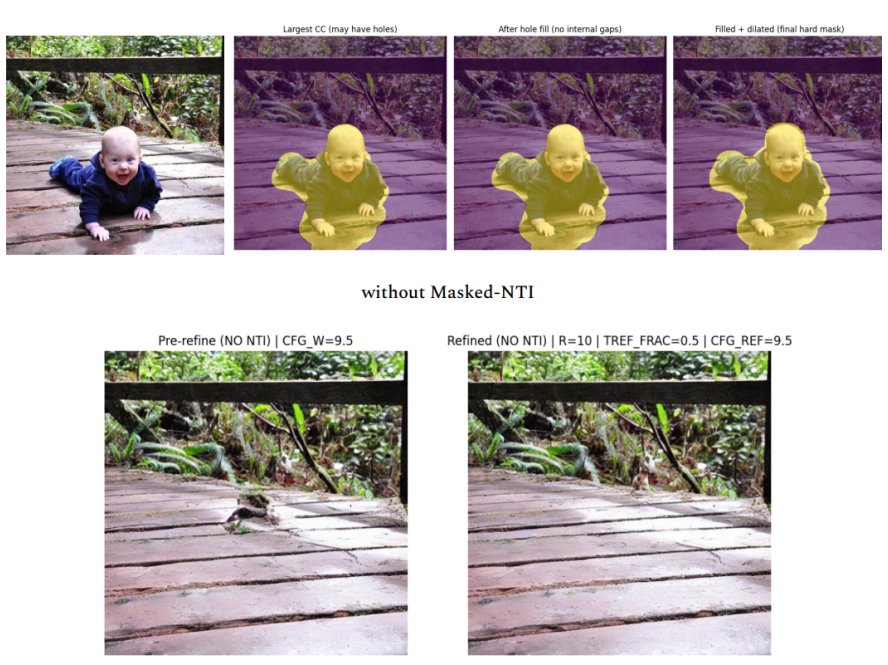

**Figure 4.** Qualitative comparison of the proposed object-removal pipeline. Input image, object mask, and result after refinement without masked NTI optimization.

Masked NTI becomes more useful when the removed region must agree with nearby structural boundaries, texture patterns, or scene geometry. Its outside-dominant objective improves the compatibility between classifier-free-guided denoising and the inversion trajectory while leaving the masked region relatively unconstrained. In challenging examples, this can reduce texture distortions, boundary inconsistencies, and local artifacts that remain in the no-NTI result.

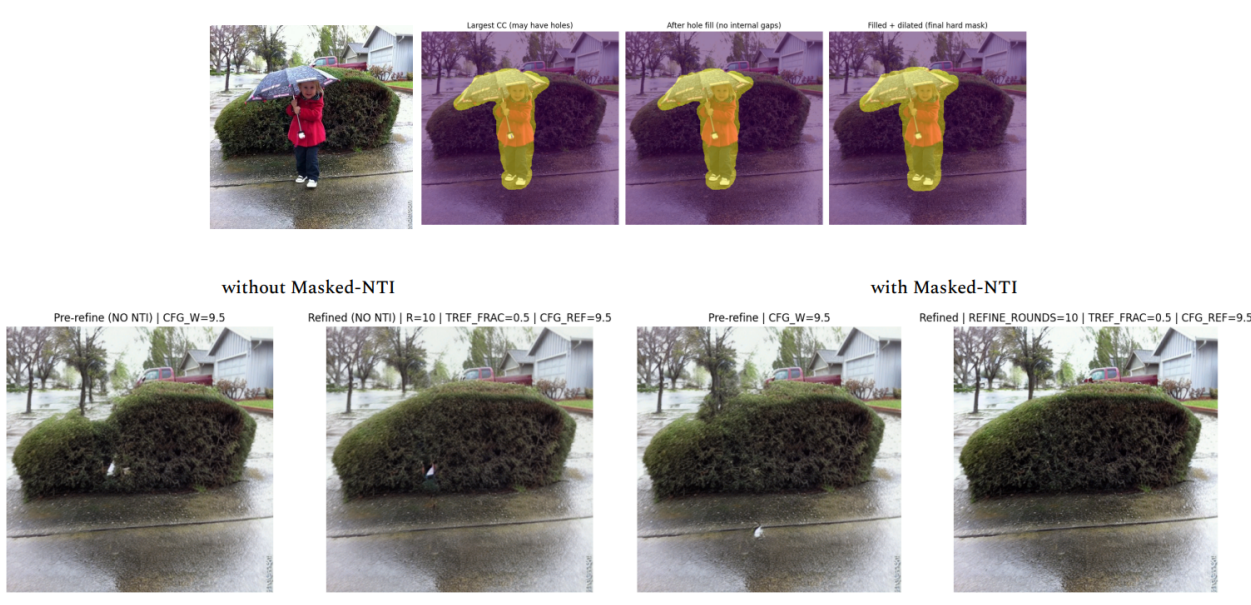

**Figure 5.** Qualitative comparison of the proposed object-removal pipeline. Input image, object mask, and results after refinement without and with masked NTI optimization. Masked NTI effectively removes the visible artifact.

The comparison between outputs before and after localized refinement shows the contribution of the refinement stage. Repeated renoise–denoise rounds reduce residual object traces and improve local texture and seam quality. This gives the model additional opportunities to revise the completion and improves texture, color, and structural consistency near the boundary. Since the outside region remains hard-anchored during every round, these improvements remain localized to the editable region. Some artifacts remain in highly challenging scenes, particularly when the target overlaps with dense structures, multiple neighboring objects, shadows, or fine background details. More aggressive settings, such as additional masked-NTI iterations, more refinement rounds, or refinement from an earlier timestep, provide greater freedom to revise the generated region but increase runtime and may reduce stability. The method therefore presents a practical trade-off between computational efficiency and output quality.

**Negative-prompt ablation.** We also tested the target-object description as a negative prompt during denoising, motivated by the timestep-dependent behavior of negative prompting reported in (Ban et al. 2024). However, this produced no consistent improvement over the final pipeline. Negative prompting acts as a global conditioning mechanism and does not spatially localize the removal, whereas decoder self-attention masking already suppresses object-region keys and hard latent anchoring restricts changes to the editable region. Under these stronger spatial controls, the additional effect of the negative prompt was negligible, and it was therefore omitted from the final method.

## 6. Limitations

The proposed method depends strongly on mask quality. Incomplete masks may leave object fragments, whereas overly large masks increase the amount of content that must be regenerated. Shadows and reflections outside the mask also remain unchanged as the preserved region is hard-anchored to the inversion trajectory; such effects must therefore be included in the editable mask. The masked-NTI variant requires per-image and per-timestep optimization, making it slower than purely forward-pass methods. Although the no-NTI variant is more efficient, it may be less reliable in structurally complex scenes. Furthermore, the quality of the generated replacement content remains limited by the pretrained diffusion model. Missing regions involving complex geometry, fine structures, or weakly represented content may therefore contain artifacts, semantic errors, or inconsistent textures. Finally, the evaluation metrics primarily measure boundary quality and local context consistency rather than complete object removal. A more comprehensive evaluation could additionally use object detectors, segmentation or vision-language models to assess whether the target object remains visible.

## 7. Conclusion

This paper presented a zero-shot object-removal framework that performs constrained latent inpainting with a frozen pretrained Stable Diffusion model. The method is designed to address the recurring technical challenges that arise in zero-shot diffusion-based image editing. It combines SAM-based mask construction, BLIP image-caption conditioning, DDIM inversion, background-weighted masked NTI, decoder self-attention masking, hard outside-mask latent anchoring, and localized renoise–denoise refinement. Together, these components suppress target-object information, preserve the surrounding scene, and improve the consistency of the synthesized region with the background. The experiments show that background-weighted masked NTI is particularly beneficial in structurally complex scenes, although the no-NTI variant is sufficient in many evaluated examples. Overall, the framework demonstrates that effective object removal can be achieved through coordinated inference-time control without task-specific training or model fine-tuning.

# Appendix

## Zero-Shot Object Removal via Attention-Guided Diffusion Inpainting

This appendix presents the full qualitative and quantitative results of the proposed object-removal pipeline. For each example, the input image and object mask are shown together with the intermediate and refined outputs of the no-NTI and masked-NTI variants. This allows the effects of decoder self-attention masking, masked NTI, and localized refinement to be compared separately.

The quantitative evaluation reports absolute and relative boundary-gradient mismatch, boundary color jump, outside-ring RGB and gradient differences, CLIP-based patch--context similarity, ResNet50-based local feature distance, ring-focused SSIM, and ring-focused LPIPS. Lower values indicate better results for all metrics except CLIP similarity and SSIM, for which higher values are better.

Since these measures focus mainly on boundary quality, local consistency, and compatibility with the surrounding context, they are interpreted together with the visual results.

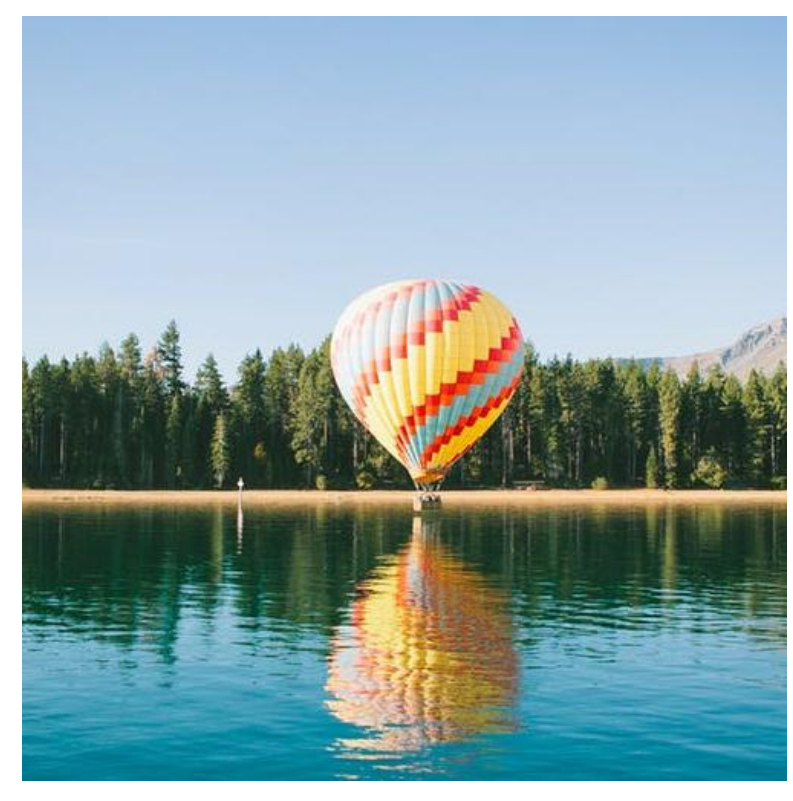

In this example, the objective is to remove the balloon together with its reflection on the water surface. In the no-NTI variant, most of the balloon and its reflection are removed successfully, but some artifacts remain in the edited region. Refinement reduces these inconsistencies, although a minor balloon-related artifact is still visible. In contrast, the masked-NTI variant removes the object and its reflection more effectively already in the intermediate result. After refinement, the remaining artifacts are eliminated, producing a clean and visually coherent final output. This example therefore shows that masked-NTI provides a clear advantage by improving removal quality and enabling artifact-free refinement.

Largest CC (may have holes)

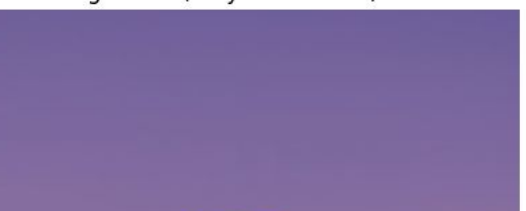

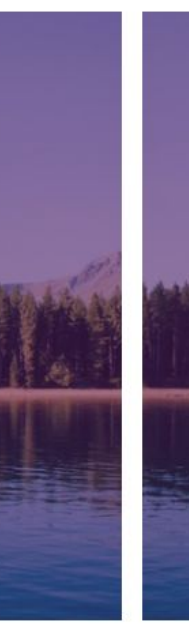

After hole fill (no internal gaps)

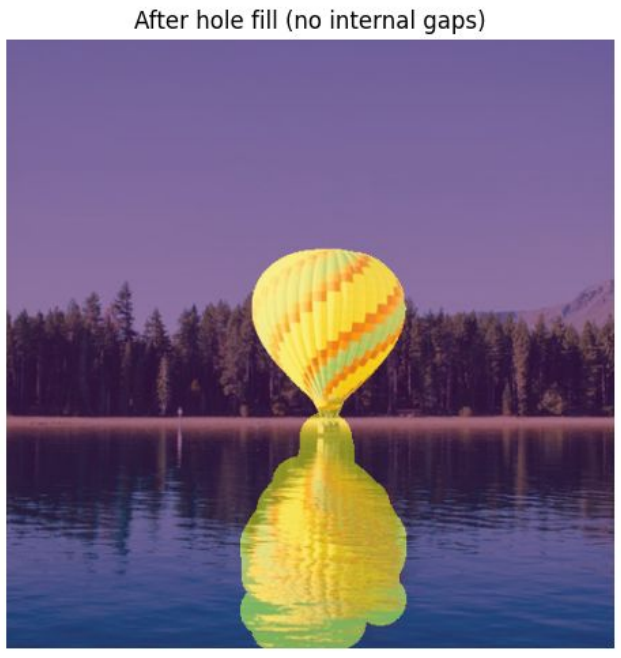

Filled + dilated (final hard mask)

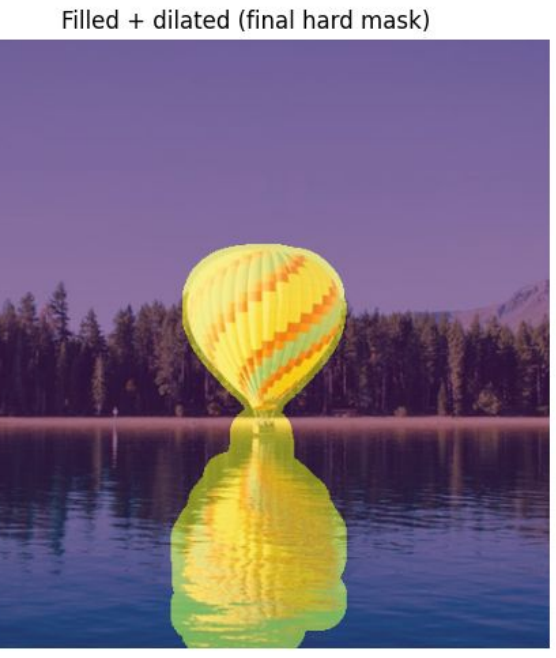

Pre-refine (NO NTI) | CFG_W=9.5

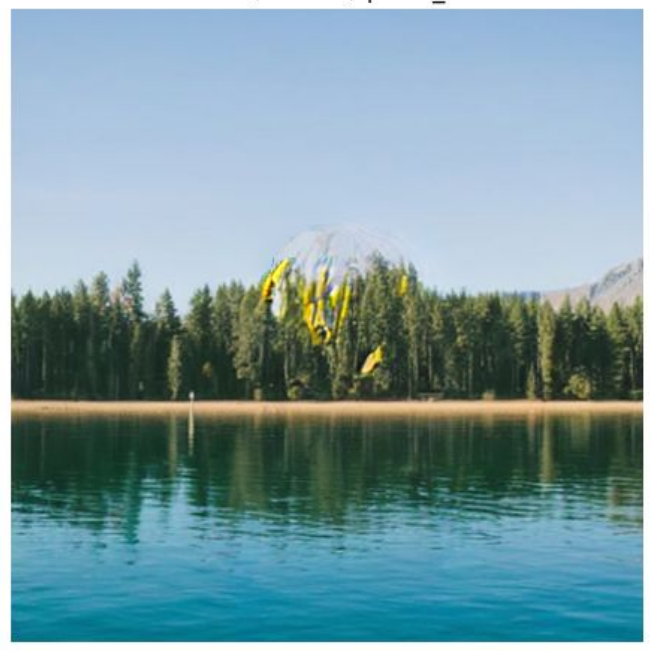

Refined (NO NTI) | R=10 | TREF_FRAC=0.55 | CFG_REF=9.5

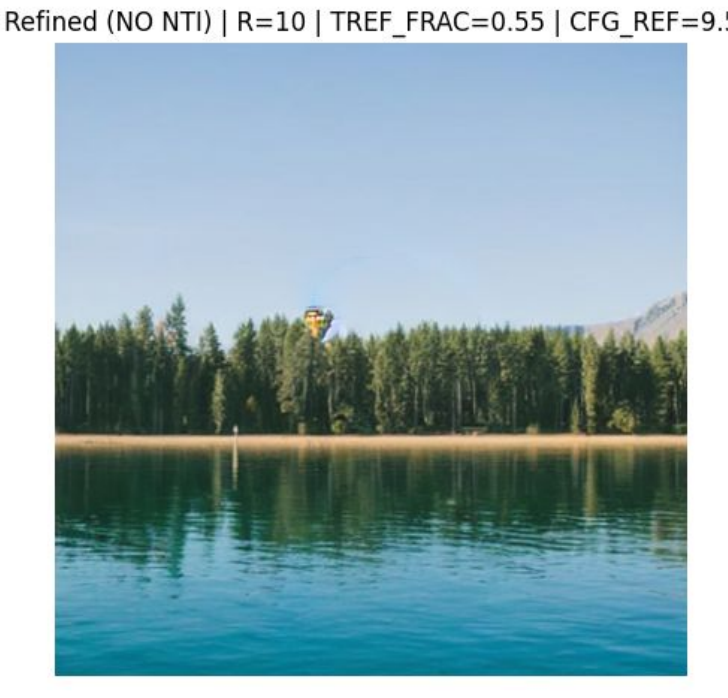

```
=================== FINAL OUTPUT EVAL (REFINED NO-NTI) ============
FINAL OUTPUT EVAL (REFINED NO-NTI)
mask: area=0.1431 | band_px=12 | ring_px=20 | ring_seam_px=8

Boundary seam (grad): in=0.2233 out=0.2008
abs_mismatch=0.0225 | rel_mismatch=0.1119  (lower better)
Boundary seam (color jump L2): 0.0161          (lower better)
Neighborhood consistency (outside rings): rgb_l2=0.0728 | grad_abs=0.0148 (lower better)
BG alignment (patch vs context): sim=0.8366 (CLIP) (higher better)
Local feature distance (patch vs context): d=1.0555 (ResNet50) (lower better)
SSIM_ring (orig vs ring_mix): 0.9647         (higher better)
LPIPS_ring (orig vs ring_mix): 0.0233        (lower better)

=================================================================
```

Pre-refine | CFG_W=9.5

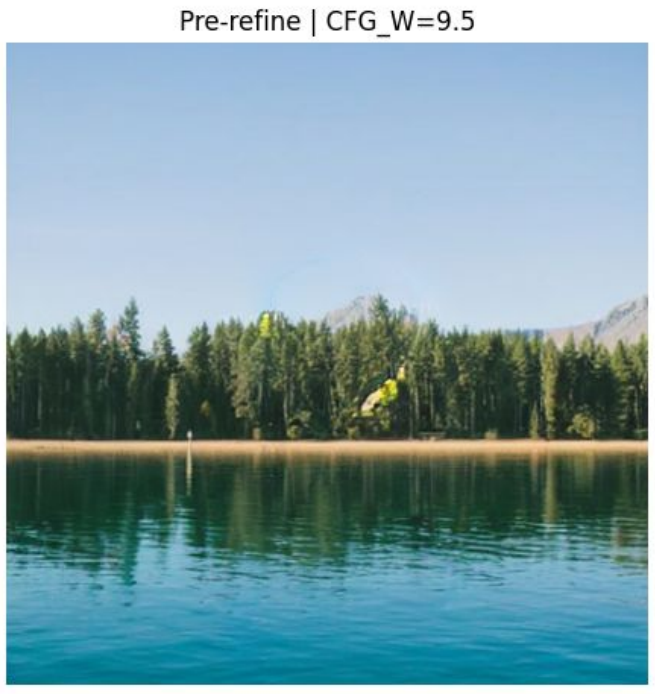

Refined | REFINE_ROUNDS=10 | TREF_FRAC=0.55 | CFG_REF=10.5

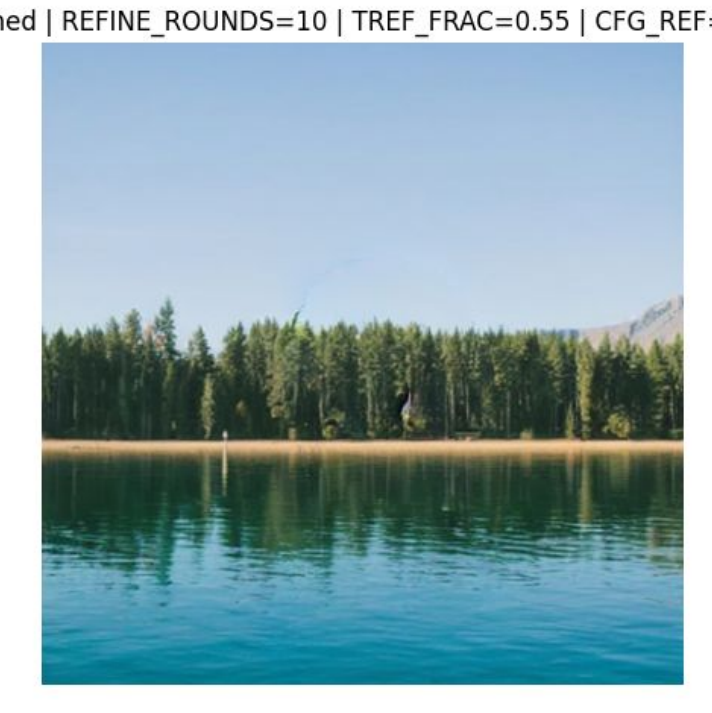

```
=================== FINAL OUTPUT EVAL (REFINED) ===================
FINAL OUTPUT EVAL (REFINED)
mask: area=0.1431 | band_px=12 | ring_px=20 | ring_seam_px=8

Boundary seam (grad): in=0.2127 out=0.2109
abs_mismatch=0.0019 | rel_mismatch=0.0089  (lower better)
Boundary seam (color jump L2): 0.0193          (lower better)
Neighborhood consistency (outside rings): rgb_l2=0.0743 | grad_abs=0.0186 (lower better)
BG alignment (patch vs context): sim=0.8301 (CLIP) (higher better)
Local feature distance (patch vs context): d=1.0496 (ResNet50) (lower better)
SSIM_ring (orig vs ring_mix): 0.9654         (higher better)
LPIPS_ring (orig vs ring_mix): 0.0214        (lower better)

=================================================================
```

The quantitative results are largely consistent with qualitative observations. Compared with the no-NTI variant, the masked-NTI result achieves substantially lower absolute and relative boundary-gradient mismatch, a lower local feature distance, and slightly improved SSIM and LPIPS, all of which indicate better local coherence and patch quality. At the same time, the no-NTI variant remains slightly better in boundary color jump, outside-ring consistency, and CLIP-based background alignment. Overall, the metrics support the qualitative observation that masked NTI improves the compatibility of the generated region with the surrounding scene and leads to a cleaner final reconstruction.

Target x0 (inv_xt[-1])

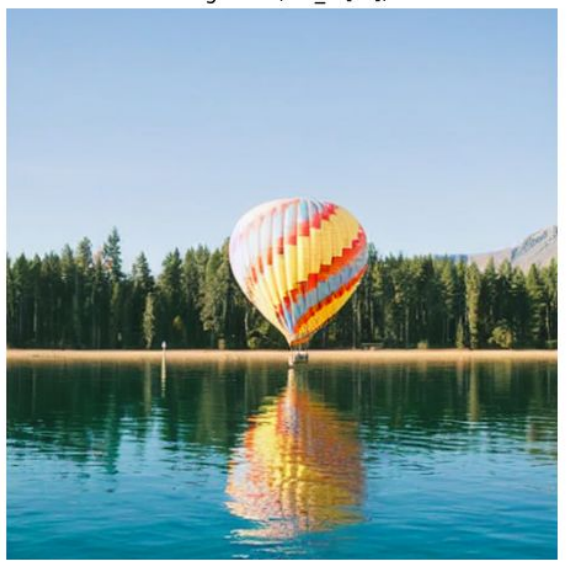

Recon FREE x0 (NO reinjection)

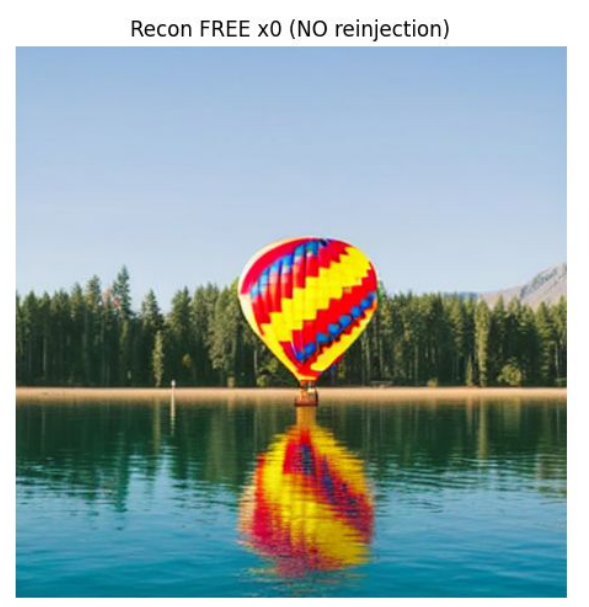

reconstruction with masked NTI

$\lambda_{in} \ll 1$

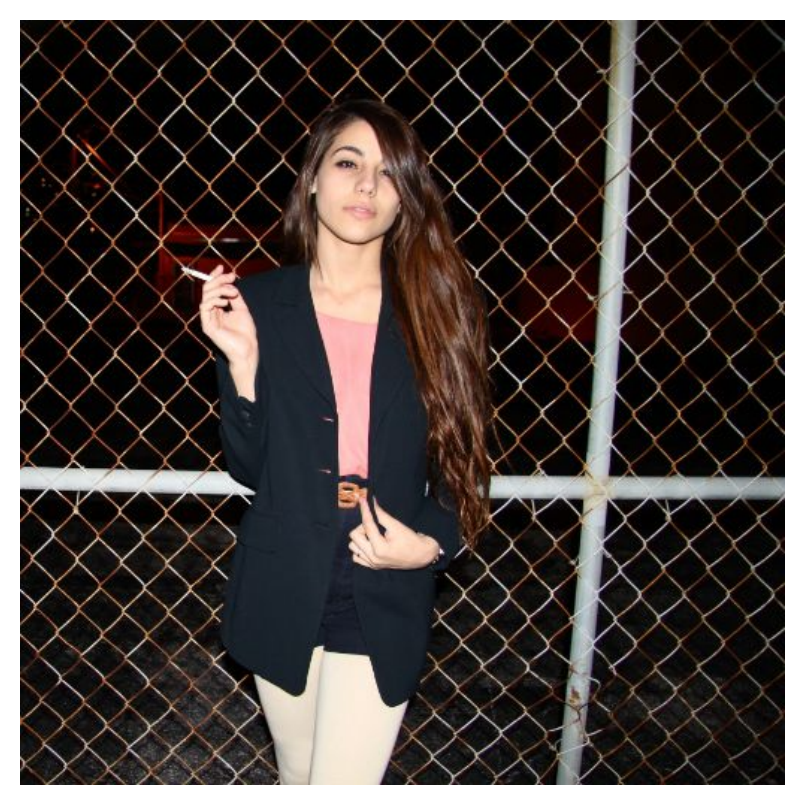

In this example, the objective is to remove the girl holding the cigarette from the scene. In the no-NTI variant, decoder self-attention masking removes most of the target object before refinement, although some artifacts remain in the edited region. The refinement stage effectively removes these inconsistencies and produces a clean final output. Compared with the masked-NTI variant, no noticeable qualitative improvement or clear advantage is observed, indicating that decoder self-attention masking followed by refinement is sufficient in this case.

Largest CC (may have holes)

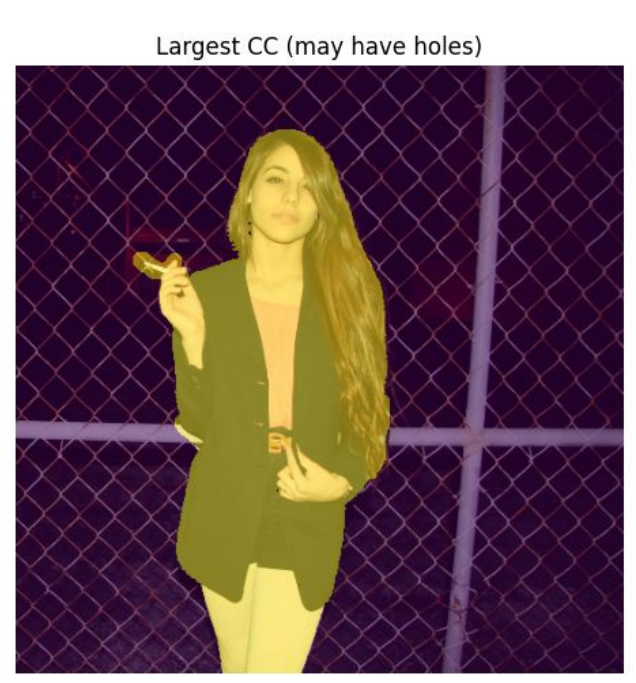

After hole fill (no internal gaps)

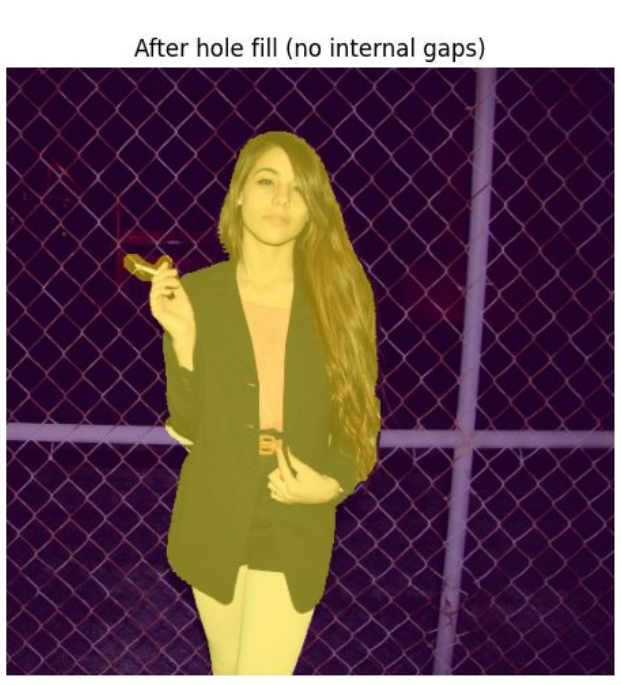

Filled + dilated (final hard mask)

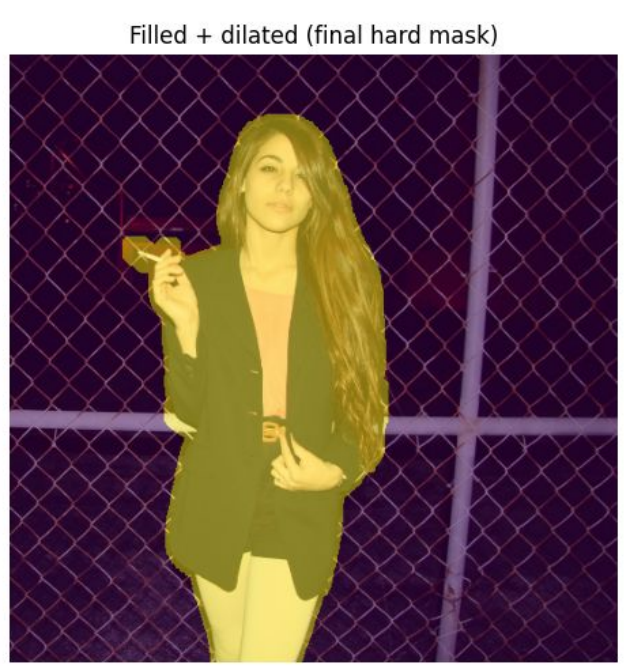

Pre-refine (NO NTI) | CFG_W=9.5

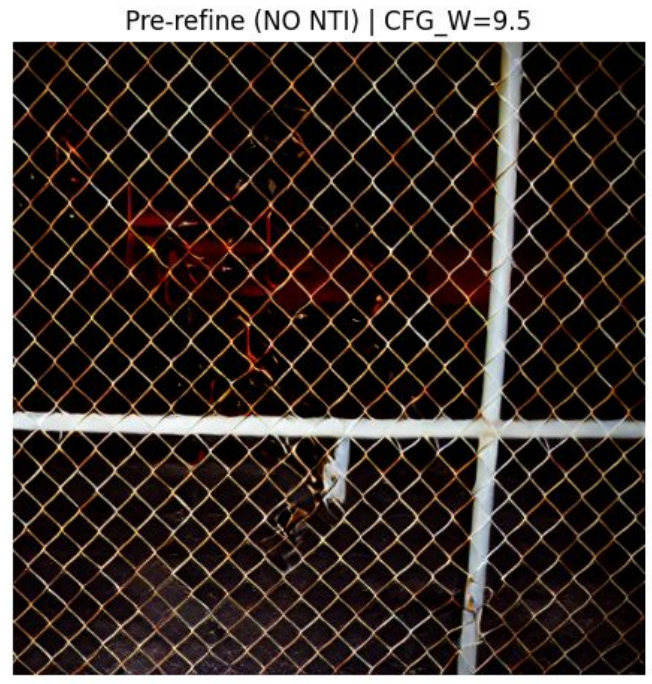

Refined (NO NTI) | R=10 | TREF_FRAC=0.55 | CFG_REF=9.5

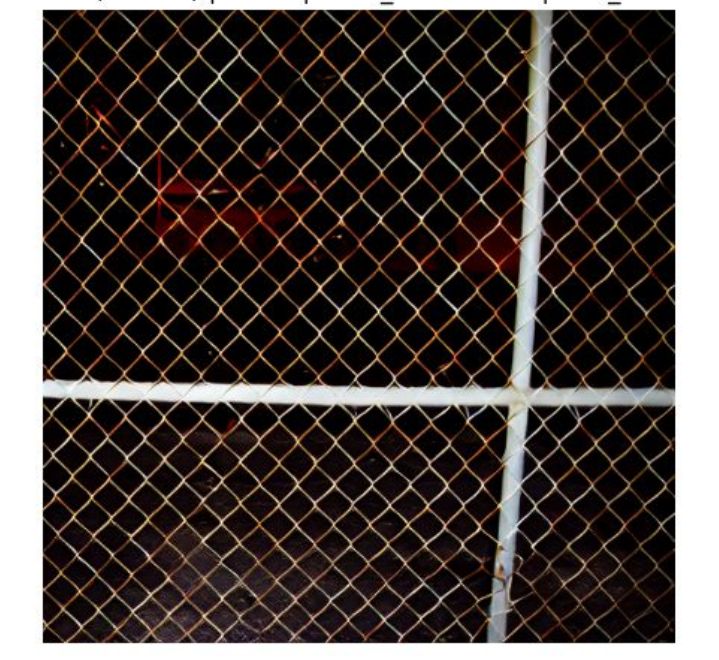

=================== FINAL OUTPUT EVAL (REFINED NO-NTI) ============
Evaluation metrics (NO-NTI)
FINAL OUTPUT EVAL (REFINED NO-NTI)
mask: area=0.2545 | band_px=12 | ring_px=20 | ring_seam_px=8

Boundary seam (grad): in=0.6371 out=0.6333
abs_mismatch=0.0038 | rel_mismatch=0.0060 (lower better)
Boundary seam (color jump L2): 0.0195 (lower better)
Neighborhood consistency (outside rings): rgb_12=0.0137 | grad_abs=0.0036 (lower better)
BG alignment (patch vs context): sim=0.8626 (CLIP) (higher better)
Local feature distance (patch vs context): d=0.7354 (ResNet50) (lower better)
SSIM_ring (orig vs ring_mix): 0.9479 (higher better)
LPIPS_ring (orig vs ring_mix): 0.0355 (lower better)

==================================================================

Pre-refine | CFG_W=9.5

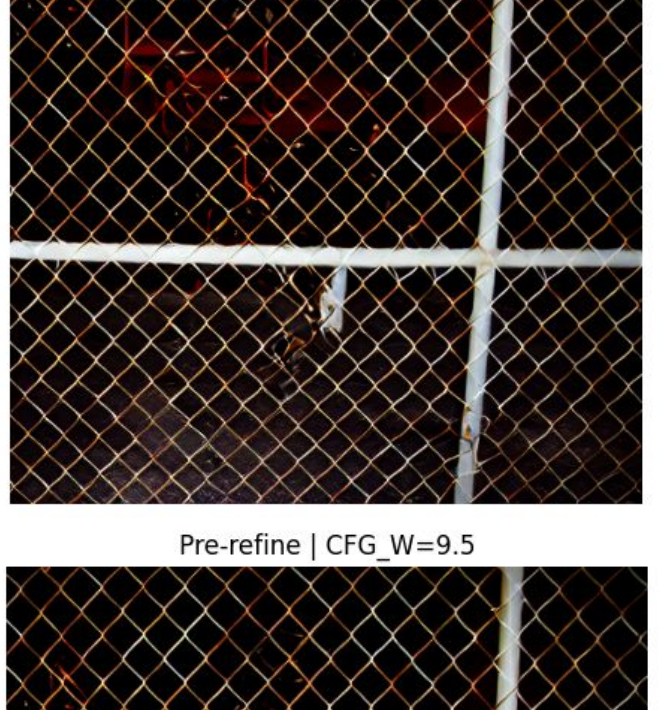

Refined | REFINE_ROUNDS=10 | TREF_FRAC=0.55 | CFG_REF=10.5

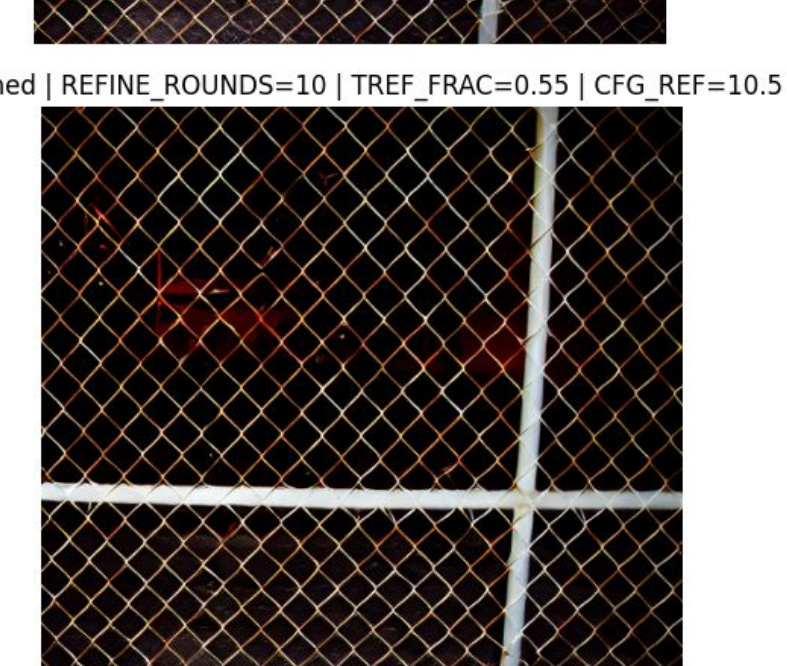

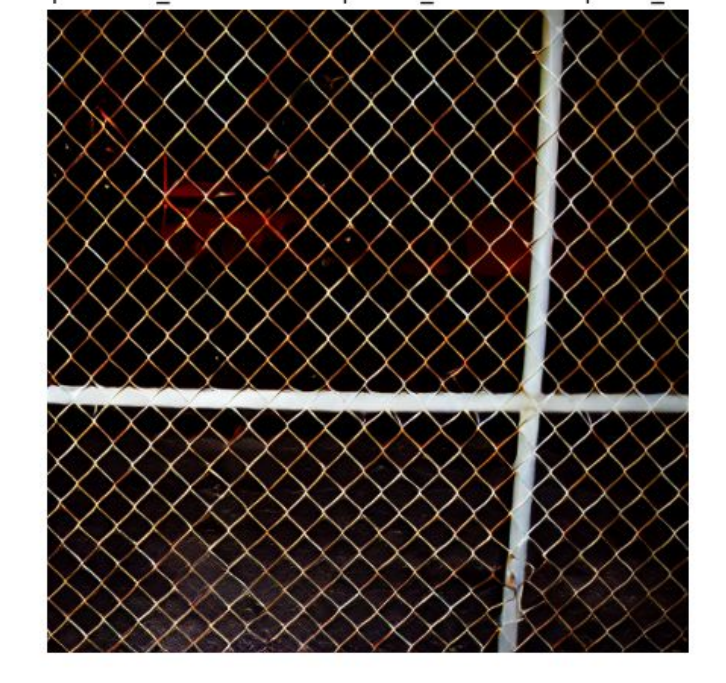

=================== FINAL OUTPUT EVAL (REFINED) ===================
FINAL OUTPUT EVAL (REFINED)
mask: area=0.2545 | band_px=12 | ring_px=20 | ring_seam_px=8

Boundary seam (grad): in=0.6273 out=0.6285
abs_mismatch=0.0013 | rel_mismatch=0.0020 (lower better)
Boundary seam (color jump L2): 0.0176 (lower better)
Neighborhood consistency (outside rings): rgb_12=0.0132 | grad_abs=0.0035 (lower better)
BG alignment (patch vs context): sim=0.8750 (CLIP) (higher better)
Local feature distance (patch vs context): d=0.7324 (ResNet50) (lower better)
SSIM_ring (orig vs ring_mix): 0.9476 (higher better)
LPIPS_ring (orig vs ring_mix): 0.0346 (lower better)

==================================================================

The quantitative results slightly favor the masked-NTI variant. It achieves lower boundary-gradient mismatch and color jump, better outside-ring RGB and gradient consistency, higher CLIP-based background alignment, lower local feature distance, and lower ring-focused LPIPS. The no-NTI result obtains only a marginally higher ring-focused SSIM. However, the differences are small, supporting the qualitative observation that both variants produce similarly successful final results and that masked NTI provides only a limited advantage in this example.

Target x0 (inv_xt[-1])

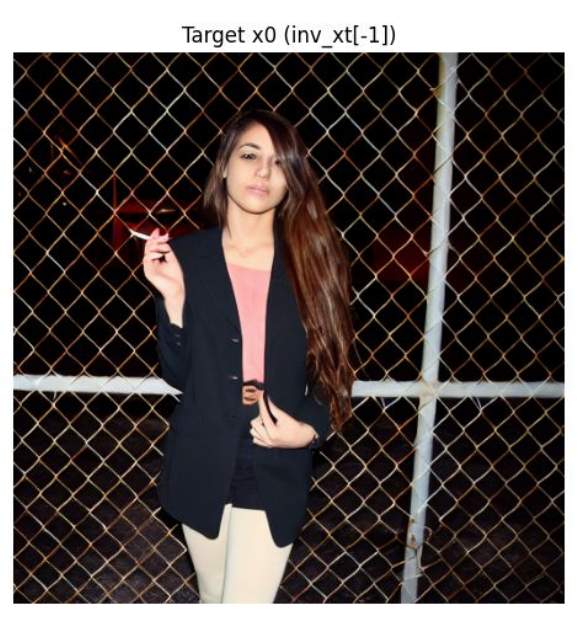

Recon FREE x0 (NO reinjection)

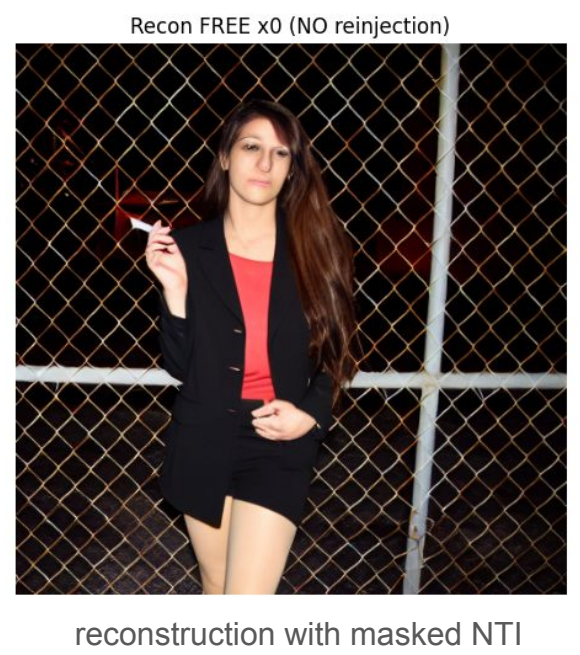

reconstruction with masked NTI

$\lambda_{in} \ll 1$

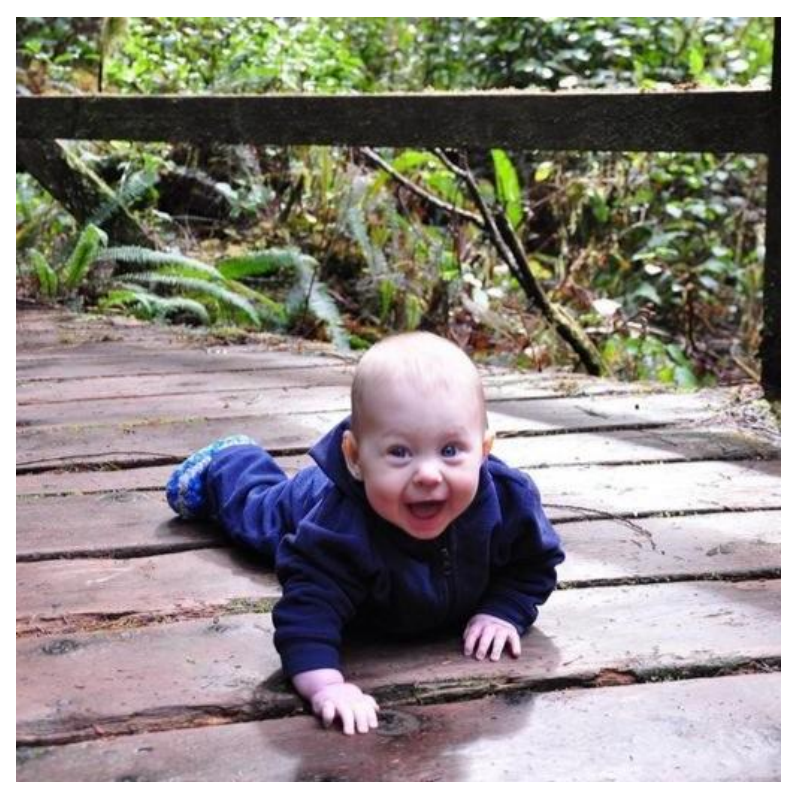

In this example, the objective is to remove the baby from the image. The no-NTI variant already removes the target object successfully using decoder self-attention masking, although some artifacts remain before refinement. The refinement stage effectively corrects these artifacts, producing a clean and visually coherent output. Compared with the NTI-based variant, no major qualitative improvement is observed, suggesting that attention masking and refinement are sufficient for this example and that the additional NTI stage provides no significant advantage.

Largest CC (may have holes)

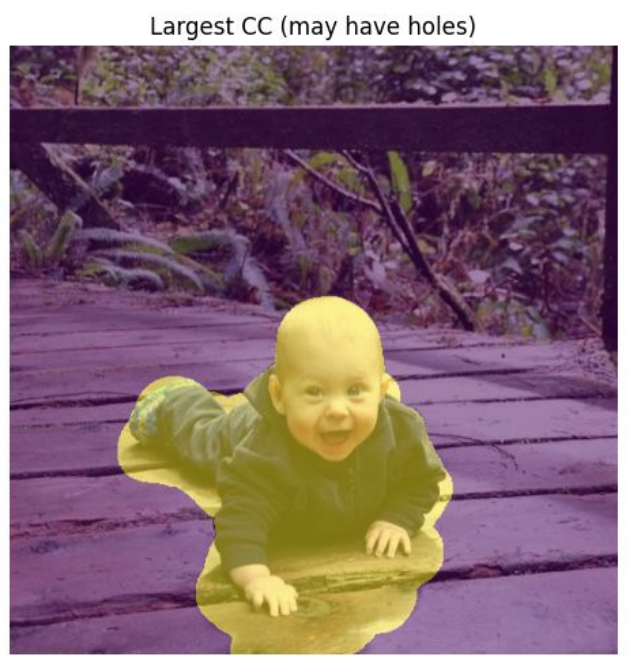

After hole fill (no internal gaps)

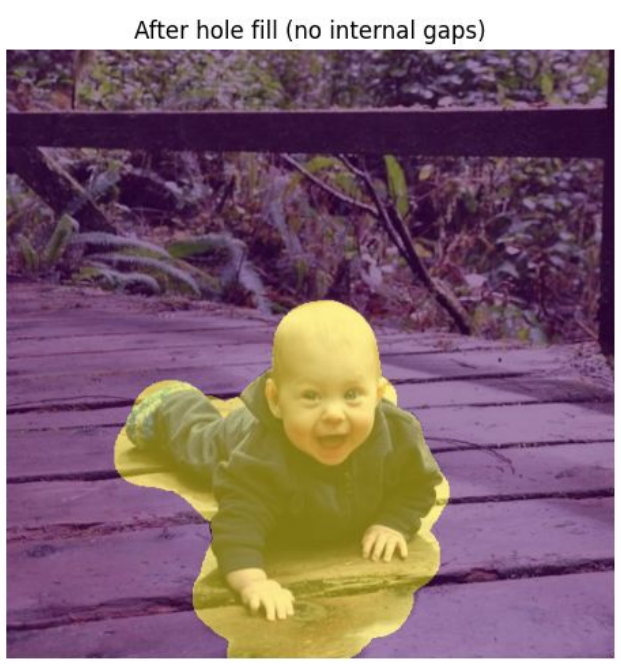

Filled + dilated (final hard mask)

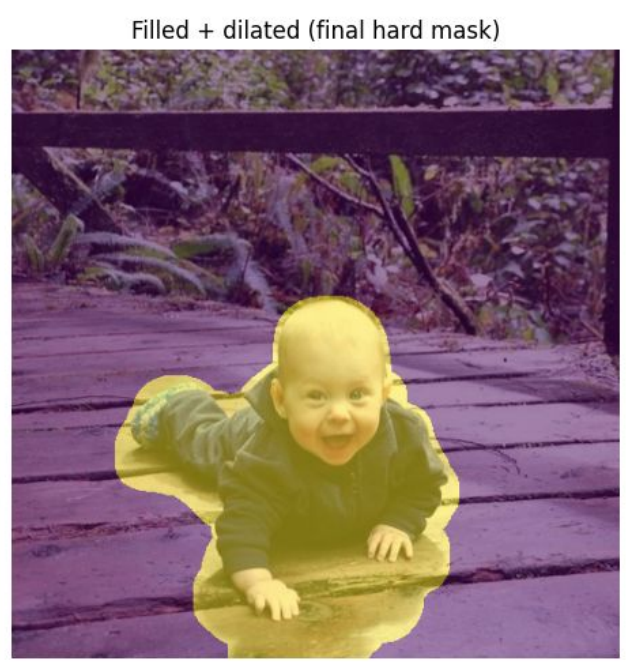

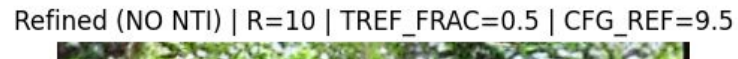

Pre-refine (NO NTI) | CFG_W=9.5

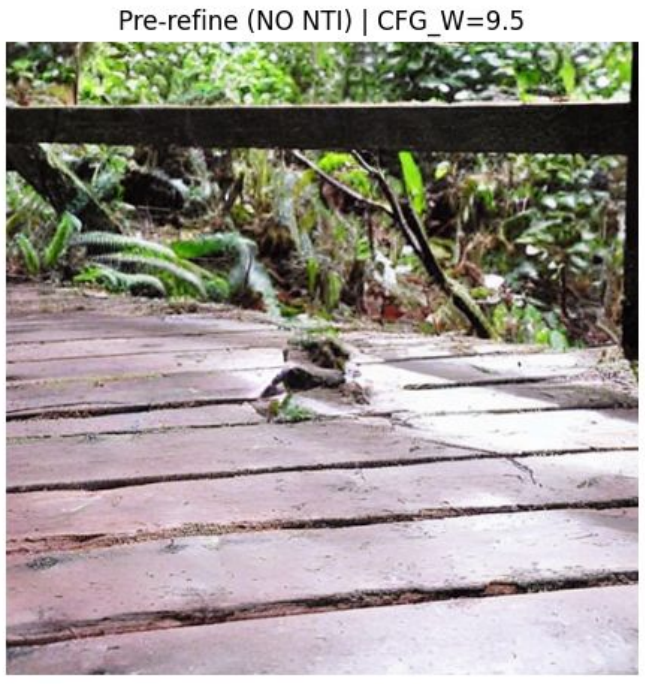

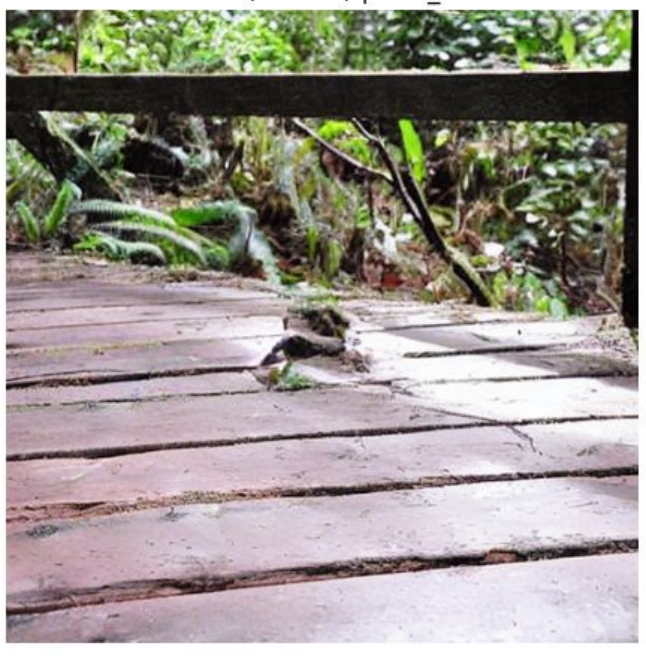

Refined (NO NTI) | R=10 | TREF_FRAC=0.5 | CFG_REF=9.5

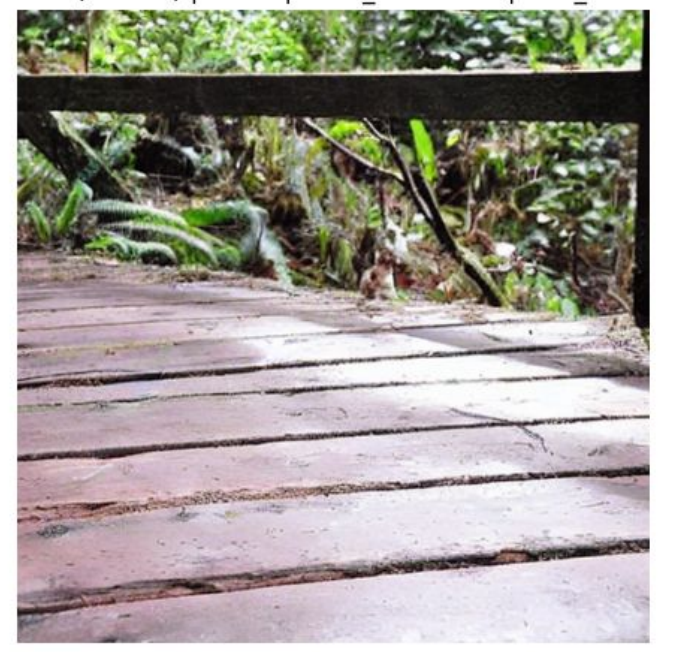

```
=================== FINAL OUTPUT EVAL (REFINED NO-NTI) ============
FINAL OUTPUT EVAL (REFINED NO-NTI)
mask: area=0.2206 | band_px=12 | ring_px=20 | ring_seam_px=8

Boundary seam (grad): in=0.4252 out=0.3961
abs_mismatch=0.0291 | rel_mismatch=0.0735  (lower better)
Boundary seam (color jump L2): 0.0298           (lower better)
Neighborhood consistency (outside rings): rgb_l2=0.0765 | grad_abs=0.0609 (lower better)
BG alignment (patch vs context): sim=0.7852 (CLIP) (higher better)
Local feature distance (patch vs context): d=0.9570 (ResNet50) (lower better)
SSIM_ring (orig vs ring_mix): 0.9606         (higher better)
LPIPS_ring (orig vs ring_mix): 0.0325       (lower better)

=================================================================
```

Pre-refine | CFG_W=9.5

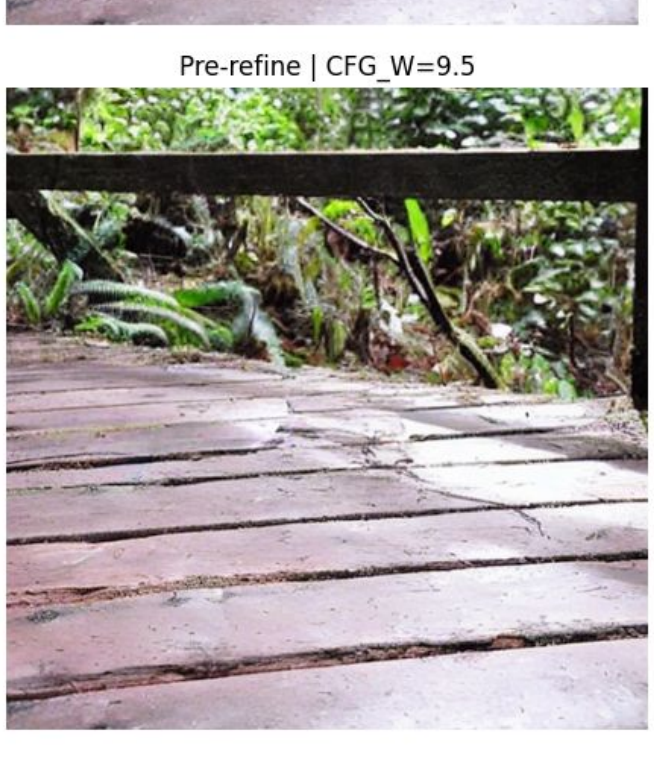

Refined | REFINE_ROUNDS=10 | TREF_FRAC=0.5 | CFG_REF=9.5

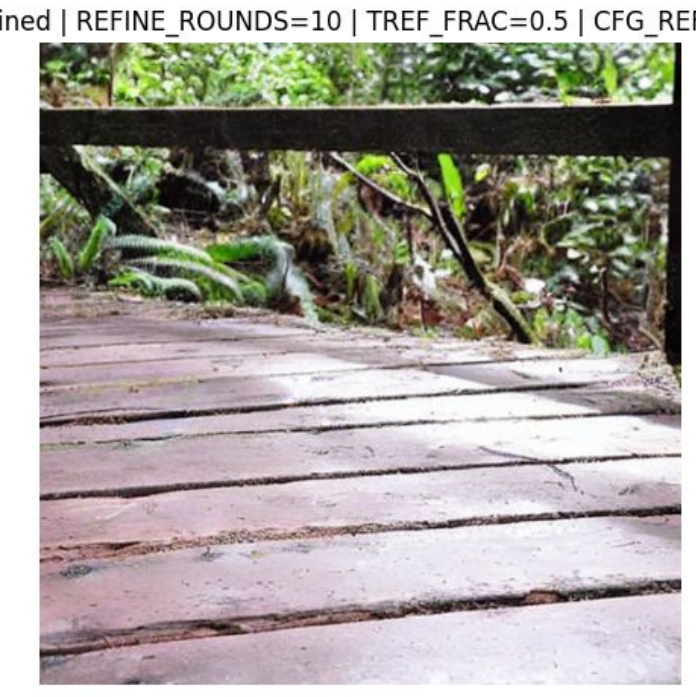

```
=================== FINAL OUTPUT EVAL (REFINED) ===================
FINAL OUTPUT EVAL (REFINED)
mask: area=0.2206 | band_px=12 | ring_px=20 | ring_seam_px=8

Boundary seam (grad): in=0.4254 out=0.3969
abs_mismatch=0.0284 | rel_mismatch=0.0716  (lower better)
Boundary seam (color jump L2): 0.0264           (lower better)
Neighborhood consistency (outside rings): rgb_l2=0.0750 | grad_abs=0.0616 (lower better)
BG alignment (patch vs context): sim=0.7752 (CLIP) (higher better)
Local feature distance (patch vs context): d=0.9297 (ResNet50) (lower better)
SSIM_ring (orig vs ring_mix): 0.9606         (higher better)
LPIPS_ring (orig vs ring_mix): 0.0315       (lower better)

=================================================================
```

The quantitative results are very similar across the two variants. The masked-NTI result achieves slightly lower boundary-gradient mismatch, color jump, outside-ring RGB difference, local feature distance, and ring-focused LPIPS, while the no-NTI result obtains slightly better CLIP-based background alignment and outside-ring gradient consistency. The ring-focused SSIM is identical for both variants. These small differences support the qualitative observation that masked NTI provides only a limited improvement in this relatively straightforward example.

Target x0 (inv_xt[-1])

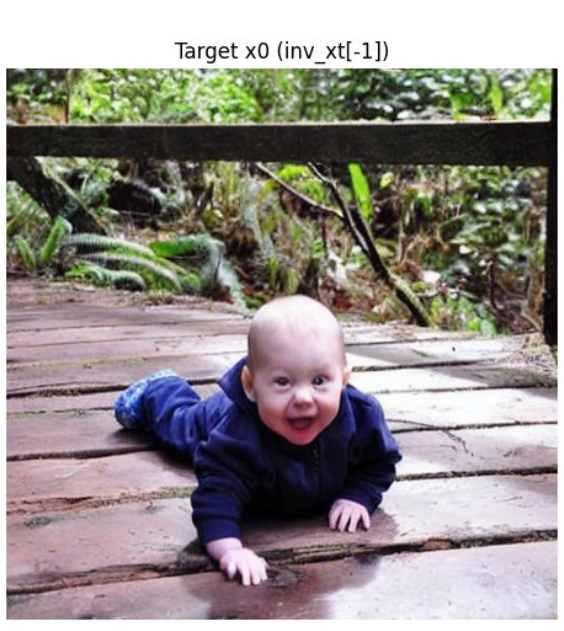

Recon FREE x0 (NO reinjection)

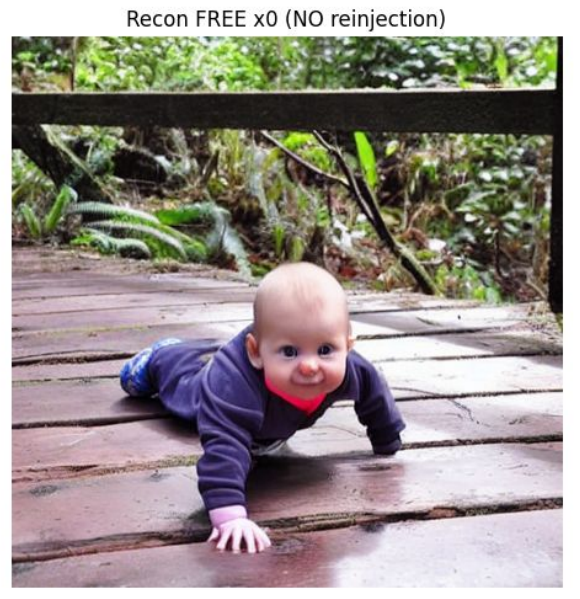

reconstruction with masked NTI

$\lambda_{in} \ll 1$

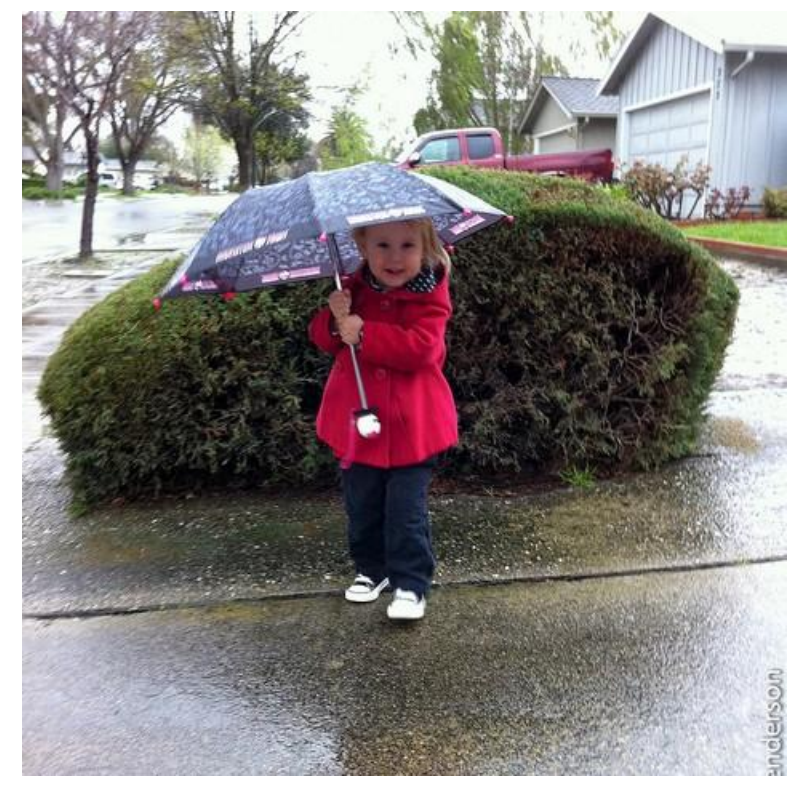

In this example, the objective is to remove the girl from the scene while preserving the surrounding background. In the no-NTI variant, the pre-refinement result after decoder self-attention masking still contains visible artifacts inside the masked region. The refinement stage reduces these inconsistencies, but some artifacts remain. In contrast, the NTI-based variant produces a noticeably cleaner intermediate result, indicating better background awareness and a more compatible denoising trajectory. After refinement, the remaining artifacts are removed, producing a clean and coherent final result in which the target object is successfully removed.

Largest CC (may have holes)

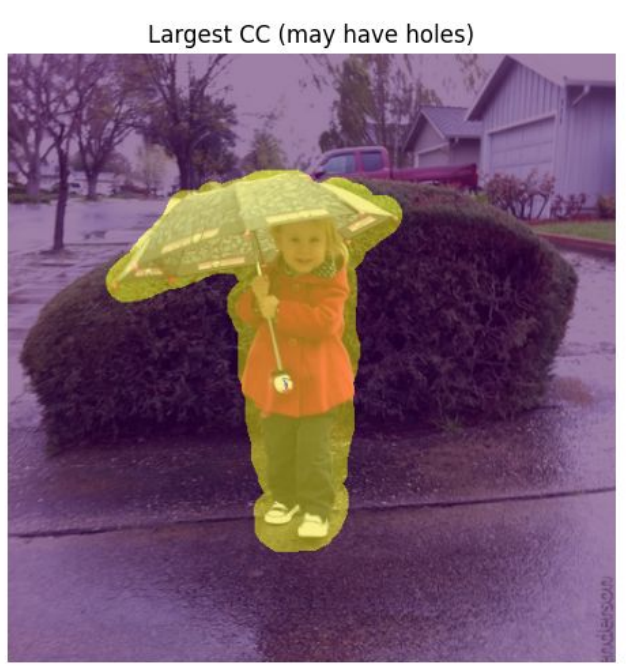

After hole fill (no internal gaps)

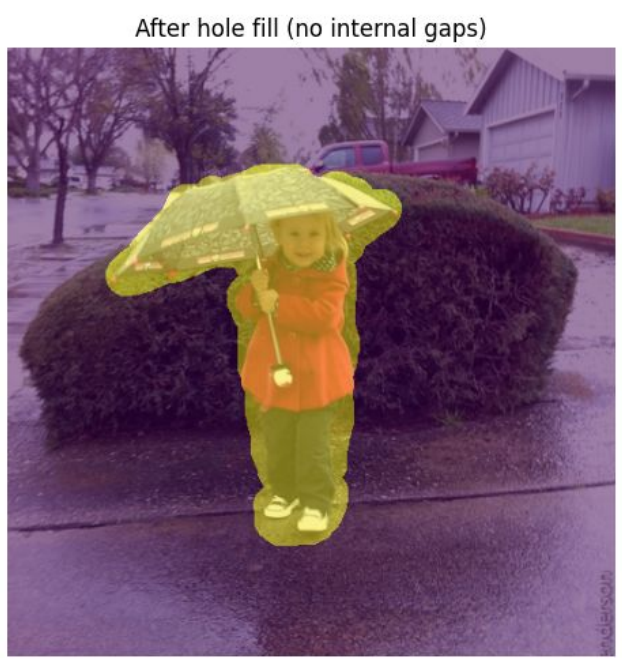

Filled + dilated (final hard mask)

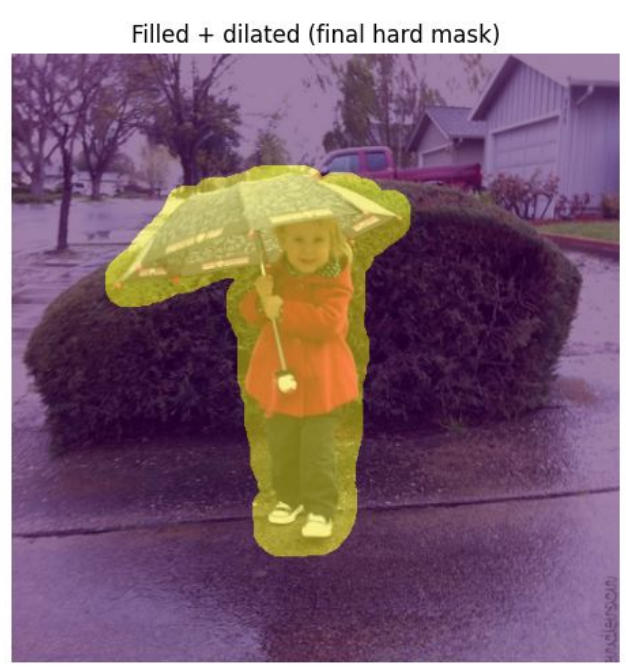

Pre-refine (NO NTI) | CFG_W=9.5

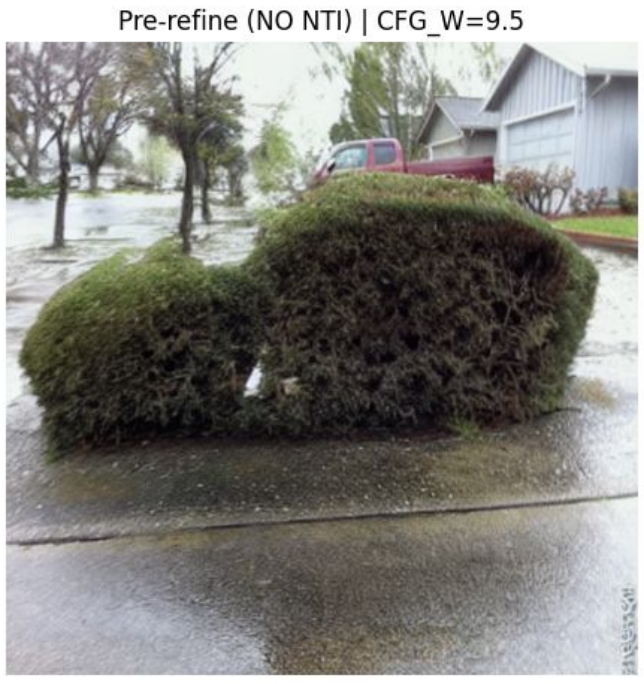

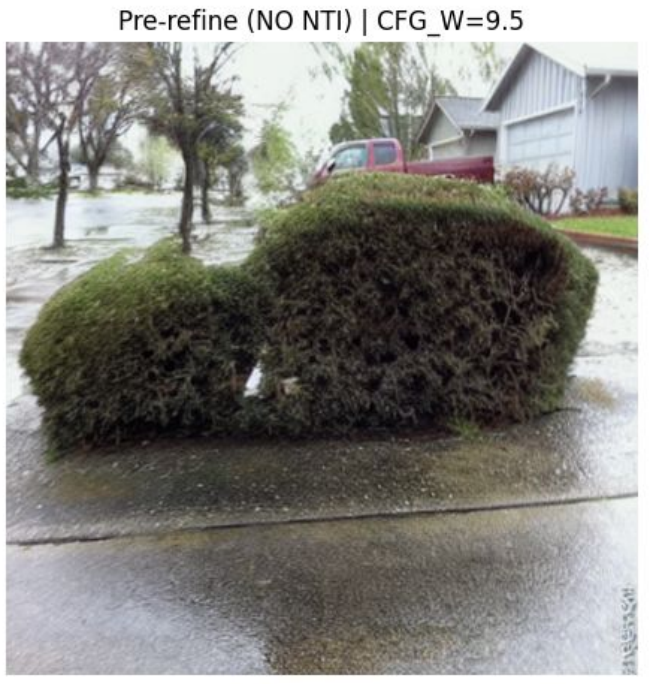

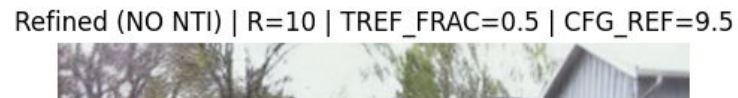

Refined (NO NTI) | R=10 | TREF_FRAC=0.5 | CFG_REF=9.5

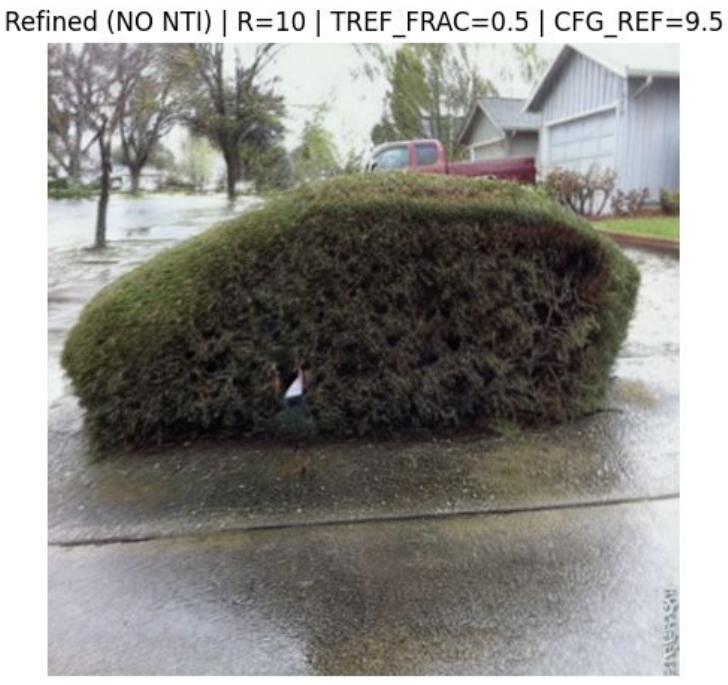

```
==================== FINAL OUTPUT EVAL (REFINED NO-NTI) =============
FINAL OUTPUT EVAL (REFINED NO-NTI)
mask: area=0.1630 | band_px=12 | ring_px=20 | ring_seam_px=8

Boundary seam (grad): in=0.2765 out=0.2632
abs_mismatch=0.0133 | rel_mismatch=0.0507  (lower better)
Boundary seam (color jump L2): 0.0595          (lower better)
Neighborhood consistency (outside rings): rgb_l2=0.0262 | grad_abs=0.0113 (lower better)
BG alignment (patch vs context): sim=0.7019 (CLIP) (higher better)
Local feature distance (patch vs context): d=0.9961 (ResNet50) (lower better)
SSIM_ring (orig vs ring_mix): 0.9466         (higher better)
LPIPS_ring (orig vs ring_mix): 0.0245        (lower better)

==================================================================
```

Pre-refine | CFG_W=9.5

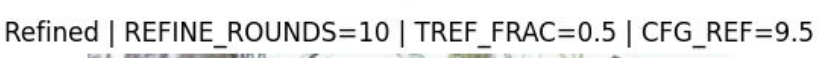

Refined | REFINE_ROUNDS=10 | TREF_FRAC=0.5 | CFG_REF=9.5

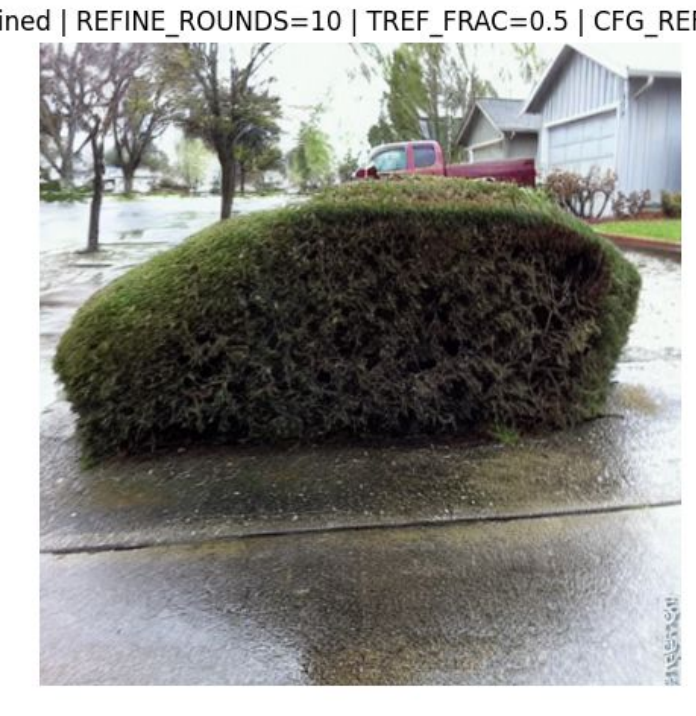

```
==================== FINAL OUTPUT EVAL (REFINED) ====================
FINAL OUTPUT EVAL (REFINED)
mask: area=0.1630 | band_px=12 | ring_px=20 | ring_seam_px=8

Boundary seam (grad): in=0.3488 out=0.3312
abs_mismatch=0.0176 | rel_mismatch=0.0532  (lower better)
Boundary seam (color jump L2): 0.0726          (lower better)
Neighborhood consistency (outside rings): rgb_l2=0.0250 | grad_abs=0.0133 (lower better)
BG alignment (patch vs context): sim=0.6926 (CLIP) (higher better)
Local feature distance (patch vs context): d=0.9560 (ResNet50) (lower better)
SSIM_ring (orig vs ring_mix): 0.9455         (higher better)
LPIPS_ring (orig vs ring_mix): 0.0188        (lower better)

====================================================================
```

The quantitative metrics show a mixed pattern and do not uniformly favor either variant. The no-NTI result achieves slightly lower boundary-gradient mismatch, color jump, and outside-ring gradient difference, whereas the masked-NTI result obtains better local feature distance and lower ring-focused LPIPS. The remaining values are close between the two variants. This indicates that the visible improvement produced by masked NTI is captured mainly by perceptual and feature-based measures, while the local seam metrics do not fully reflect the qualitative reduction of artifacts inside the removed region.

Target x0 (inv_xt[-1])

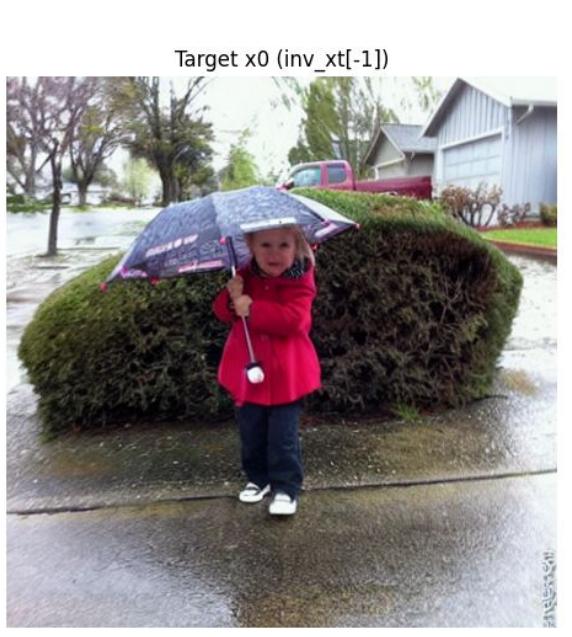

Recon FREE x0 (NO reinjection)

reconstruction with masked NTI

$\lambda_{in} \ll 1$

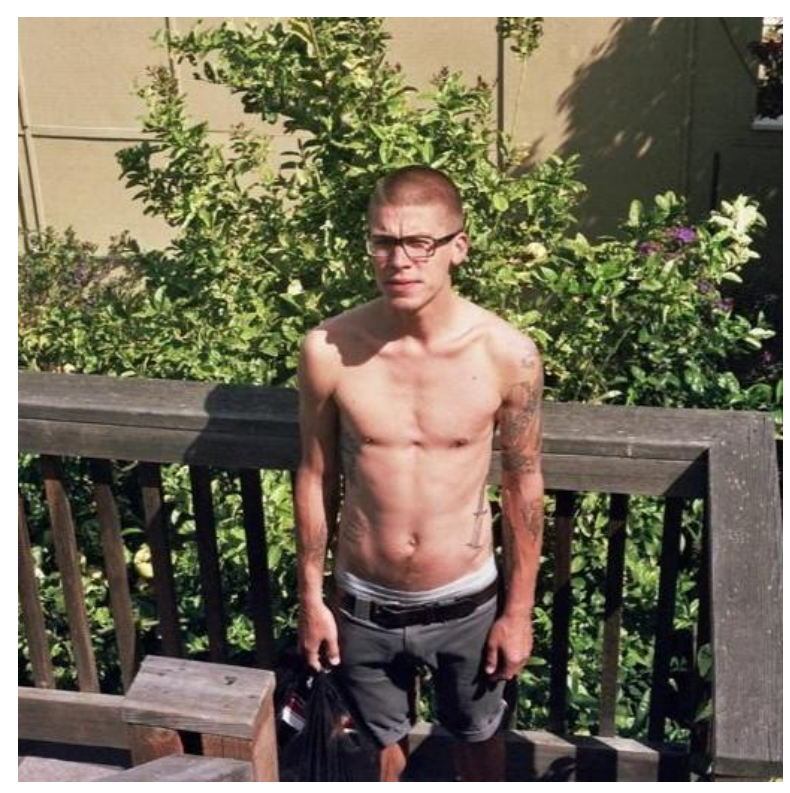

In this example, the objective is to remove the young man from the porch. The no-NTI variant produces a reasonable pre-refinement result, although some visible artifacts remain in the removed region. The refinement stage effectively removes these artifacts, resulting in a clean and coherent output. The NTI-based variant does not provide any noticeable additional improvement in this case.

Largest CC (may have holes)

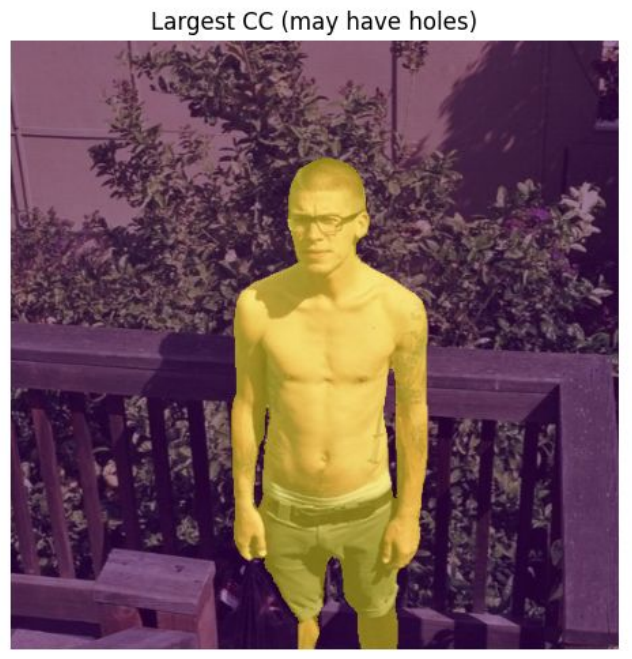

After hole fill (no internal gaps)

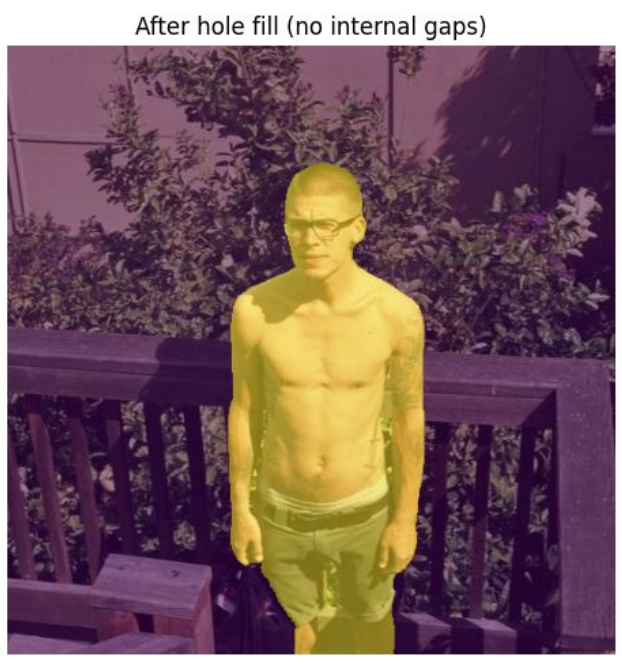

Filled + dilated (final hard mask)

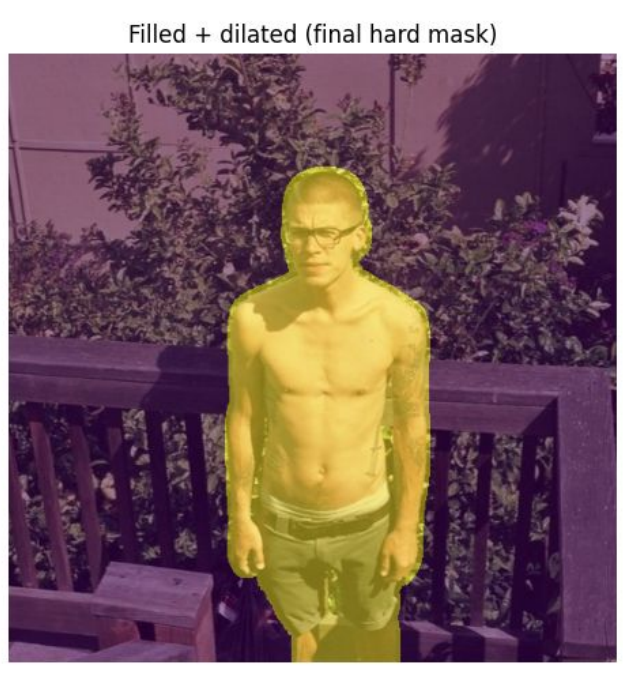

Pre-refine (NO NTI) | CFG_W=9.5

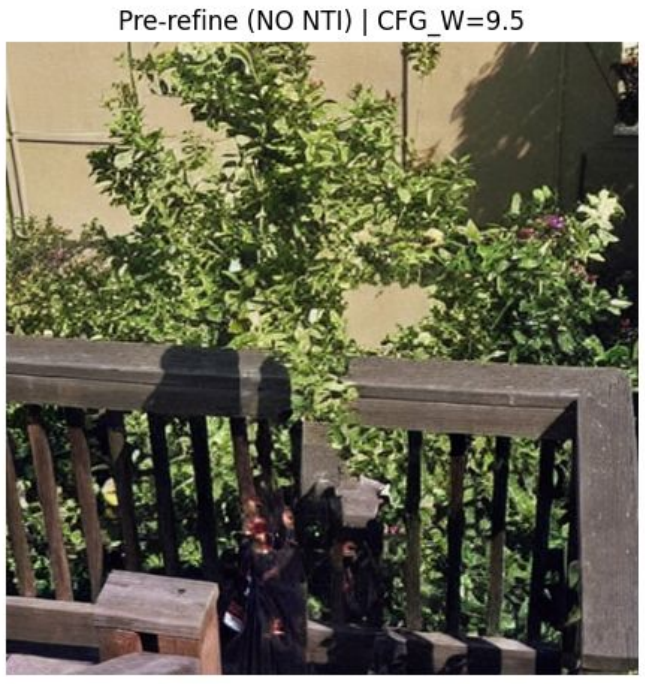

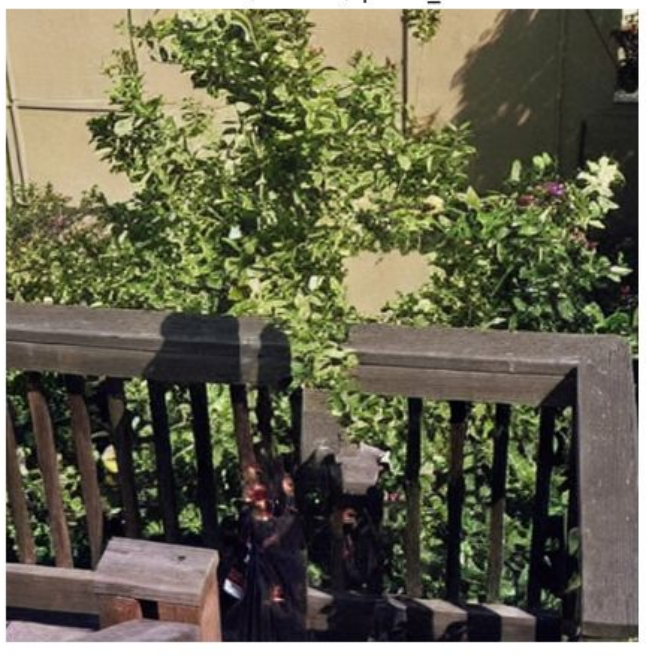

Refined (NO NTI) | R=10 | TREF_FRAC=0.5 | CFG_REF=9.5

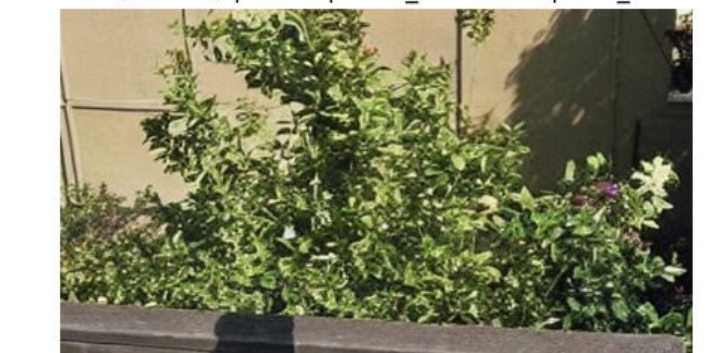

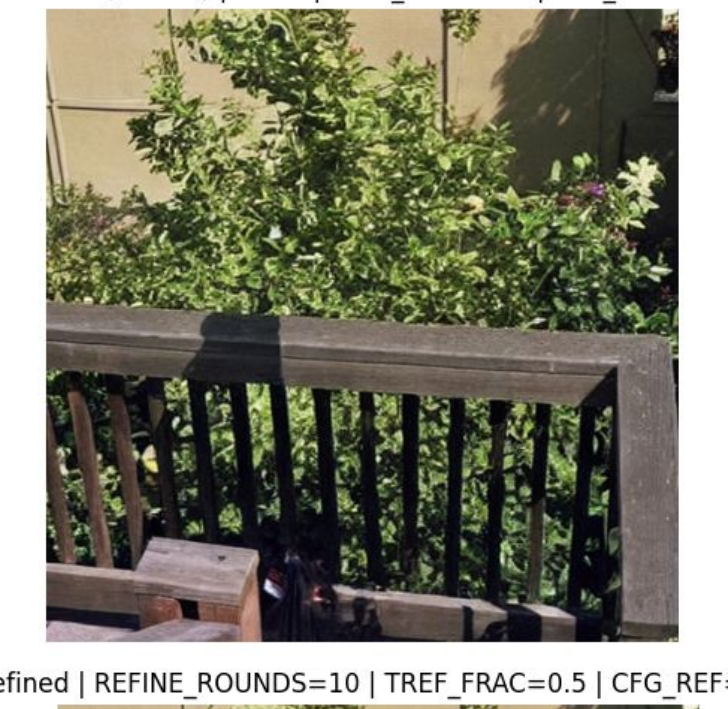

=================== FINAL OUTPUT EVAL (REFINED NO-NTI) ============
FINAL OUTPUT EVAL (REFINED NO-NTI)
mask: area=0.2105 | band_px=12 | ring_px=20 | ring_seam_px=8

Boundary seam (grad): in=0.6056 out=0.5798
abs_mismatch=0.0258 | rel_mismatch=0.0445 (lower better)
Boundary seam (color jump L2): 0.0252 (lower better)
Neighborhood consistency (outside rings): rgb_l2=0.1040 | grad_abs=0.0925 (lower better)
BG alignment (patch vs context): sim=0.7564 (CLIP) (higher better)
Local feature distance (patch vs context): d=0.9675 (ResNet50) (lower better)
SSIM_ring (orig vs ring_mix): 0.9431 (higher better)
LPIPS_ring (orig vs ring_mix): 0.0328 (lower better)

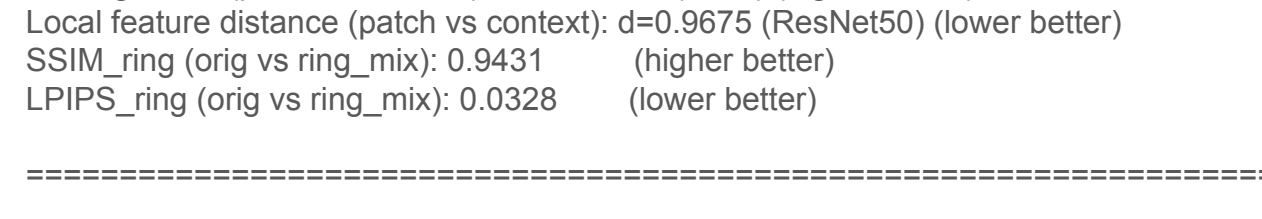

================================================================

Pre-refine | CFG_W=9.5

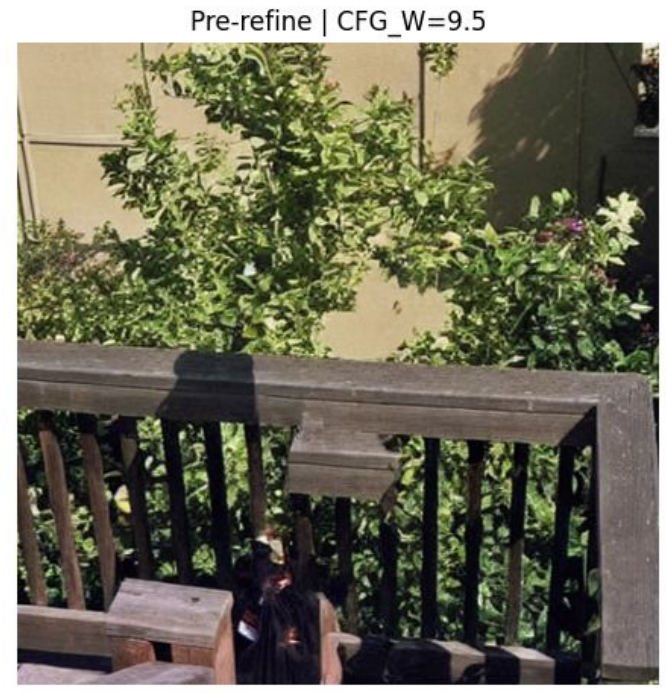

Refined | REFINE_ROUNDS=10 | TREF_FRAC=0.5 | CFG_REF=9.5

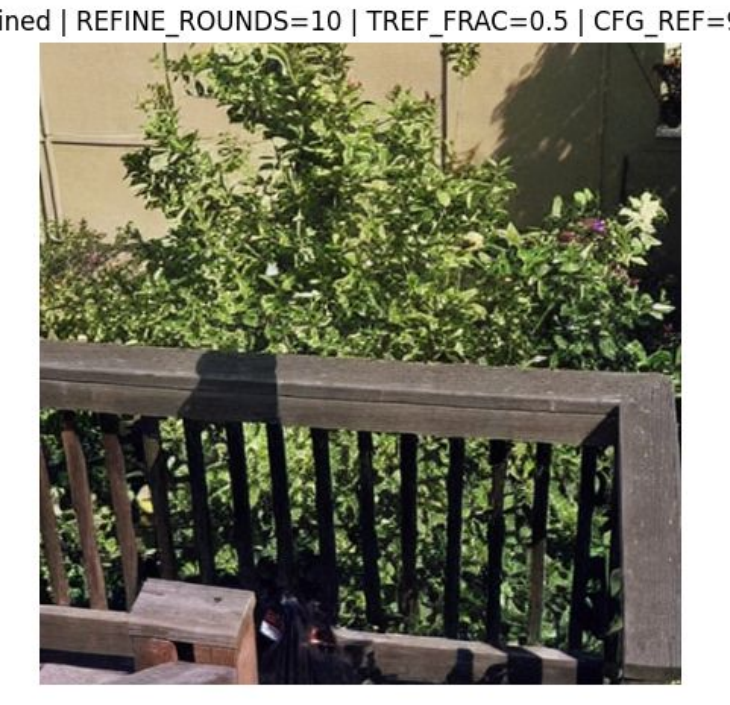

=================== FINAL OUTPUT EVAL (REFINED) ===================
FINAL OUTPUT EVAL (REFINED)
mask: area=0.2105 | band_px=12 | ring_px=20 | ring_seam_px=8

Boundary seam (grad): in=0.6061 out=0.5800
abs_mismatch=0.0261 | rel_mismatch=0.0450 (lower better)
Boundary seam (color jump L2): 0.0270 (lower better)
Neighborhood consistency (outside rings): rgb_l2=0.1078 | grad_abs=0.0926 (lower better)
BG alignment (patch vs context): sim=0.7555 (CLIP) (higher better)
Local feature distance (patch vs context): d=0.9637 (ResNet50) (lower better)
SSIM_ring (orig vs ring_mix): 0.9435 (higher better)
LPIPS_ring (orig vs ring_mix): 0.0324 (lower better)

=================================================================

The quantitative results are very similar across the two variants. The no-NTI result achieves slightly lower boundary-gradient mismatch, color jump, outside-ring RGB difference, and slightly better CLIP-based background alignment. In contrast, the masked-NTI result obtains marginally better local feature distance, ring-focused SSIM, and ring-focused LPIPS. These differences are minor and support the qualitative observation that masked NTI provides no significant advantage for this example.

Target x0 (inv_xt[-1])

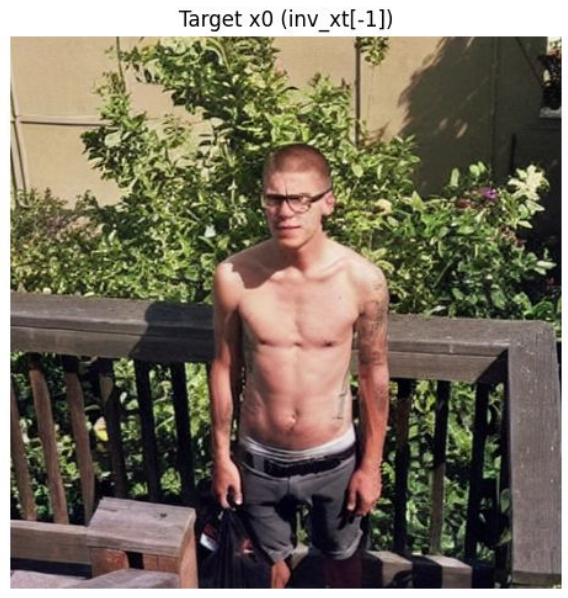

Recon FREE x0 (NO reinjection)

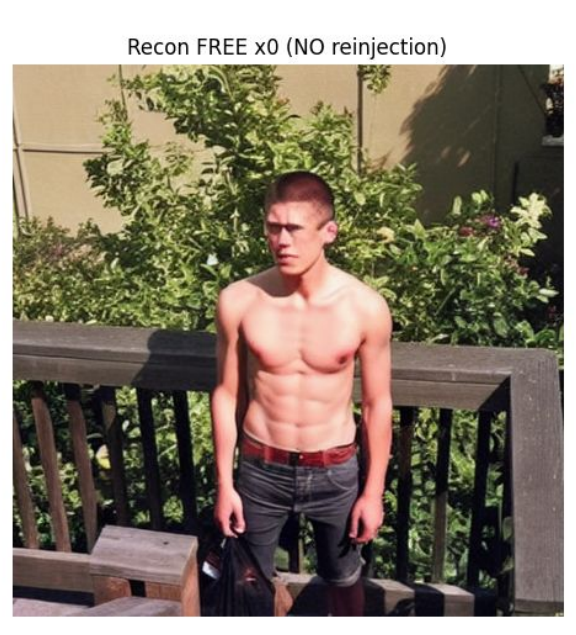

reconstruction with masked NTI

$\lambda_{in} \ll 1$

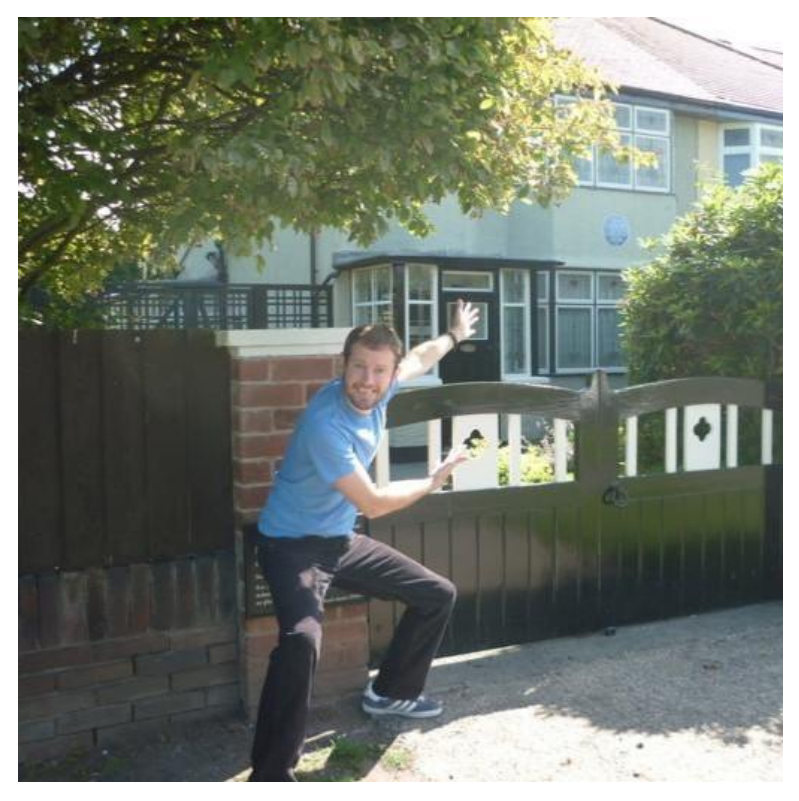

In this example, the objective is to remove the man from the scene. In the no-NTI variant, decoder self-attention masking successfully removes the target object, but visible artifacts remain in the removed region. Refinement reduces these artifacts, although inconsistencies persist in the wall structure behind the removed object. The NTI-based variant shows a similar behavior: refinement improves the output, but noticeable inconsistencies remain in the wall and around the entrance door frame. This example can therefore be considered a failure case, likely due to the structural complexity of the background, which makes coherent reconstruction of the missing region difficult.

Largest CC (may have holes)

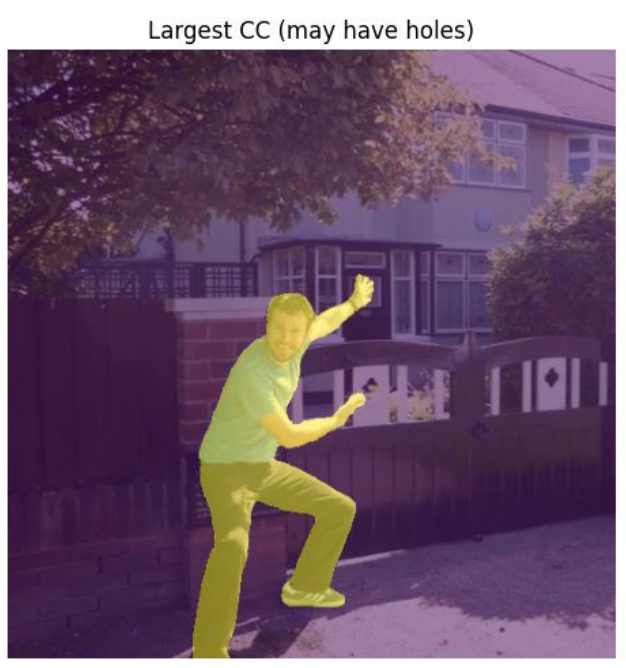

After hole fill (no internal gaps)

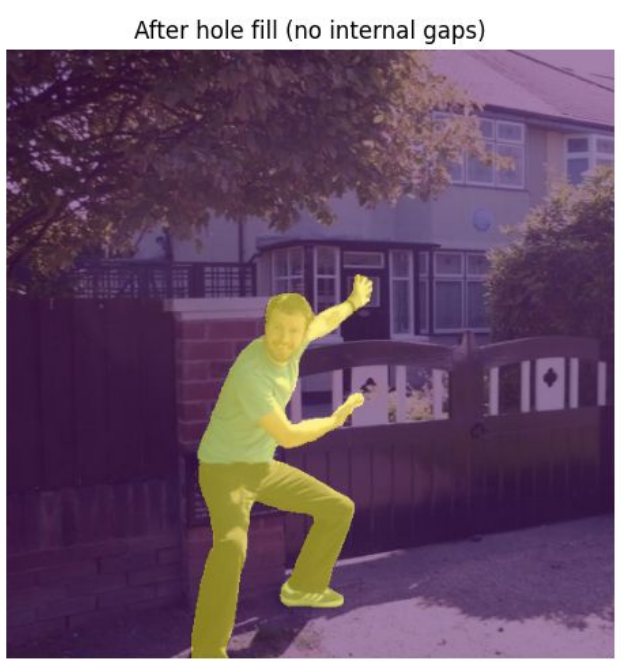

Filled + dilated (final hard mask)

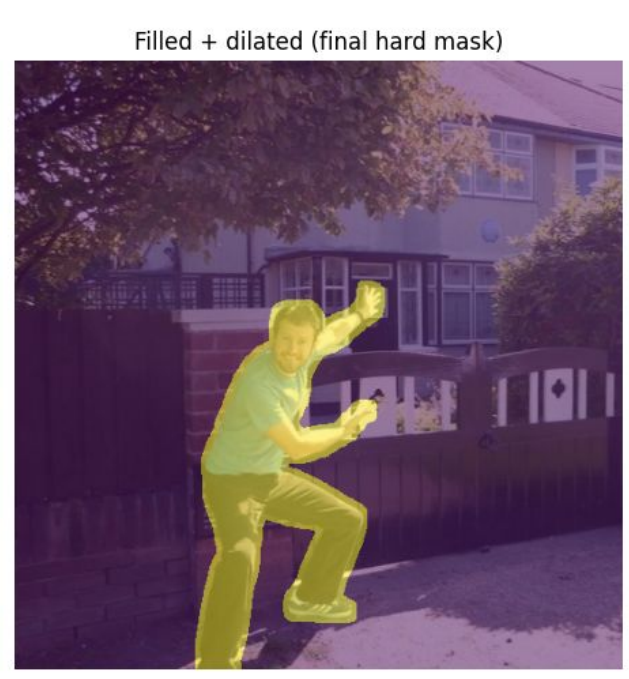

Pre-refine (NO NTI) | CFG_W=9.5

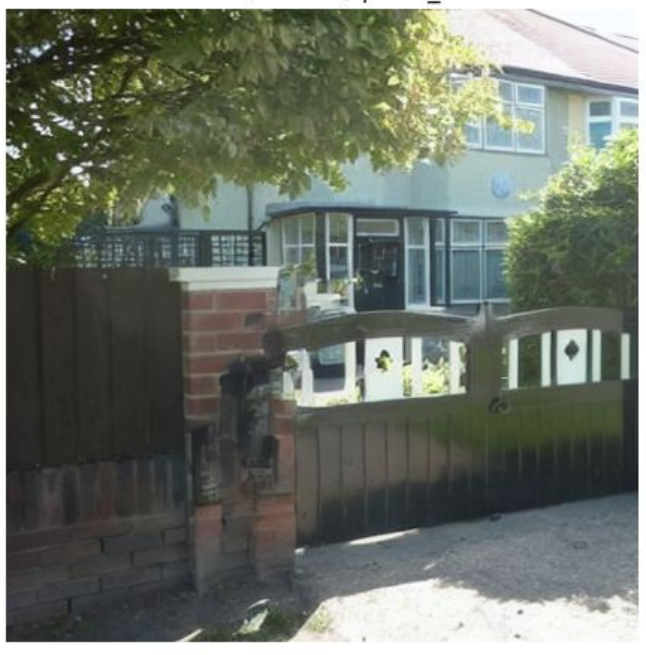

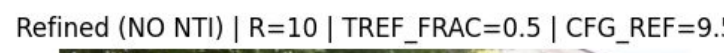

Refined (NO NTI) | R=10 | TREF_FRAC=0.5 | CFG_REF=9.5

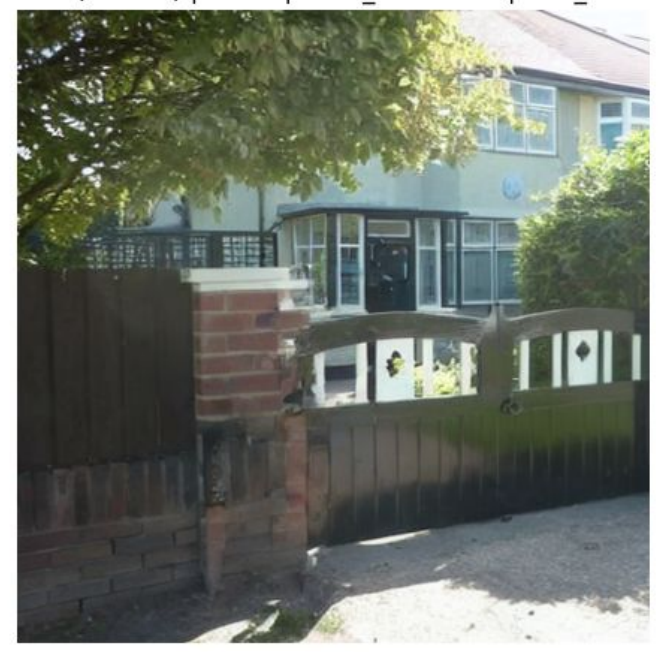

```
=================== FINAL OUTPUT EVAL (REFINED NO-NTI) ============
FINAL OUTPUT EVAL (REFINED NO-NTI)
mask: area=0.1040 | band_px=12 | ring_px=20 | ring_seam_px=8

Boundary seam (grad): in=0.2505 out=0.2677
abs_mismatch=0.0172 | rel_mismatch=0.0642  (lower better)
Boundary seam (color jump L2): 0.0387           (lower better)
Neighborhood consistency (outside rings): rgb_l2=0.0754 | grad_abs=0.0229 (lower better)
BG alignment (patch vs context): sim=0.7140 (CLIP) (higher better)
Local feature distance (patch vs context): d=1.1783 (ResNet50) (lower better)
SSIM_ring (orig vs ring_mix): 0.9491          (higher better)
LPIPS_ring (orig vs ring_mix): 0.0468        (lower better)

=================================================================
```

Pre-refine | CFG_W=9.5

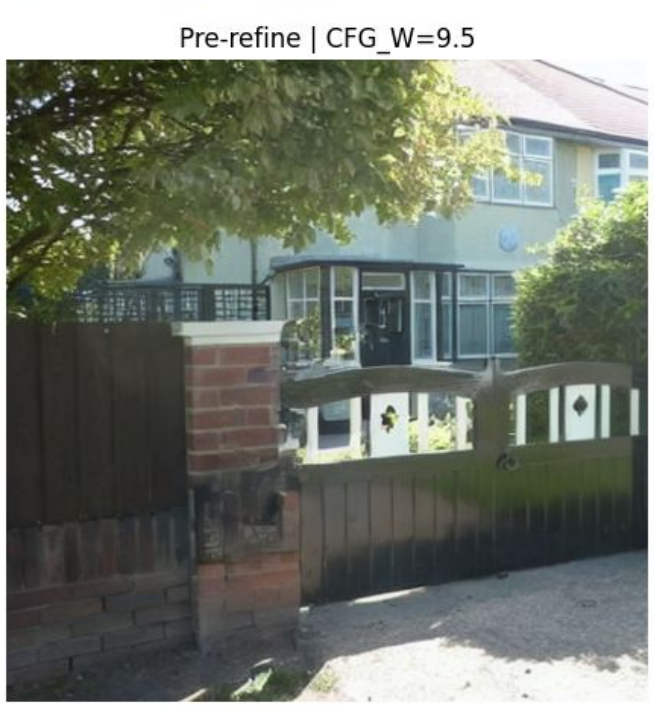

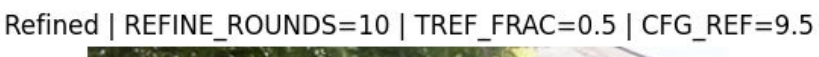

Refined | REFINE_ROUNDS=10 | TREF_FRAC=0.5 | CFG_REF=9.5

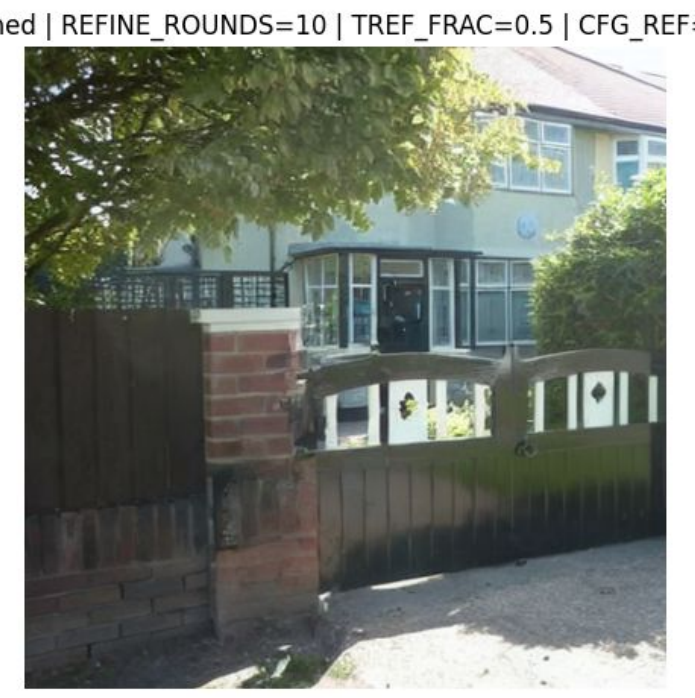

```
=================== FINAL OUTPUT EVAL (REFINED) ===================
FINAL OUTPUT EVAL (REFINED)
mask: area=0.1040 | band_px=12 | ring_px=20 | ring_seam_px=8

Boundary seam (grad): in=0.2506 out=0.2641
abs_mismatch=0.0135 | rel_mismatch=0.0512  (lower better)
Boundary seam (color jump L2): 0.0375           (lower better)
Neighborhood consistency (outside rings): rgb_l2=0.0762 | grad_abs=0.0205 (lower better)
BG alignment (patch vs context): sim=0.6902 (CLIP) (higher better)
Local feature distance (patch vs context): d=1.1705 (ResNet50) (lower better)
SSIM_ring (orig vs ring_mix): 0.9492          (higher better)
LPIPS_ring (orig vs ring_mix): 0.0477        (lower better)

=================================================================
```

The quantitative results show only small and mixed differences between the two variants. The masked-NTI result achieves lower boundary-gradient mismatch, color jump, outside-ring gradient difference, and local feature distance, together with a marginally higher ring-focused SSIM. In contrast, the no-NTI result obtains slightly better outside-ring RGB consistency, CLIP-based background alignment, and ring-focused LPIPS. However, these local metrics do not fully capture the visible structural errors in the reconstructed wall and door frame, emphasizing the need to interpret them together with qualitative inspection.

Target x0 (inv_xt[-1])

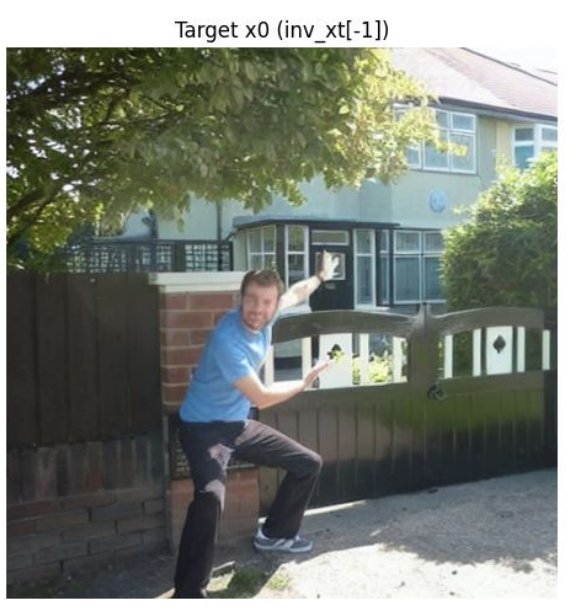

Recon FREE x0 (NO reinjection)

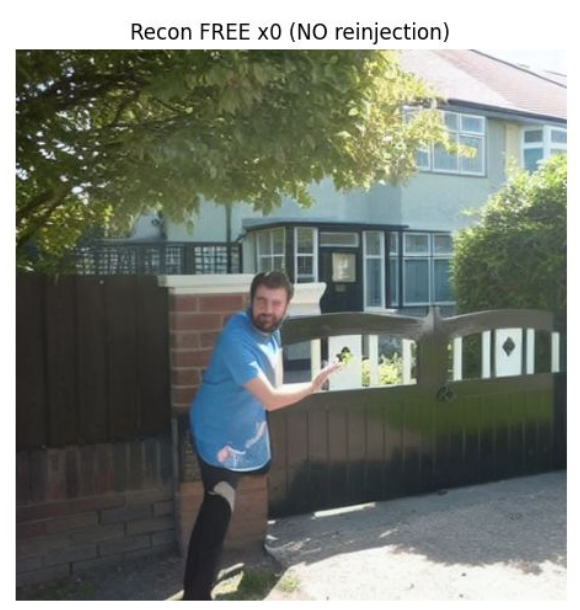

reconstruction with masked NTI

$\lambda_{in} \ll 1$

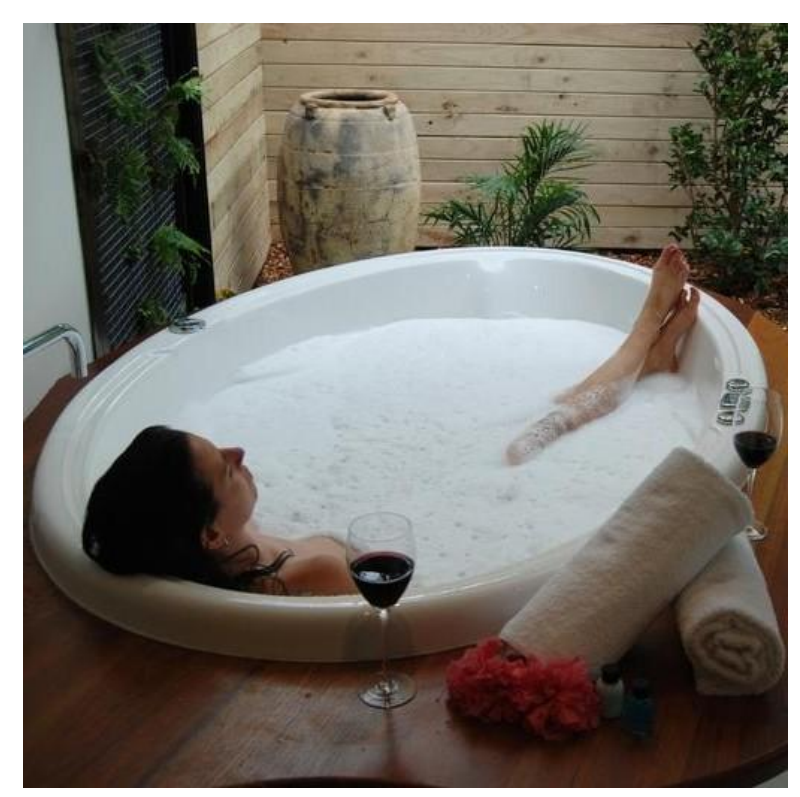

In this example, the objective is to remove the woman from the bathtub. In the no-NTI variant, decoder self-attention masking successfully removes the target object, but substantial artifacts remain in the masked region. Refinement removes most of these artifacts and improves the overall visual quality, although it introduces a new inconsistency in the form of an additional wine glass. In contrast, the refined NTI-based variant produces a cleaner and more coherent final result, suggesting that NTI improves the compatibility between the generated content and the surrounding scene. Further improvements may be possible with a stronger refinement configuration and more masked-NTI inner iterations.

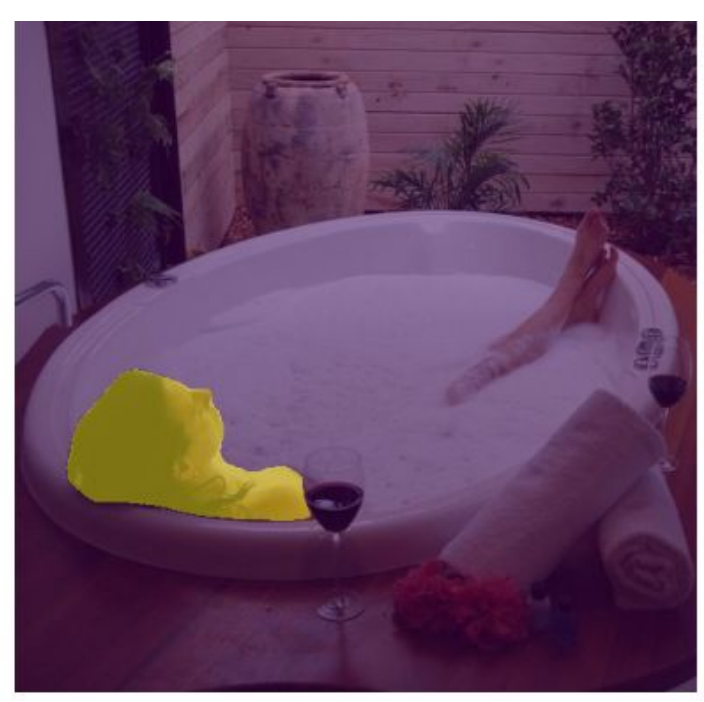

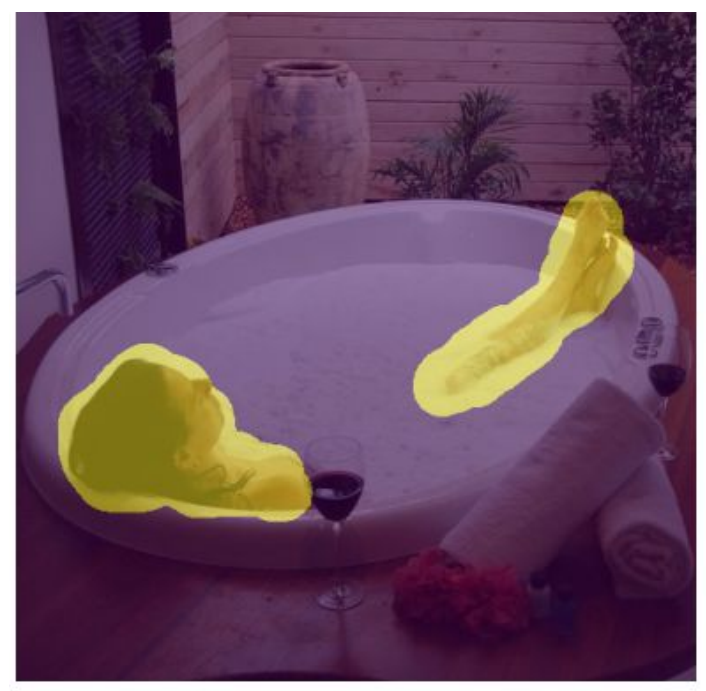

Pre-refine (NO NTI) | CFG_W=9.5

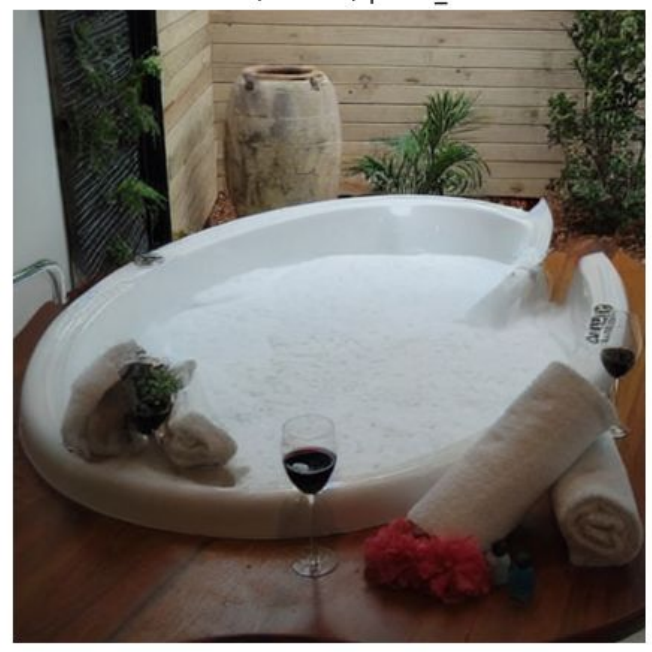

Refined (NO NTI) | R=10 | TREF_FRAC=0.5 | CFG_REF=9.5

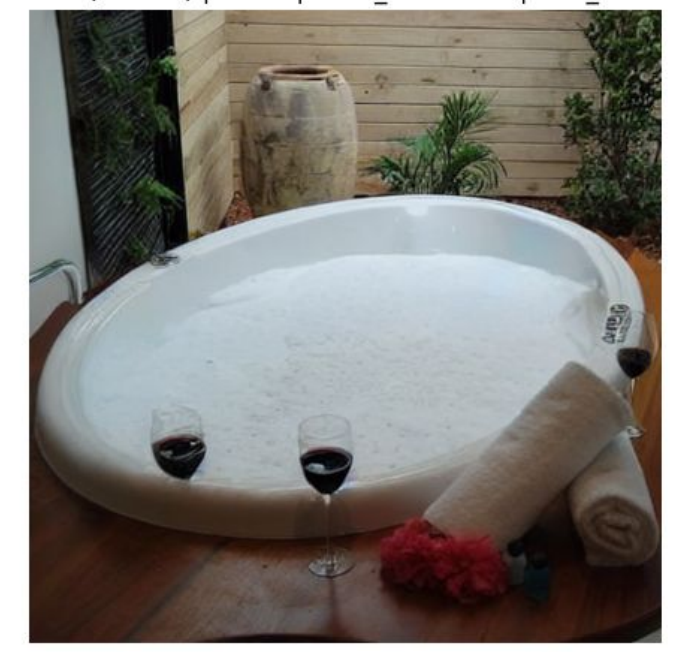

==================== FINAL OUTPUT EVAL (REFINED NO-NTI) ============
FINAL OUTPUT EVAL (REFINED NO-NTI)
mask: area=0.1084 | band_px=12 | ring_px=20 | ring_seam_px=8

Boundary seam (grad): in=0.1207 out=0.1254
abs_mismatch=0.0047 | rel_mismatch=0.0376 (lower better)
Boundary seam (color jump L2): 0.1334 (lower better)
Neighborhood consistency (outside rings): rgb_l2=0.1324 | grad_abs=0.0484 (lower better)
BG alignment (patch vs context): sim=0.6026 (CLIP) (higher better)
Local feature distance (patch vs context): d=1.1439 (ResNet50) (lower better)
SSIM_ring (orig vs ring_mix): 0.9720 (higher better)
LPIPS_ring (orig vs ring_mix): 0.0312 (lower better)

==================================================================

Pre-refine | CFG_W=9.5

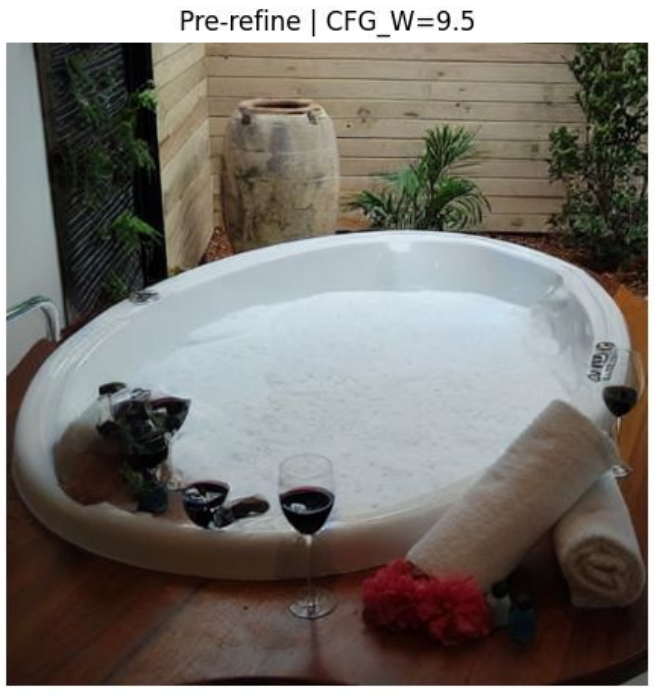

Refined | REFINE_ROUNDS=10 | TREF_FRAC=0.5 | CFG_REF=9.5

==================== FINAL OUTPUT EVAL (REFINED) ====================
FINAL OUTPUT EVAL (REFINED)
mask: area=0.1084 | band_px=12 | ring_px=20 | ring_seam_px=8

Boundary seam (grad): in=0.1137 out=0.1286
abs_mismatch=0.0149 | rel_mismatch=0.1158 (lower better)
Boundary seam (color jump L2): 0.1563 (lower better)
Neighborhood consistency (outside rings): rgb_l2=0.1355 | grad_abs=0.0492 (lower better)
BG alignment (patch vs context): sim=0.6319 (CLIP) (higher better)
Local feature distance (patch vs context): d=1.1617 (ResNet50) (lower better)
SSIM_ring (orig vs ring_mix): 0.9715 (higher better)
LPIPS_ring (orig vs ring_mix): 0.0323 (lower better)

==================================================================

The quantitative results are mixed. The refined no-NTI result achieves lower boundary-gradient mismatch, lower color jump, slightly better outside-ring consistency, lower local feature distance, slightly higher ring-focused SSIM, and lower ring-focused LPIPS. The refined masked-NTI result is better only in CLIP-based background alignment. This indicates that the local metrics do not fully reflect the qualitative difference observed in this example, particularly the semantic inconsistency introduced by the extra wine glass in the no-NTI result. The quantitative measures should therefore be interpreted together with visual inspection.

Target x0 (inv_xt[-1])

Recon FREE x0 (NO reinjection)

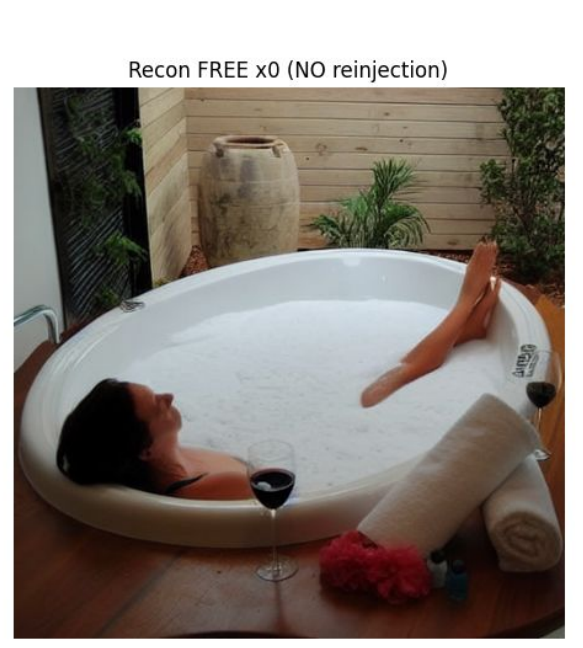

reconstruction with masked NTI

$\lambda_{in} \ll 1$

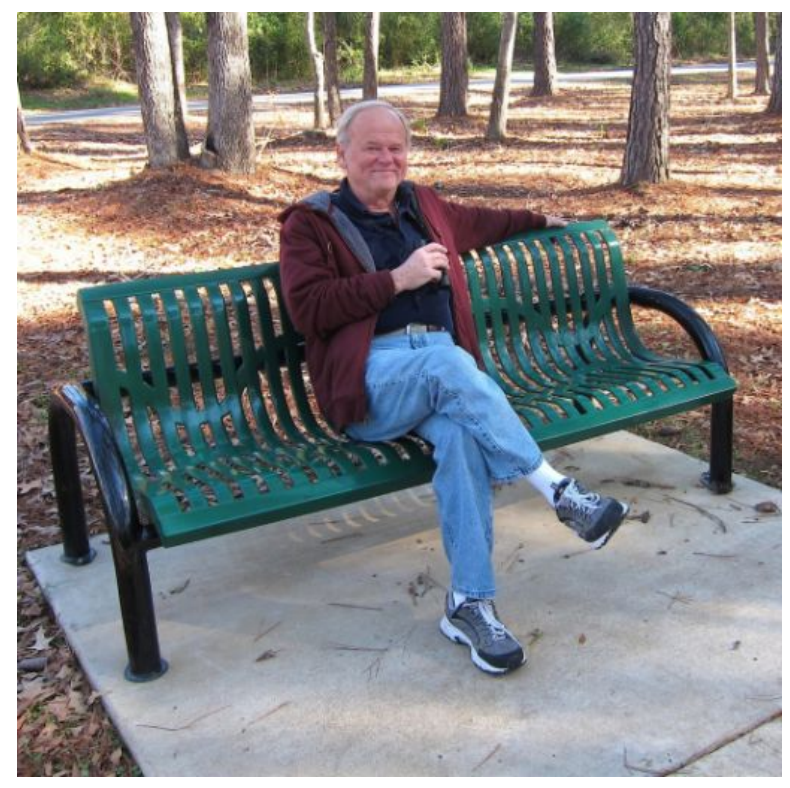

In this example, the objective is to remove the man from the bench. In the no-NTI variant, decoder self-attention masking removes most of the target object, although some artifacts remain before refinement. The refinement stage effectively corrects these artifacts, producing a clean and coherent final output. Compared with the masked-NTI variant, no major qualitative improvement is observed.

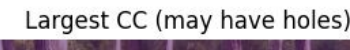

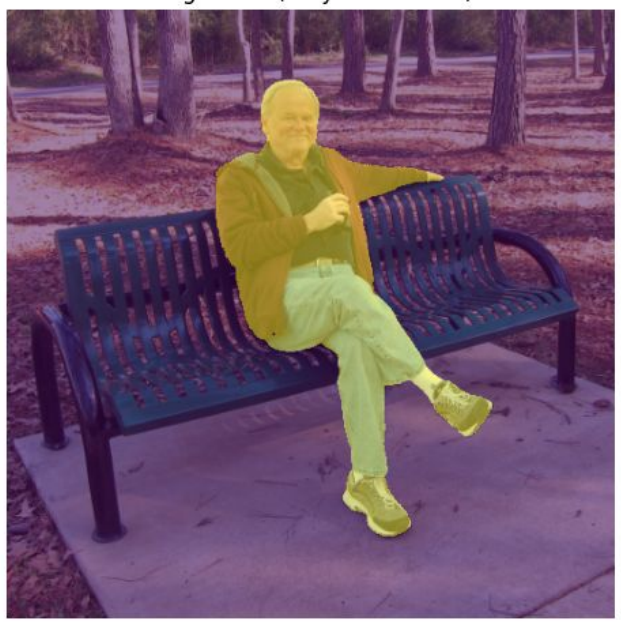

After hole fill (no internal gaps)

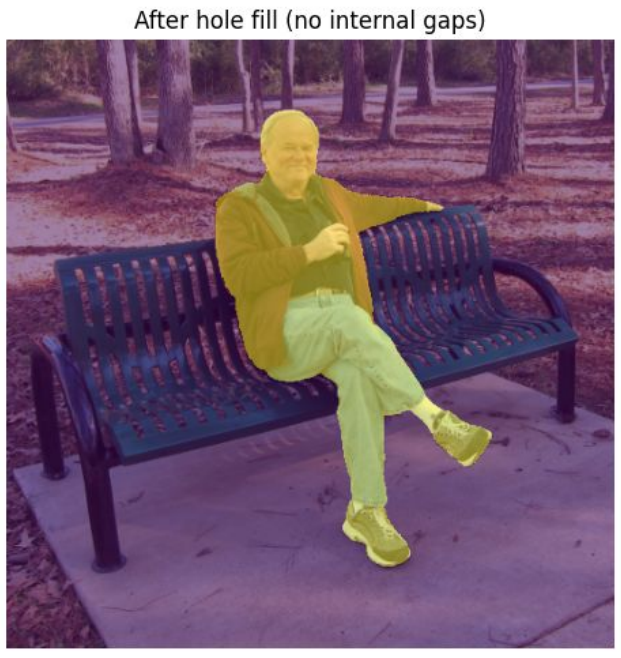

Filled + dilated (final hard mask)

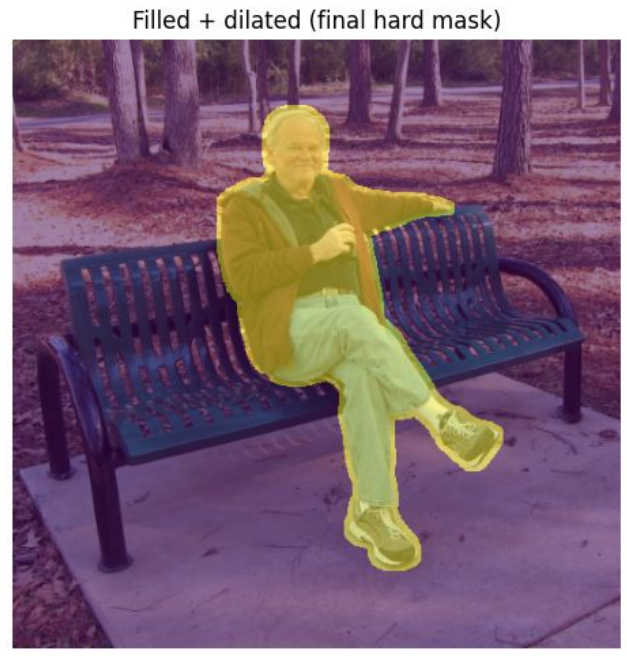

Pre-refine (NO NTI) | CFG_W=9.5

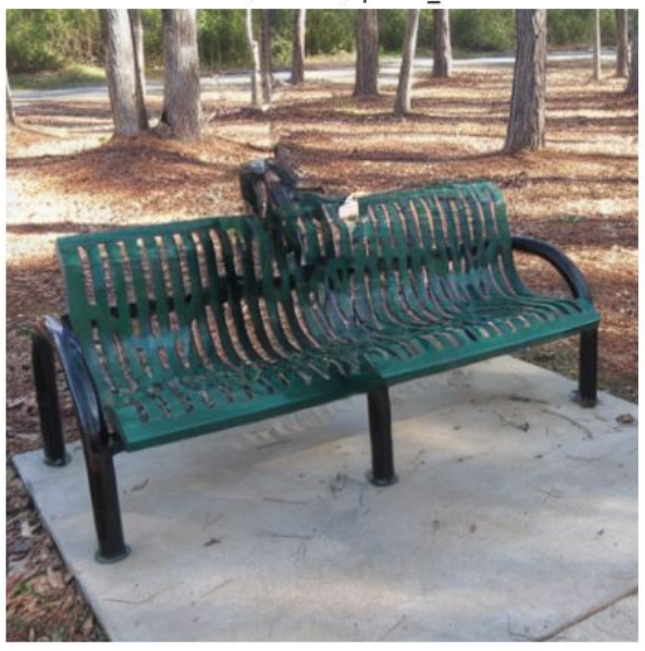

Refined (NO NTI) | R=10 | TREF_FRAC=0.55 | CFG_REF=9.5

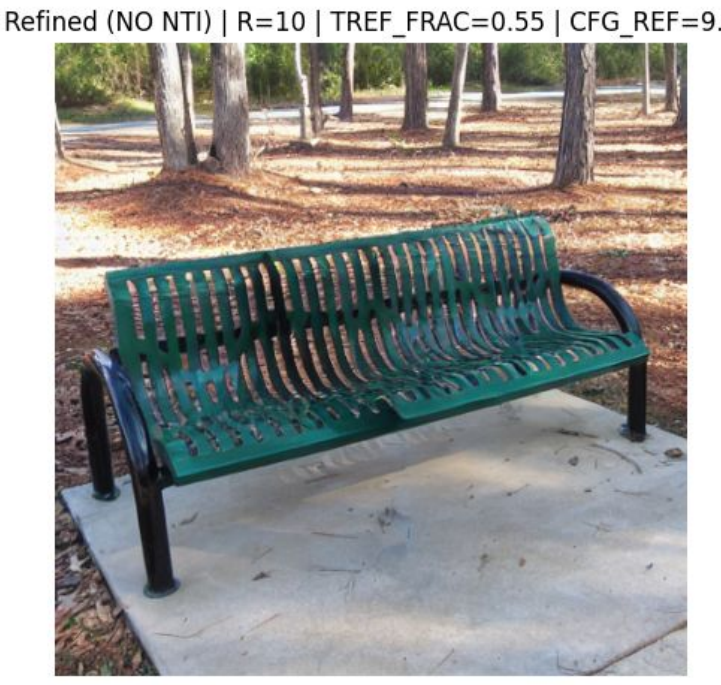

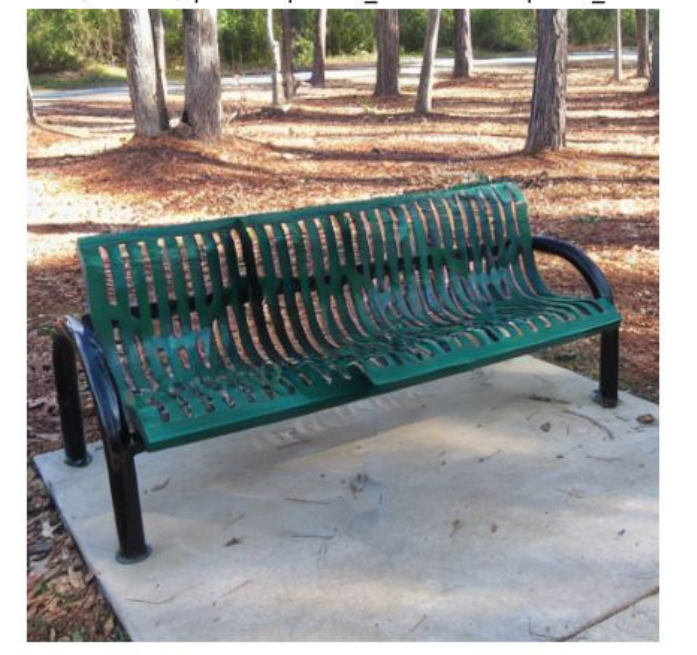

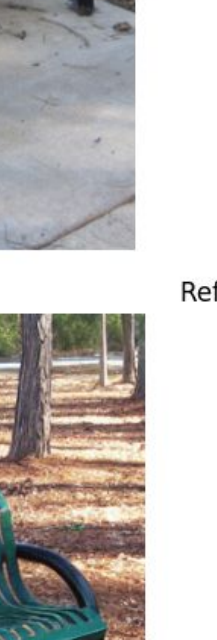

=================== FINAL OUTPUT EVAL (REFINED NO-NTI) ============
FINAL OUTPUT EVAL (REFINED NO-NTI)
mask: area=0.1459 | band_px=12 | ring_px=20 | ring_seam_px=8

Boundary seam (grad): in=0.3538 out=0.3645
abs_mismatch=0.0107 | rel_mismatch=0.0292 (lower better)
Boundary seam (color jump L2): 0.0115 (lower better)
Neighborhood consistency (outside rings): rgb_l2=0.0133 | grad_abs=0.0136 (lower better)
BG alignment (patch vs context): sim=0.6417 (CLIP) (higher better)
Local feature distance (patch vs context): d=1.1097 (ResNet50) (lower better)
SSIM_ring (orig vs ring_mix): 0.9380 (higher better)
LPIPS_ring (orig vs ring_mix): 0.0430 (lower better)

=================================================================

Pre-refine | CFG_W=9.5

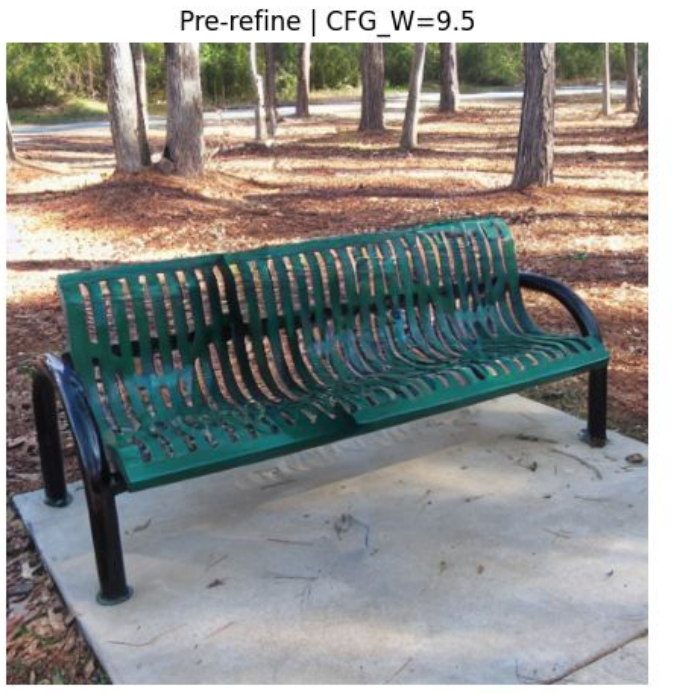

Refined | REFINE_ROUNDS=10 | TREF_FRAC=0.55 | CFG_REF=9.5

=================== FINAL OUTPUT EVAL (REFINED) ===================
FINAL OUTPUT EVAL (REFINED)
mask: area=0.1459 | band_px=12 | ring_px=20 | ring_seam_px=8

Boundary seam (grad): in=0.3551 out=0.3653
abs_mismatch=0.0102 | rel_mismatch=0.0279 (lower better)
Boundary seam (color jump L2): 0.0125 (lower better)
Neighborhood consistency (outside rings): rgb_l2=0.0133 | grad_abs=0.0141 (lower better)
BG alignment (patch vs context): sim=0.6335 (CLIP) (higher better)
Local feature distance (patch vs context): d=1.1107 (ResNet50) (lower better)
SSIM_ring (orig vs ring_mix): 0.9379 (higher better)
LPIPS_ring (orig vs ring_mix): 0.0432 (lower better)

==================================================================

The quantitative results are very similar. The masked-NTI variant achieves slightly lower absolute and relative boundary-gradient mismatch, while the no-NTI result performs marginally better in boundary color jump, outside-ring gradient consistency, CLIP-based background alignment, local feature distance, ring-focused SSIM, and ring-focused LPIPS. The outside-ring RGB difference is identical. These small differences support the qualitative observation that masked NTI provides no clear advantage for this example.

Target x0 (inv_xt[-1])

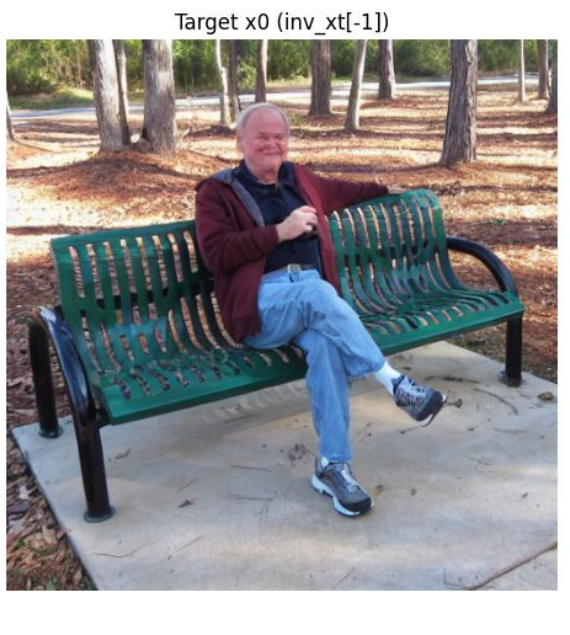

Recon FREE x0 (NO reinjection)

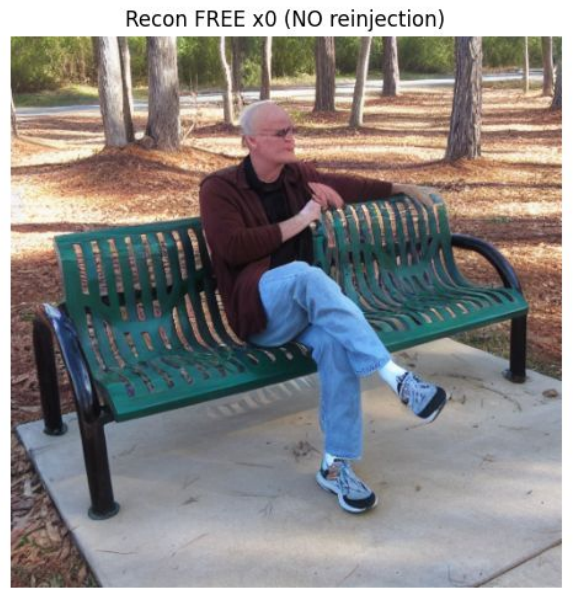

reconstruction with masked NTI

$\lambda_{in} \ll 1$

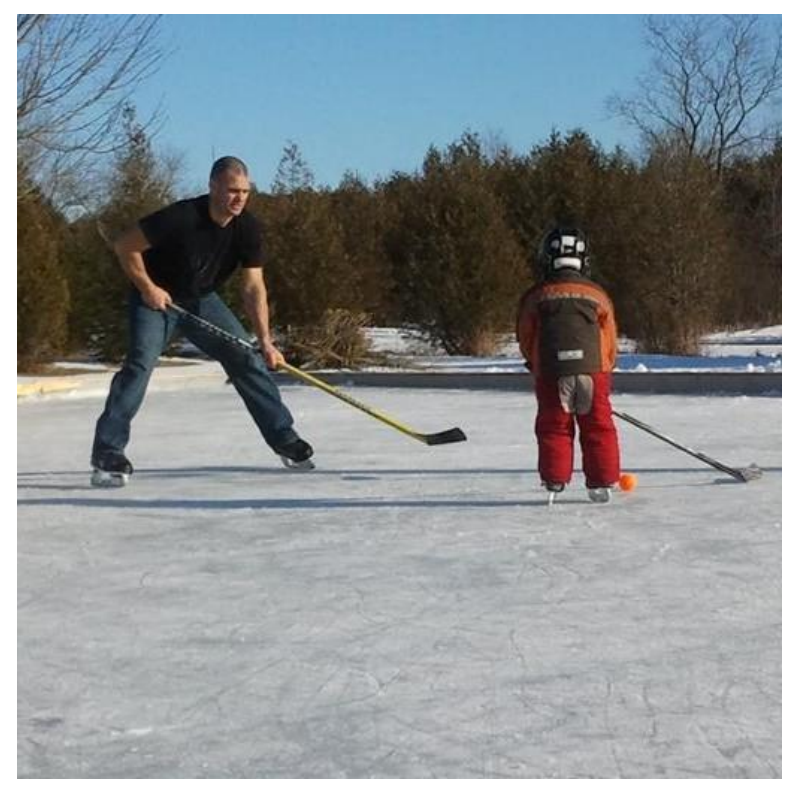

In this example, the objective is to remove the child holding the hockey stick from the image. In the no-NTI variant, decoder self-attention masking partially removes the target object, but some remnants remain before refinement. The refinement stage effectively removes these traces and produces a clean, visually coherent background. Compared with the masked-NTI variant, no substantial qualitative improvement is observed, suggesting that self-attention masking followed by refinement is sufficient in this case and that the additional NTI stage provides no clear advantage.

Largest CC (may have holes)

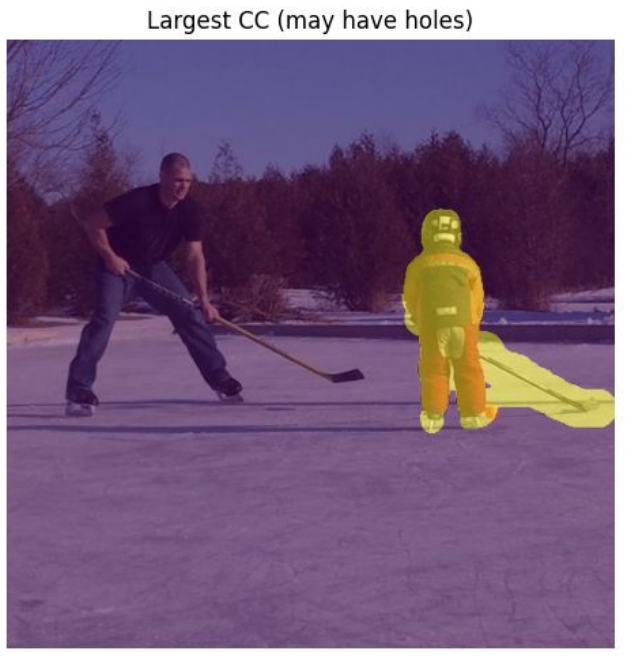

After hole fill (no internal gaps)

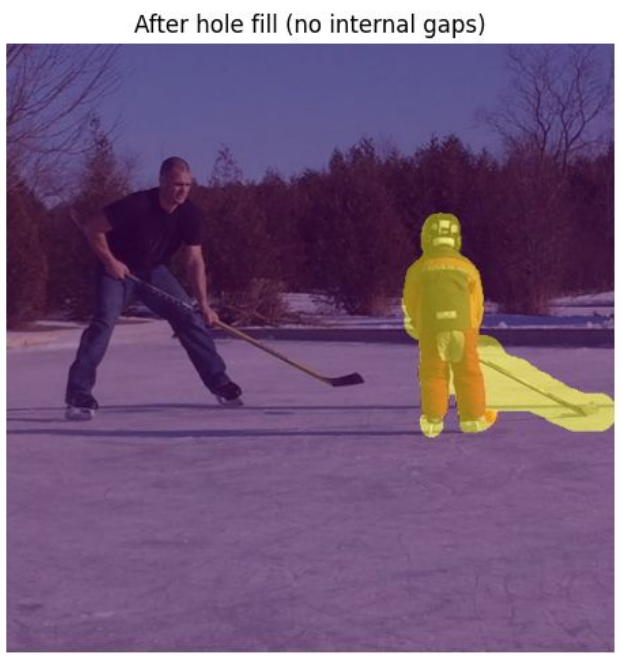

Filled + dilated (final hard mask)

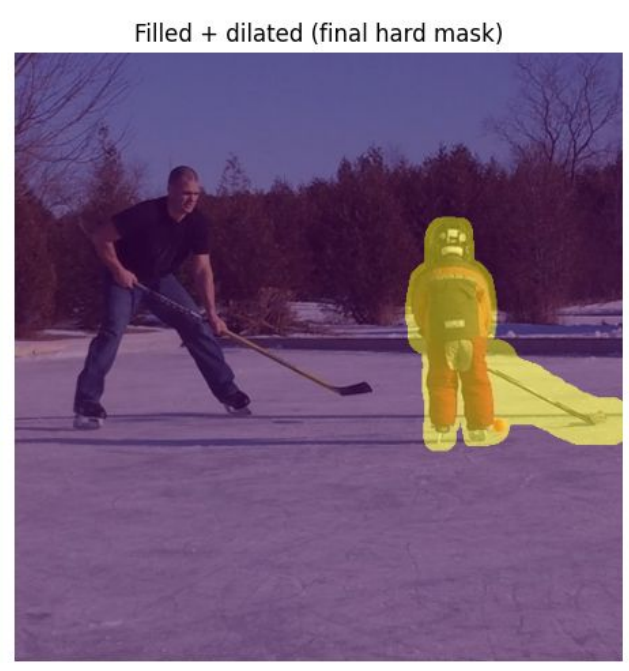

Pre-refine (NO NTI) | CFG_W=9.5

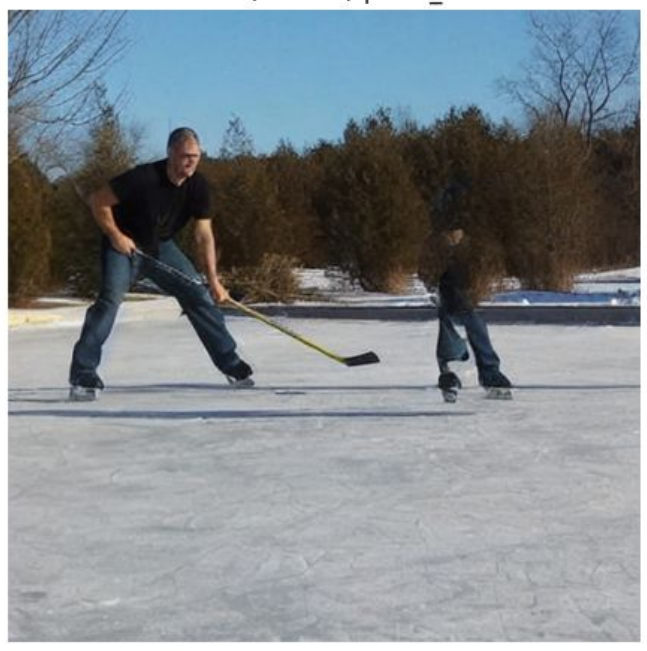

Refined (NO NTI) | R=10 | TREF_FRAC=0.5 | CFG_REF=9.5

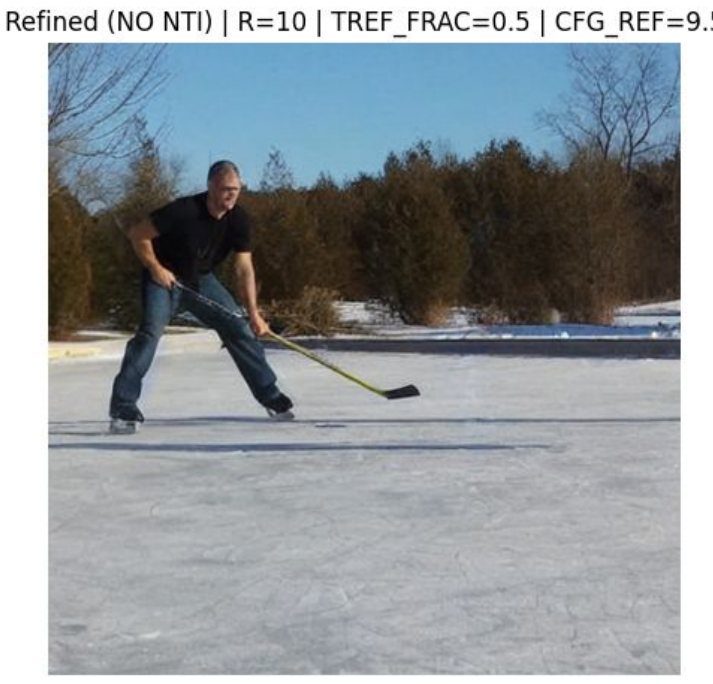

```
=================== FINAL OUTPUT EVAL (REFINED NO-NTI) ============
FINAL OUTPUT EVAL (REFINED NO-NTI)
mask: area=0.0659 | band_px=12 | ring_px=20 | ring_seam_px=8

Boundary seam (grad): in=0.2409 out=0.2103
abs_mismatch=0.0306 | rel_mismatch=0.1457  (lower better)
Boundary seam (color jump L2): 0.0400           (lower better)
Neighborhood consistency (outside rings): rgb_l2=0.0772 | grad_abs=0.0108 (lower better)
BG alignment (patch vs context): sim=0.5168 (CLIP) (higher better)
Local feature distance (patch vs context): d=1.2546 (ResNet50) (lower better)
SSIM_ring (orig vs ring_mix): 0.9780          (higher better)
LPIPS_ring (orig vs ring_mix): 0.0115         (lower better)

=================================================================
```

Pre-refine | CFG_W=9.5

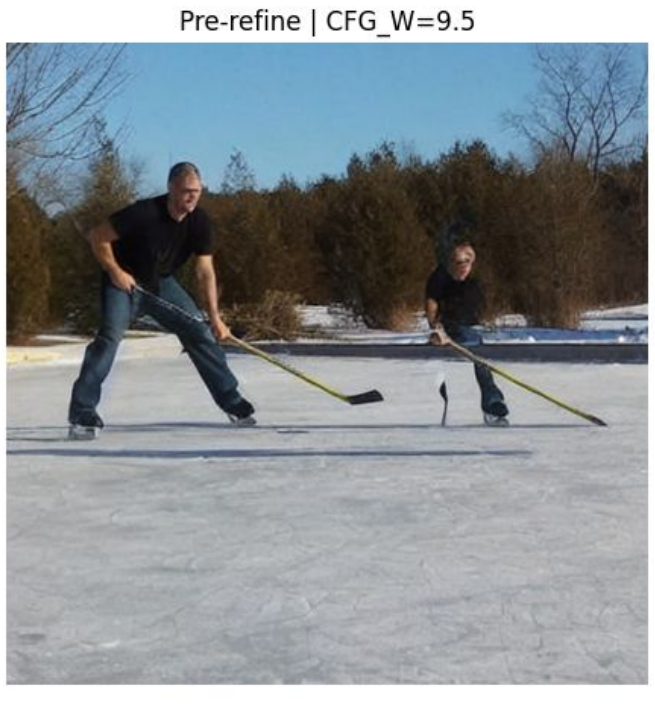

Refined | REFINE_ROUNDS=17 | TREF_FRAC=0.55 | CFG_REF=10.5

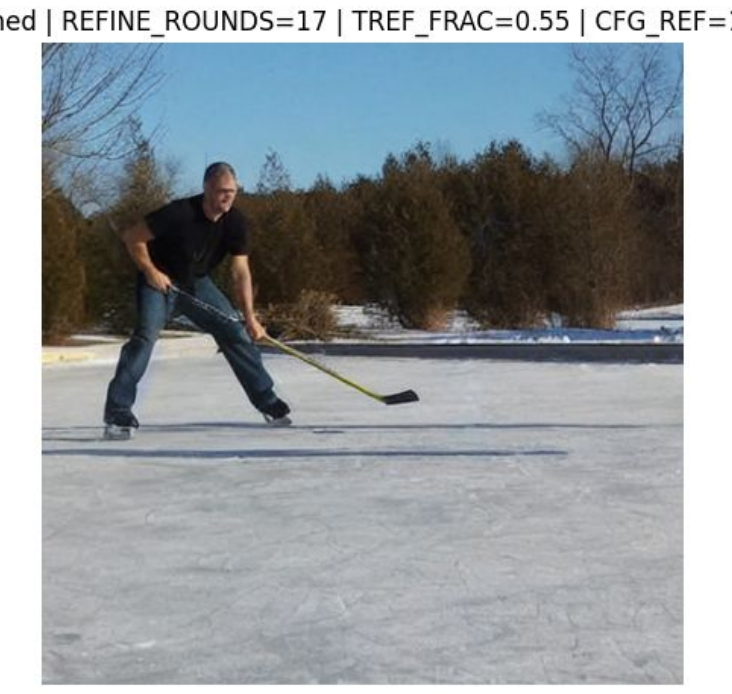

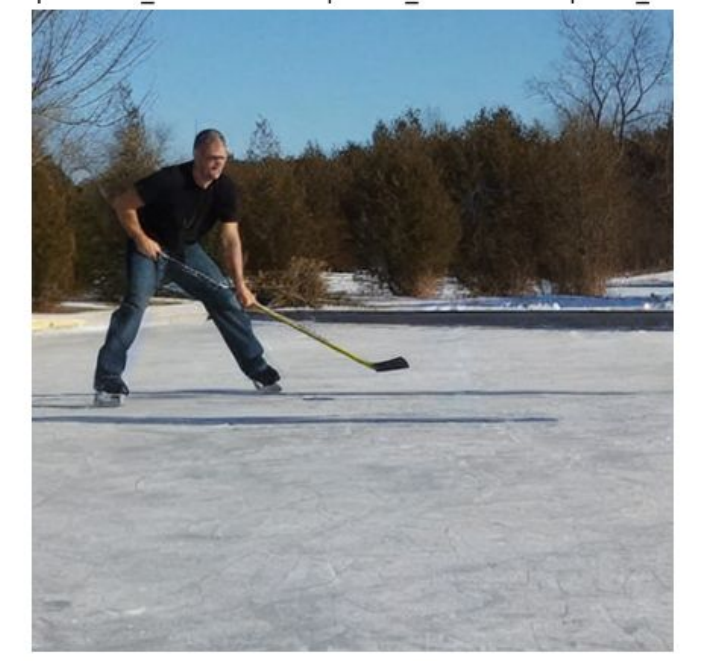

```
=================== FINAL OUTPUT EVAL (REFINED) ===================
FINAL OUTPUT EVAL (REFINED)
mask: area=0.0659 | band_px=12 | ring_px=20 | ring_seam_px=8

Boundary seam (grad): in=0.2481 out=0.2154
abs_mismatch=0.0327 | rel_mismatch=0.1520  (lower better)
Boundary seam (color jump L2): 0.0424           (lower better)
Neighborhood consistency (outside rings): rgb_l2=0.0773 | grad_abs=0.0077 (lower better)
BG alignment (patch vs context): sim=0.5223 (CLIP) (higher better)
Local feature distance (patch vs context): d=1.2375 (ResNet50) (lower better)
SSIM_ring (orig vs ring_mix): 0.9773          (higher better)
LPIPS_ring (orig vs ring_mix): 0.0122         (lower better)

=================================================================
```

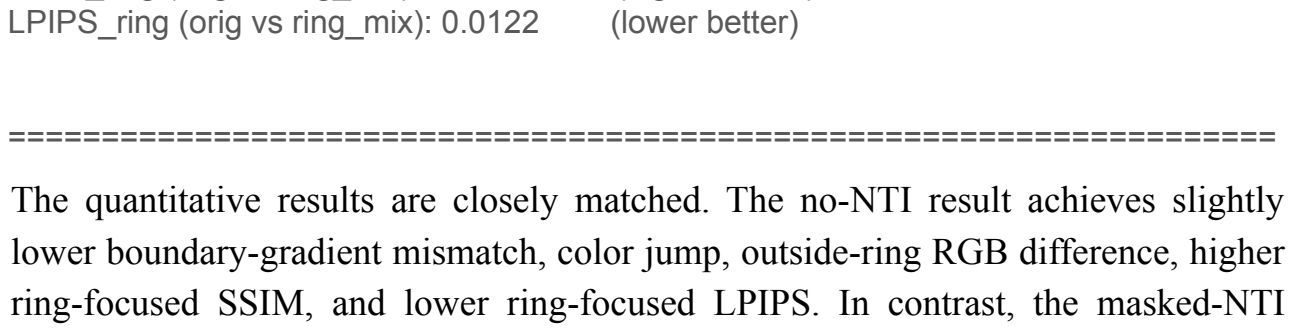

The quantitative results are closely matched. The no-NTI result achieves slightly lower boundary-gradient mismatch, color jump, outside-ring RGB difference, higher ring-focused SSIM, and lower ring-focused LPIPS. In contrast, the masked-NTI result obtains slightly better outside-ring gradient consistency, CLIP-based background alignment, and local feature distance. These small and mixed differences support the qualitative observation that masked NTI provides no significant benefit for this example.

Target x0 (inv_xt[-1])

Recon FREE x0 (NO reinjection)

reconstruction with masked NTI

$\lambda_{in} \ll 1$

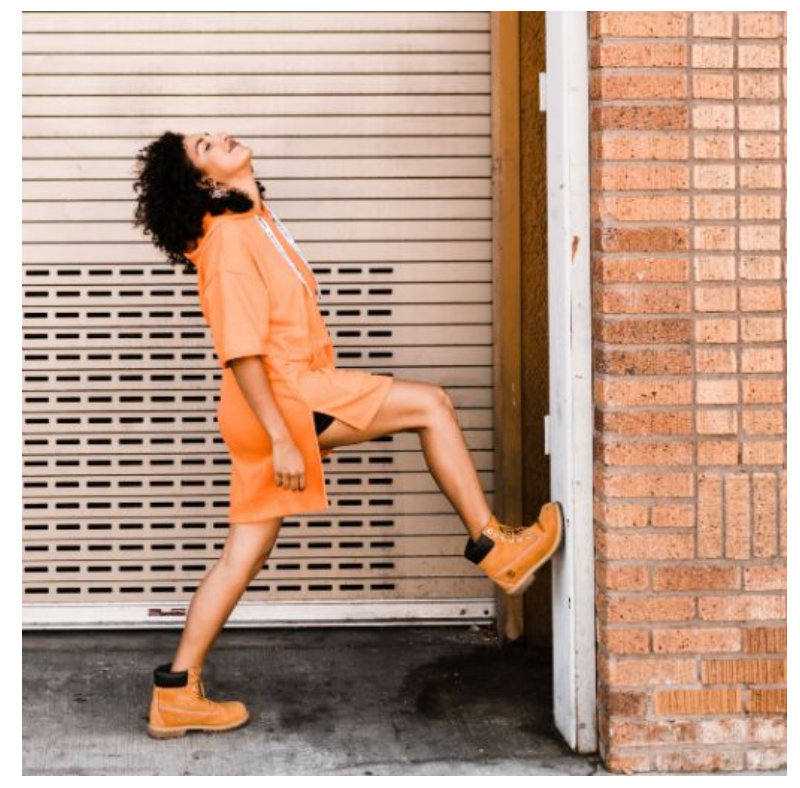

In this example, the objective is to remove the woman from the scene. In the no-NTI variant, decoder self-attention masking removes most of the target object, but visible artifacts remain in the reconstructed background. Refinement reduces these artifacts, although some inconsistencies persist near the woman's shoe. In contrast, the masked-NTI variant resolves this region more effectively. After refinement, the remaining artifacts are removed, producing a clean and visually coherent final result. This example shows that masked-NTI can provide a clear advantage when refinement alone cannot fully eliminate localized artifacts.

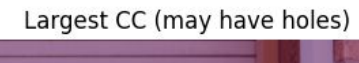

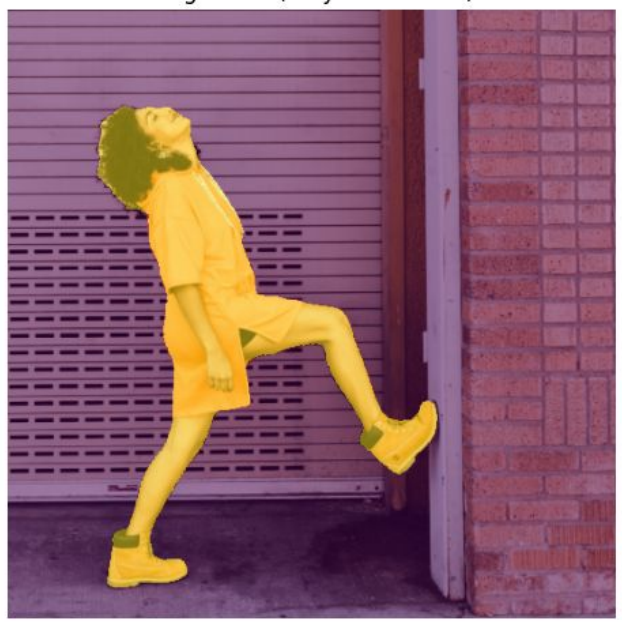

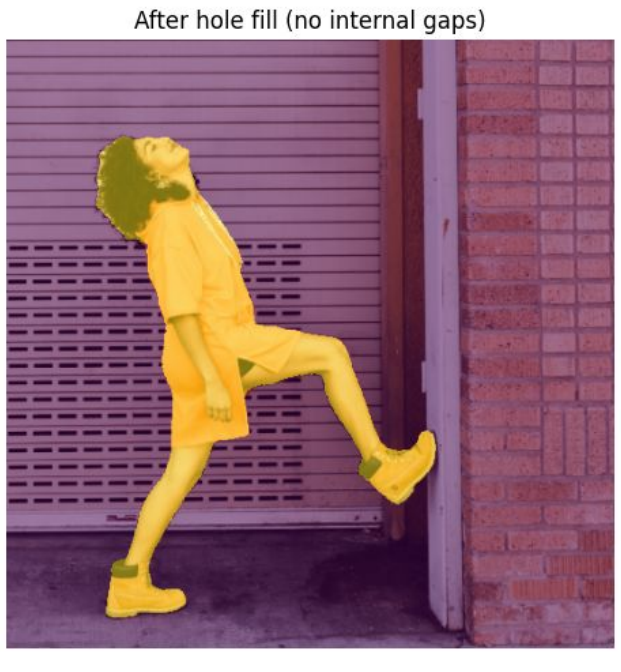

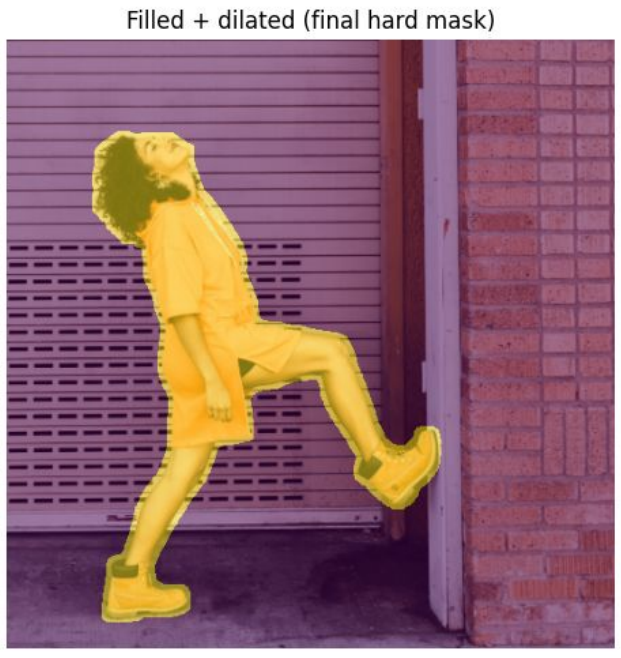

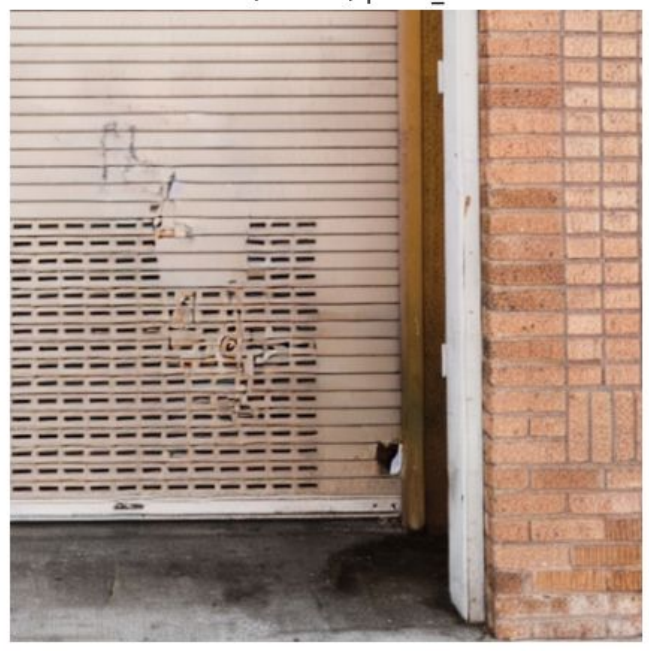

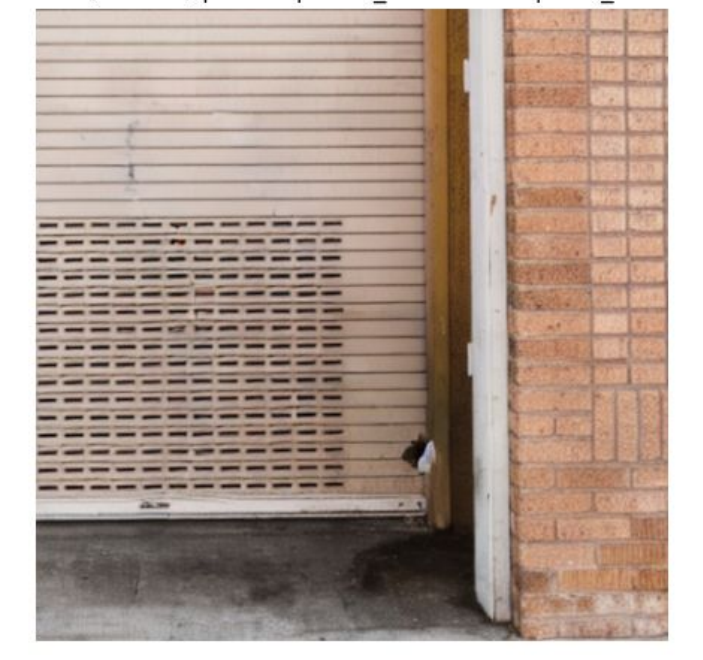

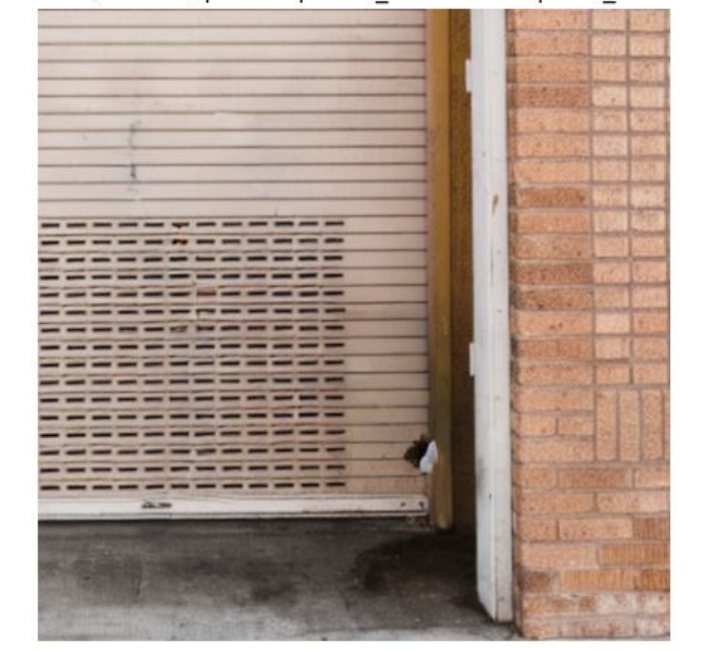

```
=================== FINAL OUTPUT EVAL (REFINED NO-NTI) ============
FINAL OUTPUT EVAL (REFINED NO-NTI)
mask: area=0.1487 | band_px=12 | ring_px=20 | ring_seam_px=8

Boundary seam (grad): in=0.4236 out=0.4086
abs_mismatch=0.0149 | rel_mismatch=0.0365  (lower better)
Boundary seam (color jump L2): 0.0308          (lower better)
Neighborhood consistency (outside rings): rgb_l2=0.0273 | grad_abs=0.0136 (lower better)
BG alignment (patch vs context): sim=0.7493 (CLIP) (higher better)
Local feature distance (patch vs context): d=1.0199 (ResNet50) (lower better)
SSIM_ring (orig vs ring_mix): 0.9402         (higher better)
LPIPS_ring (orig vs ring_mix): 0.0501        (lower better)

=================================================================
```

Pre-refine | CFG_W=9.5

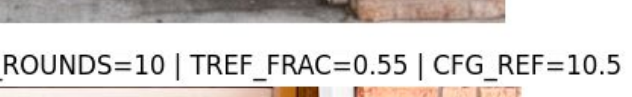

```
=================== FINAL OUTPUT EVAL (REFINED) ===================
FINAL OUTPUT EVAL (REFINED)
mask: area=0.1487 | band_px=12 | ring_px=20 | ring_seam_px=8

Boundary seam (grad): in=0.7861 out=0.7680
abs_mismatch=0.0181 | rel_mismatch=0.0235  (lower better)
Boundary seam (color jump L2): 0.0390          (lower better)
Neighborhood consistency (outside rings): rgb_l2=0.0389 | grad_abs=0.0269 (lower better)
BG alignment (patch vs context): sim=0.7436 (CLIP) (higher better)
Local feature distance (patch vs context): d=0.9922 (ResNet50) (lower better)
SSIM_ring (orig vs ring_mix): 0.9318         (higher better)
LPIPS_ring (orig vs ring_mix): 0.0792        (lower better)

=================================================================
```

The quantitative results are mixed. The masked-NTI result achieves a lower relative boundary-gradient mismatch and a lower local feature distance, which is consistent with the improved qualitative appearance in the edited region. At the same time, the no-NTI variant performs better on several other metrics, including absolute boundary-gradient mismatch, boundary color jump, outside-ring consistency, CLIP-based background alignment, SSIM, and LPIPS. Thus, for this example, the qualitative comparison indicates a clearer benefit of masked NTI than the aggregate local metrics alone, suggesting that the main improvement is concentrated in a specific problematic region rather than reflected uniformly across all evaluation measures.

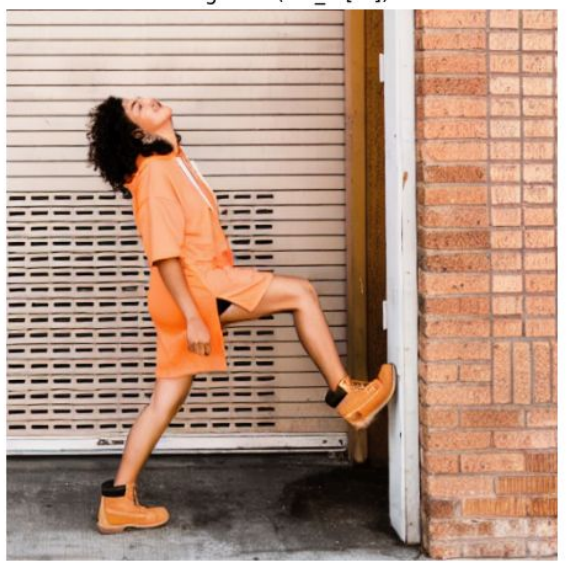

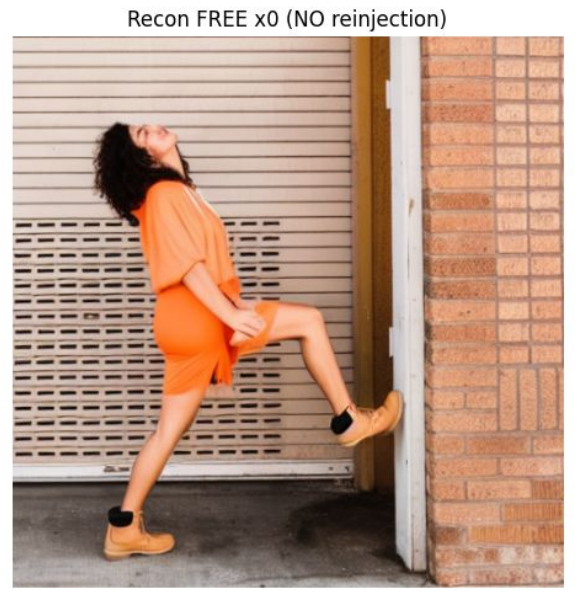

reconstruction with masked NTI

$\lambda_{in} \ll 1$

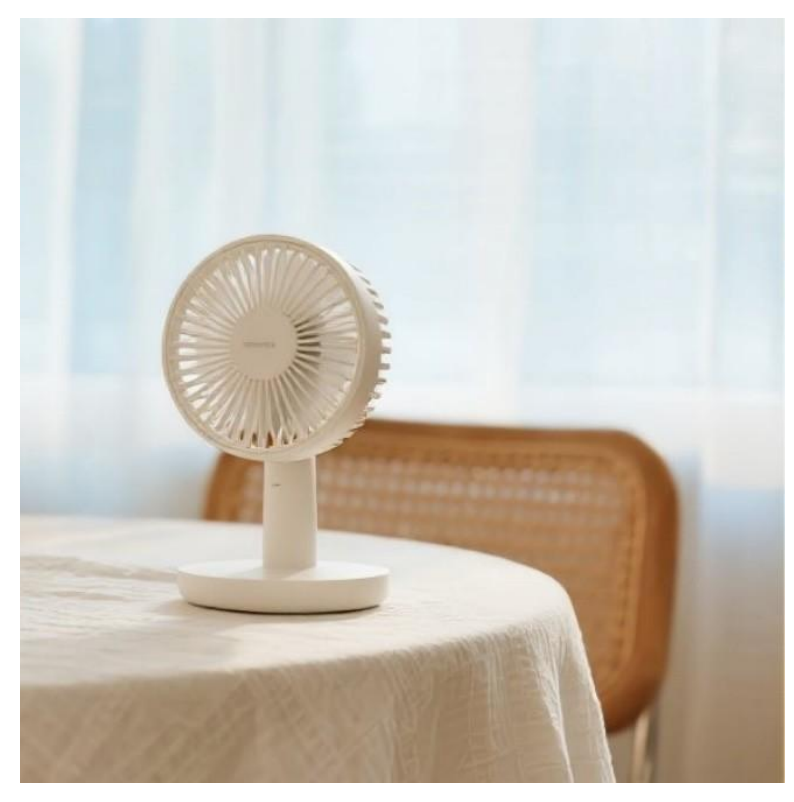

In this example, the objective is to remove the small object from the table. Decoder self-attention masking removes the target object successfully, with no noticeable artifacts remaining in the edited region. As a result, neither the refinement stage nor the masked-NTI variant provides a visible improvement. This example therefore represents a simple successful case in which decoder self-attention masking alone is sufficient to produce a clean object-removal result.

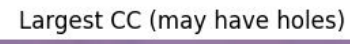
Largest CC (may have holes)

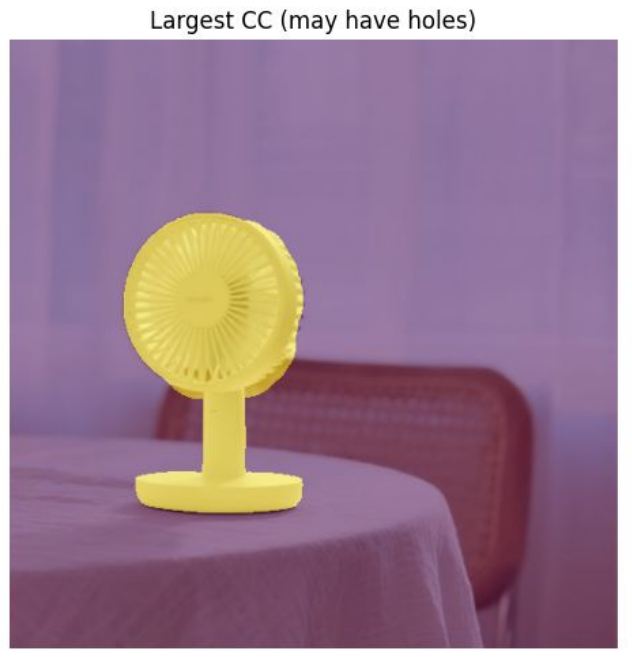

After hole fill (no internal gaps)

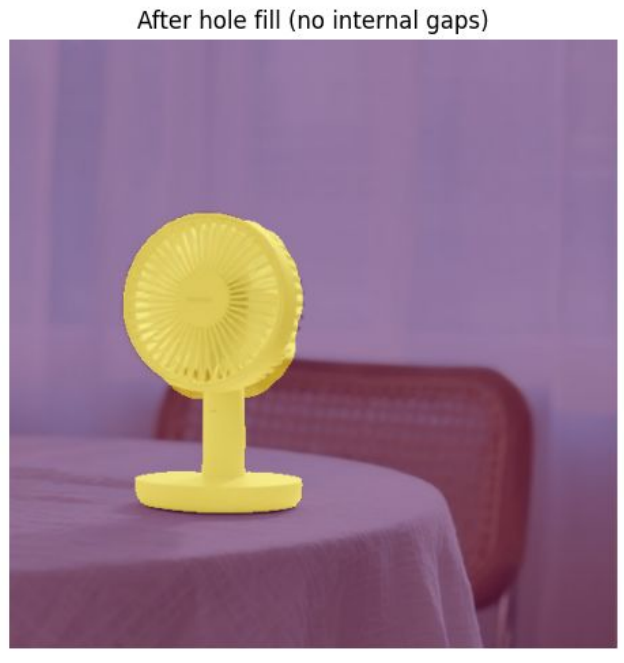

Filled + dilated (final hard mask)

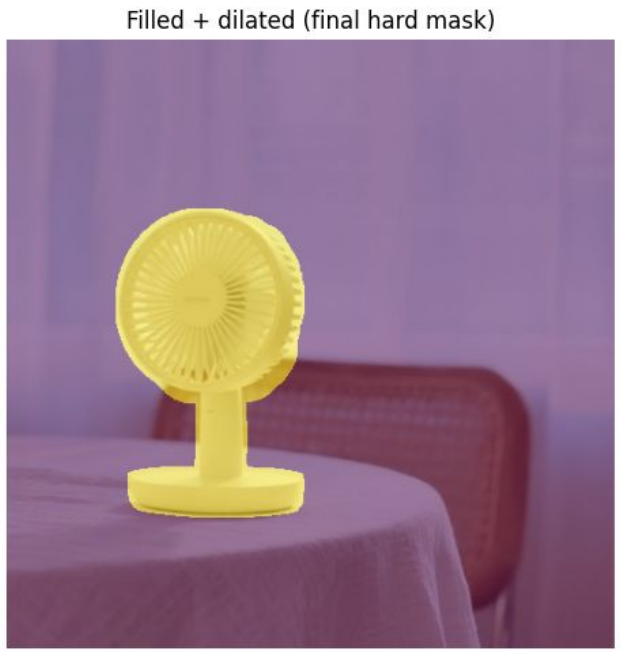

Pre-refine (NO NTI) | CFG_W=9.5

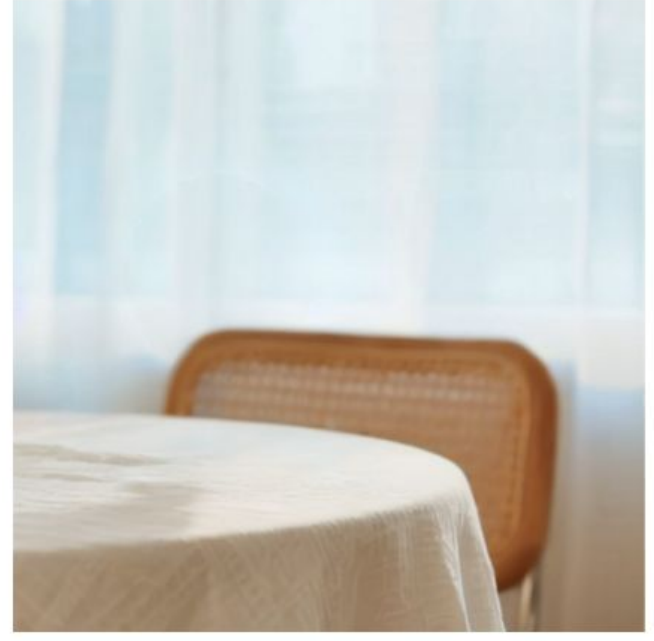

Refined (NO NTI) | R=10 | TREF_FRAC=0.55 | CFG_REF=9.5

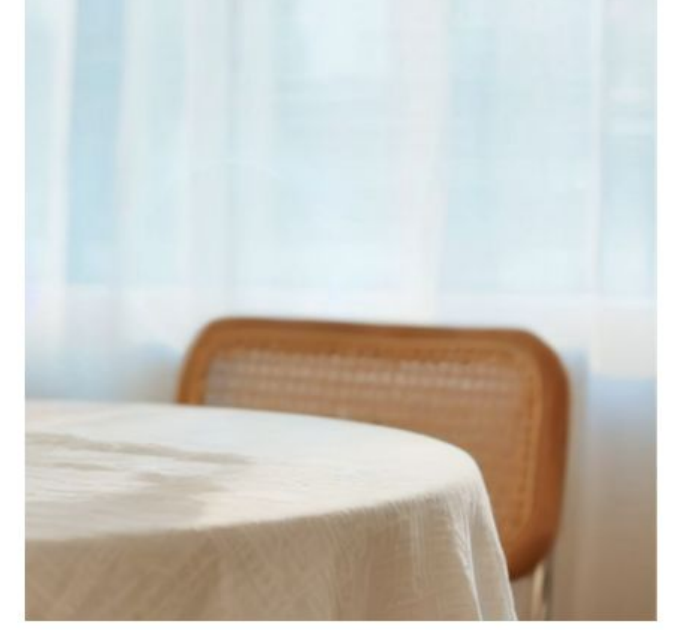

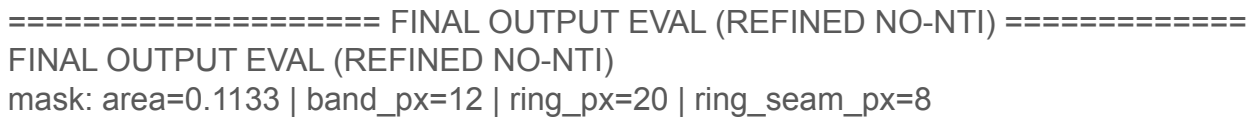

```
=================== FINAL OUTPUT EVAL (REFINED NO-NTI) ============
FINAL OUTPUT EVAL (REFINED NO-NTI)
mask: area=0.1133 | band_px=12 | ring_px=20 | ring_seam_px=8

Boundary seam (grad): in=0.0493 out=0.0462
abs_mismatch=0.0031 | rel_mismatch=0.0671  (lower better)
Boundary seam (color jump L2): 0.0191           (lower better)
Neighborhood consistency (outside rings): rgb_l2=0.0452 | grad_abs=0.0063 (lower better)
BG alignment (patch vs context): sim=0.6810 (CLIP) (higher better)
Local feature distance (patch vs context): d=0.9642 (ResNet50) (lower better)
SSIM_ring (orig vs ring_mix): 0.9723         (higher better)
LPIPS_ring (orig vs ring_mix): 0.0367        (lower better)

=================================================================
```

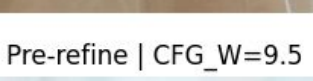
Pre-refine | CFG_W=9.5

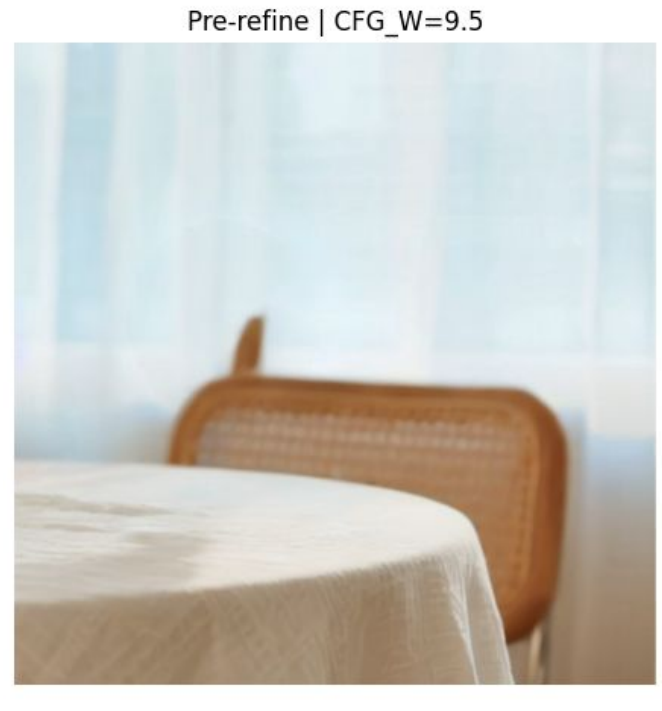

Refined | REFINE_ROUNDS=10 | TREF_FRAC=0.55 | CFG_REF=9.5

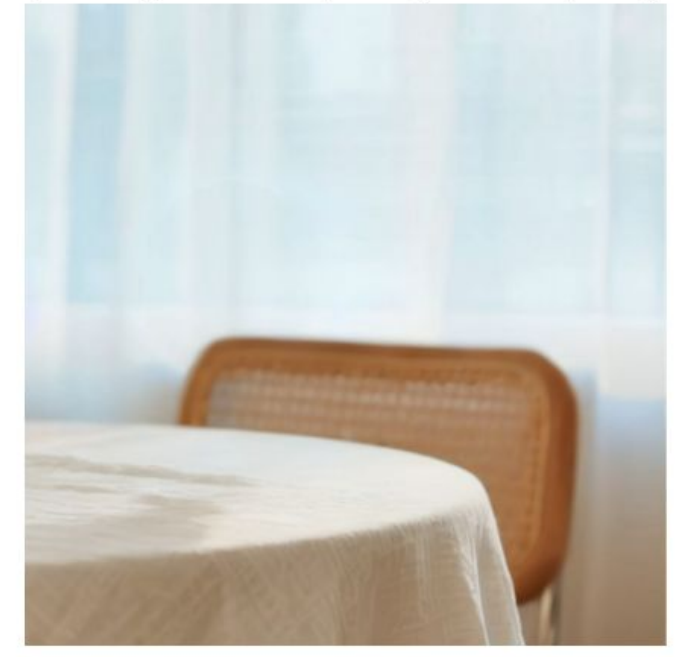

```
=================== FINAL OUTPUT EVAL (REFINED) ===================
FINAL OUTPUT EVAL (REFINED)
mask: area=0.1133 | band_px=12 | ring_px=20 | ring_seam_px=8

Boundary seam (grad): in=0.0496 out=0.0459
abs_mismatch=0.0036 | rel_mismatch=0.0792  (lower better)
Boundary seam (color jump L2): 0.0176           (lower better)
Neighborhood consistency (outside rings): rgb_l2=0.0458 | grad_abs=0.0064 (lower better)
BG alignment (patch vs context): sim=0.6834 (CLIP) (higher better)
Local feature distance (patch vs context): d=0.9634 (ResNet50) (lower better)
SSIM_ring (orig vs ring_mix): 0.9722         (higher better)
LPIPS_ring (orig vs ring_mix): 0.0367        (lower better)

=================================================================
```

The quantitative results are nearly identical across the two variants. The masked-NTI result achieves slightly lower boundary color jump, marginally higher CLIP-based background alignment, and a slightly lower local feature distance. In contrast, the no-NTI result obtains slightly lower absolute and relative boundary-gradient mismatch, better outside-ring consistency, and a marginally higher ring-focused SSIM, while ring-focused LPIPS is identical. These negligible differences support the qualitative observation that masked NTI provides no meaningful advantage in this simple case.

Target x0 (inv_xt[-1])

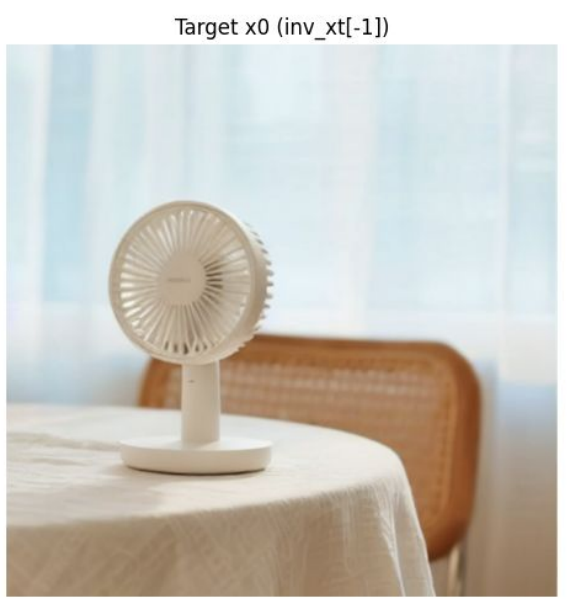

Recon FREE x0 (NO reinjection)

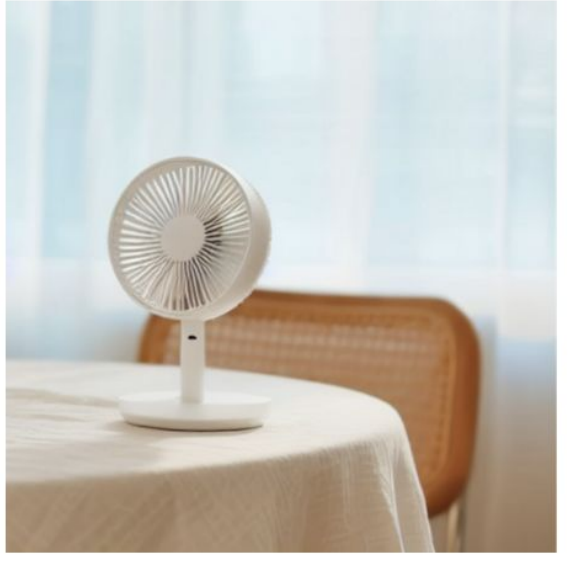

reconstruction with masked NTI

$\lambda_{in} \ll 1$

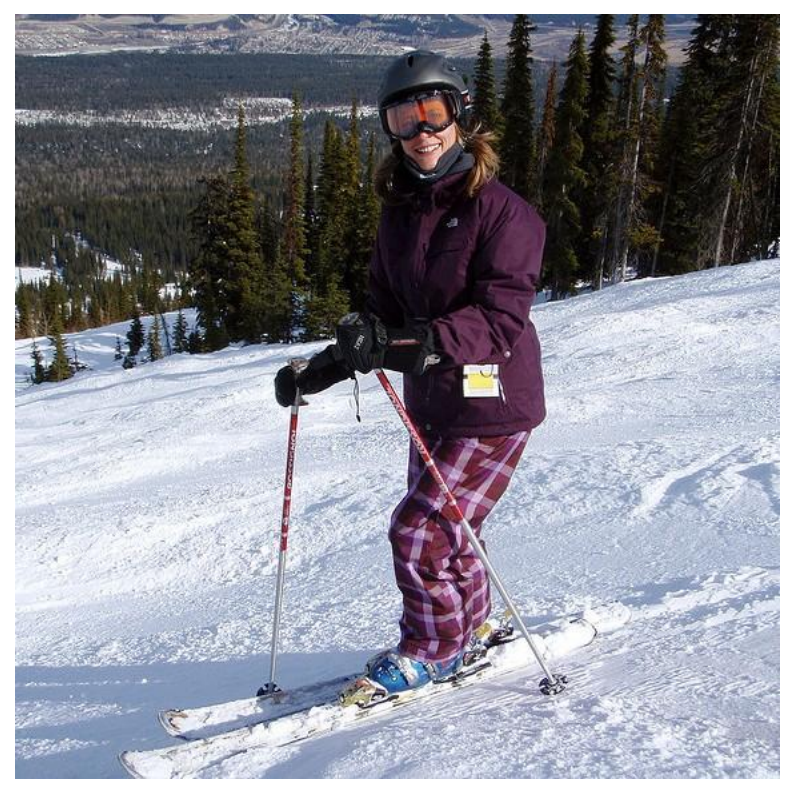

In this example, the objective is to remove the skiing woman and her ski poles from the image. In the no-NTI variant, decoder self-attention masking removes the main object, although some artifacts remain in the edited region. The refinement stage effectively removes these artifacts, producing a cleaner and more coherent output. However, the woman's shadow was not fully covered by the mask and was therefore restored by the hard background lock. This example highlights the importance of sufficient mask coverage and also shows a limitation of hard background locking, which preserves all unmasked content even when it belongs to the object being removed. Compared with the masked-NTI variant, only minor differences are observed and no substantial qualitative improvement is achieved. This suggests that decoder self-attention masking followed by refinement is sufficient in this case, while the additional masked-NTI stage provides no clear advantage.

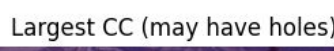

Largest CC (may have holes) | After hole fill (no internal gaps) | Filled + dilated (final hard mask)

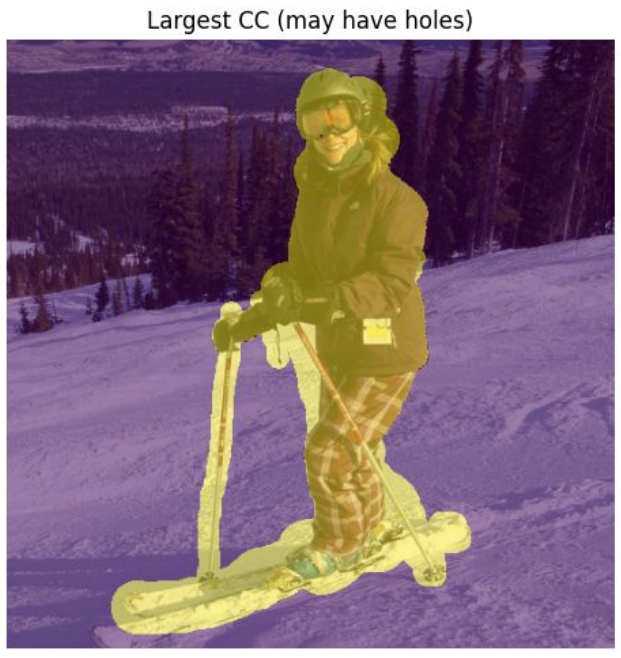

Pre-refine (NO NTI) | CFG_W=9.5

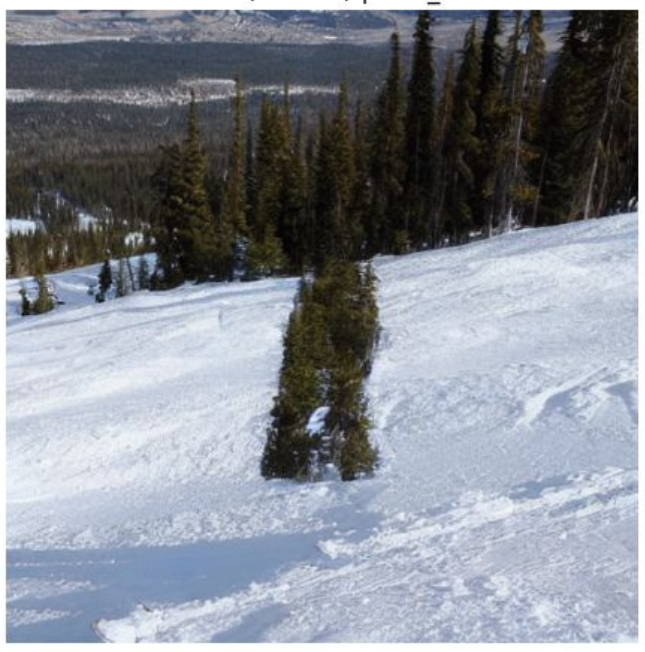

Refined (NO NTI) | R=10 | TREF_FRAC=0.55 | CFG_REF=9.5

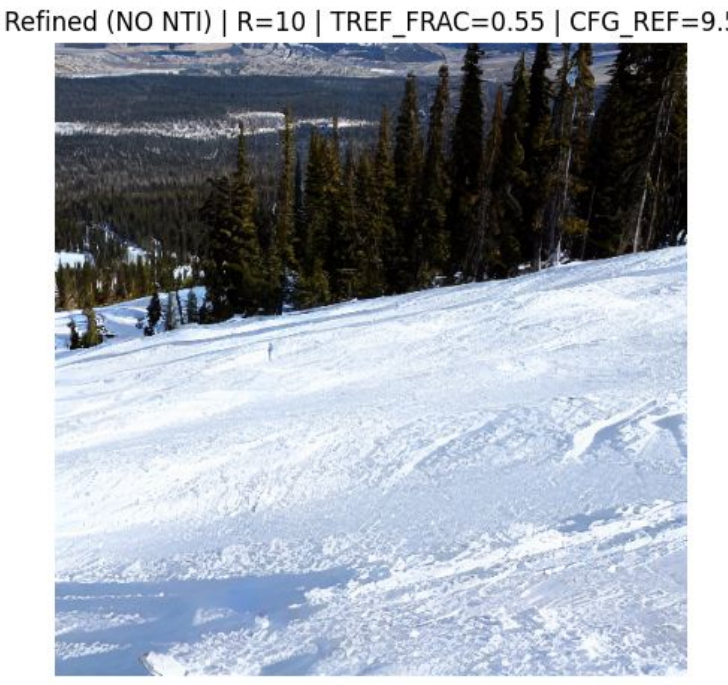

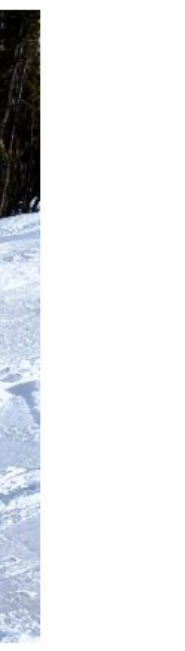

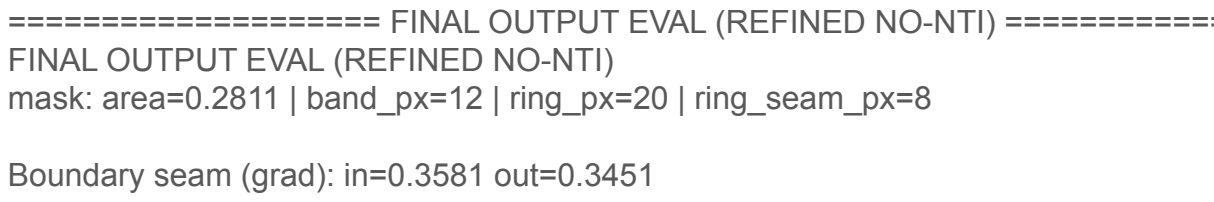

=================== FINAL OUTPUT EVAL (REFINED NO-NTI) ============
FINAL OUTPUT EVAL (REFINED NO-NTI)
mask: area=0.2811 | band_px=12 | ring_px=20 | ring_seam_px=8

Boundary seam (grad): in=0.3581 out=0.3451
abs_mismatch=0.0130 | rel_mismatch=0.0377 (lower better)
Boundary seam (color jump L2): 0.0323 (lower better)
Neighborhood consistency (outside rings): rgb_l2=0.0075 | grad_abs=0.0012 (lower better)
BG alignment (patch vs context): sim=0.7733 (CLIP) (higher better)
Local feature distance (patch vs context): d=0.7594 (ResNet50) (lower better)
SSIM_ring (orig vs ring_mix): 0.9176 (higher better)
LPIPS_ring (orig vs ring_mix): 0.0480 (lower better)

=================================================================

Pre-refine | CFG_W=9.5

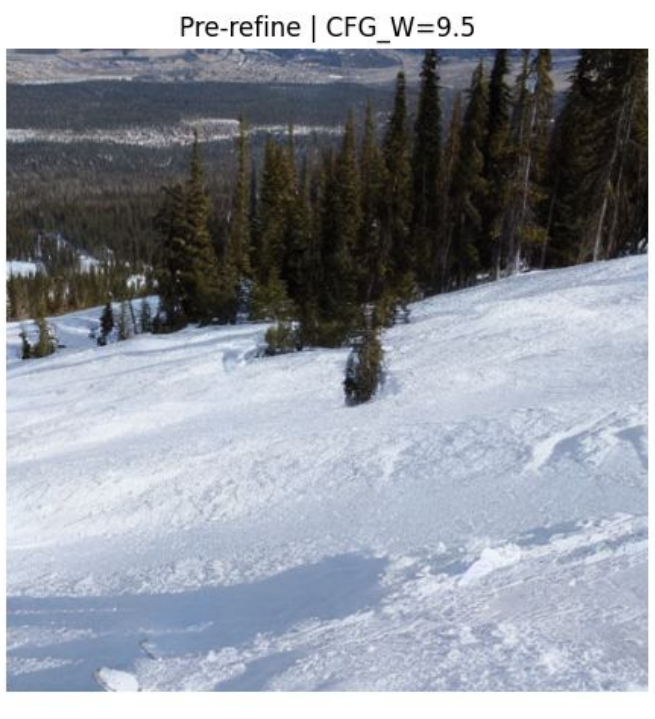

Refined | REFINE_ROUNDS=10 | TREF_FRAC=0.55 | CFG_REF=9.5

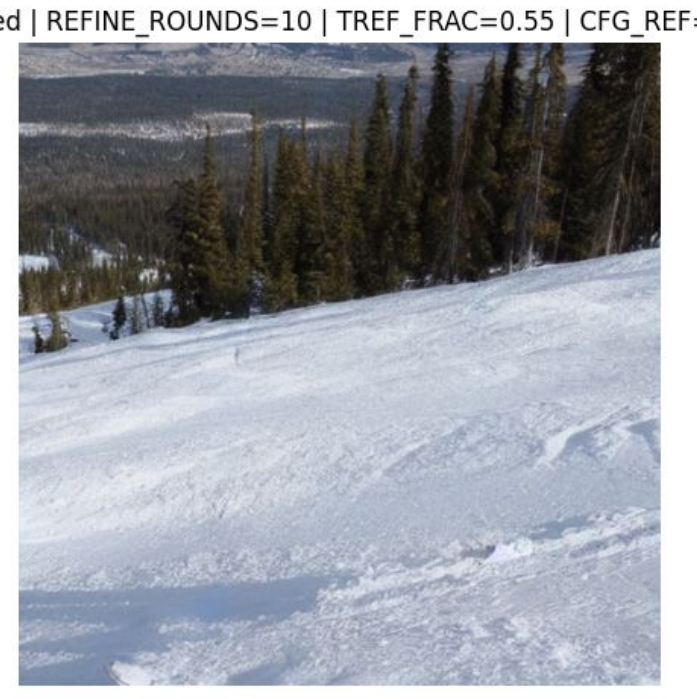

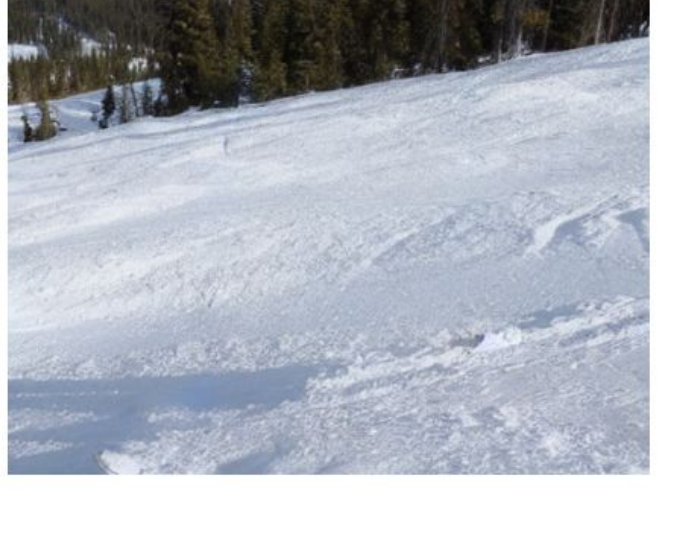

=================== FINAL OUTPUT EVAL (REFINED) ===================
FINAL OUTPUT EVAL (REFINED)
mask: area=0.2811 | band_px=12 | ring_px=20 | ring_seam_px=8

Boundary seam (grad): in=0.2412 out=0.2394
abs_mismatch=0.0018 | rel_mismatch=0.0076 (lower better)
Boundary seam (color jump L2): 0.0310 (lower better)
Neighborhood consistency (outside rings): rgb_l2=0.0064 | grad_abs=0.0042 (lower better)
BG alignment (patch vs context): sim=0.7609 (CLIP) (higher better)
Local feature distance (patch vs context): d=0.7954 (ResNet50) (lower better)
SSIM_ring (orig vs ring_mix): 0.9251 (higher better)
LPIPS_ring (orig vs ring_mix): 0.0393 (lower better)

=================================================================

The quantitative results are mixed. The masked-NTI result achieves substantially lower absolute and relative boundary-gradient mismatch, slightly lower boundary color jump and outside-ring RGB difference, higher ring-focused SSIM, and lower ring-focused LPIPS. In contrast, the no-NTI result obtains better outside-ring gradient consistency, higher CLIP-based background alignment, and a lower local feature distance. Although several boundary-focused metrics favor masked NTI, the qualitative difference remains limited. Moreover, these metrics do not capture the preserved shadow, since it lies outside the mask, further emphasizing the need to evaluate mask coverage and object-removal completeness through visual inspection.

Target x0 (inv_xt[-1])

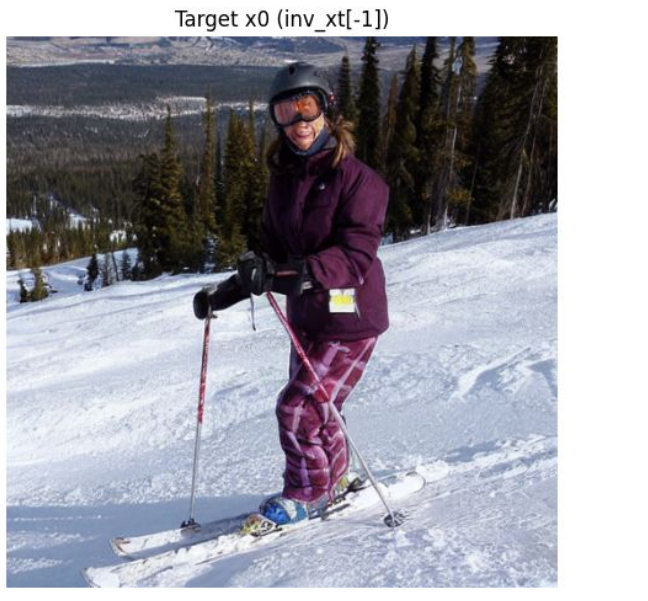

Recon FREE x0 (NO reinjection)

reconstruction with masked NTI

$\lambda_{in} \ll 1$

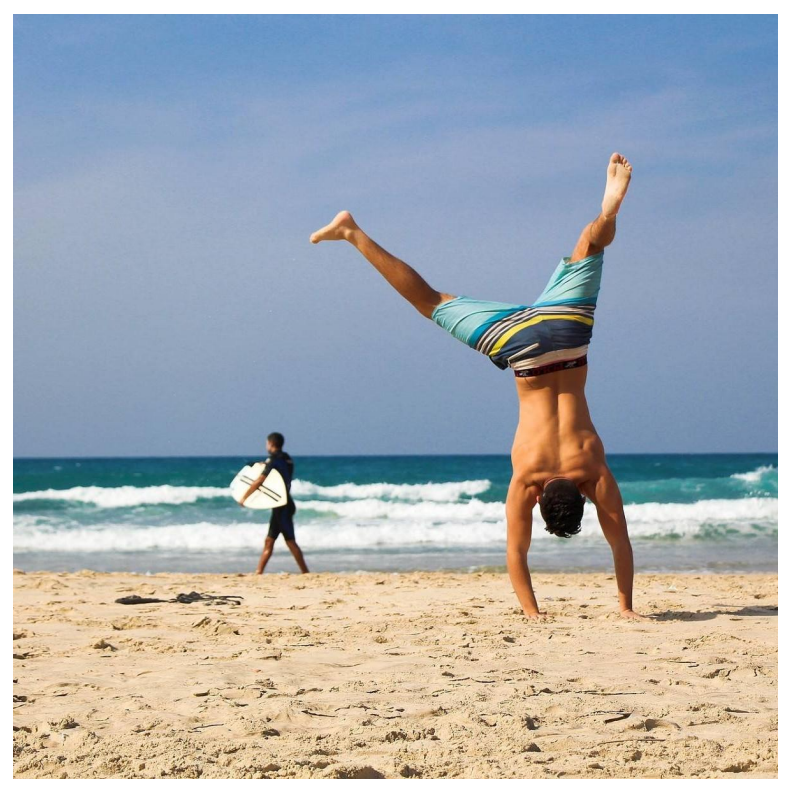

In this example, the objective is to remove the man from the beach. In the no-NTI variant, visible artifacts remain in the edited region, and refinement provides only a slight improvement, with significant inconsistencies still present. In contrast, the masked-NTI variant produces a noticeably cleaner intermediate result by removing these artifacts more effectively. Refinement further improves the visual quality, although some minor artifacts remain. These could potentially be reduced with a stronger refinement configuration, such as additional refinement rounds or refinement over a larger portion of the denoising trajectory.

Largest CC (may have holes)

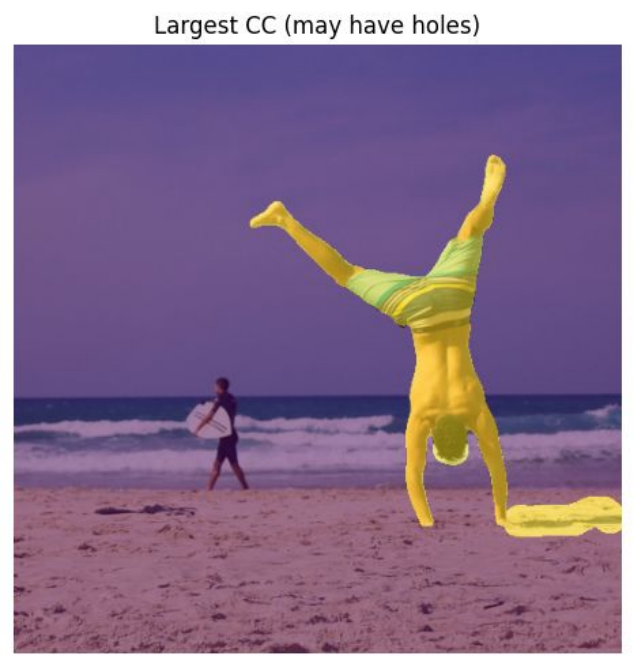

After hole fill (no internal gaps)

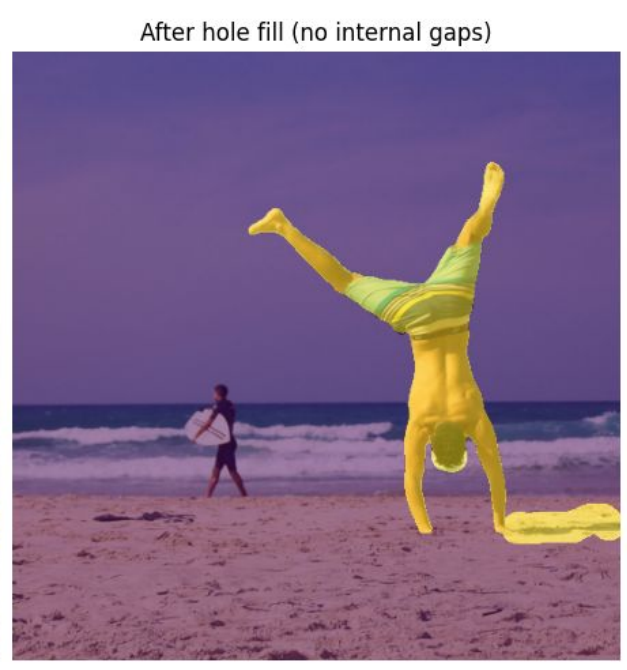

Filled + dilated (final hard mask)

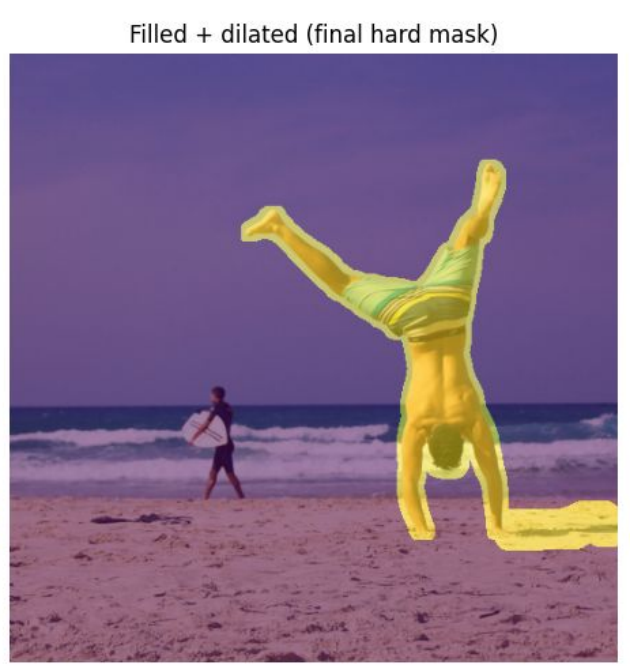

Pre-refine (NO NTI) | CFG_W=9.5

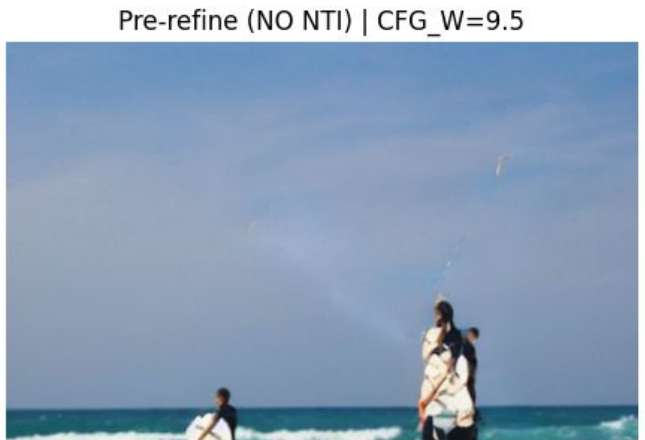

Refined (NO NTI) | R=20 | TREF_FRAC=0.3 | CFG_REF=10.5

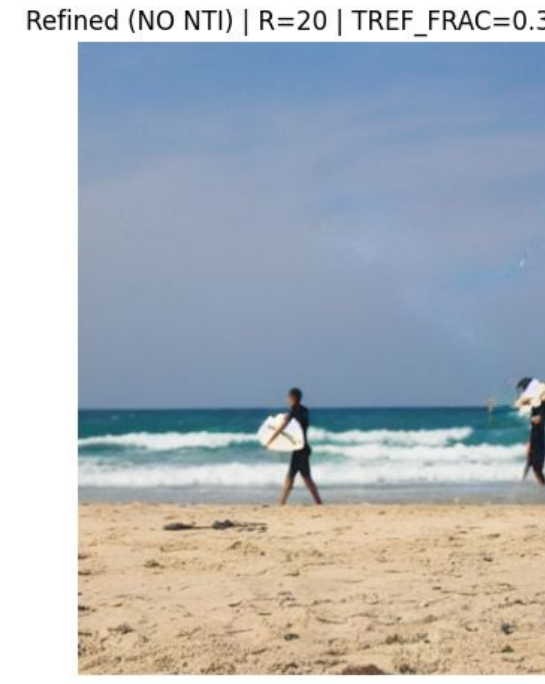

```
=================== FINAL OUTPUT EVAL (REFINED NO-NTI) ============
FINAL OUTPUT EVAL (REFINED NO-NTI)
mask: area=0.0971 | band_px=12 | ring_px=20 | ring_seam_px=8

Boundary seam (grad): in=0.1276 out=0.1017
abs_mismatch=0.0259 | rel_mismatch=0.2544  (lower better)
Boundary seam (color jump L2): 0.0147           (lower better)
Neighborhood consistency (outside rings): rgb_l2=0.0155 | grad_abs=0.0011 (lower better)
BG alignment (patch vs context): sim=0.8155 (CLIP) (higher better)
Local feature distance (patch vs context): d=1.0661 (ResNet50) (lower better)
SSIM_ring (orig vs ring_mix): 0.9489         (higher better)
LPIPS_ring (orig vs ring_mix): 0.0480        (lower better)

=================================================================
```

Pre-refine | CFG_W=9.5

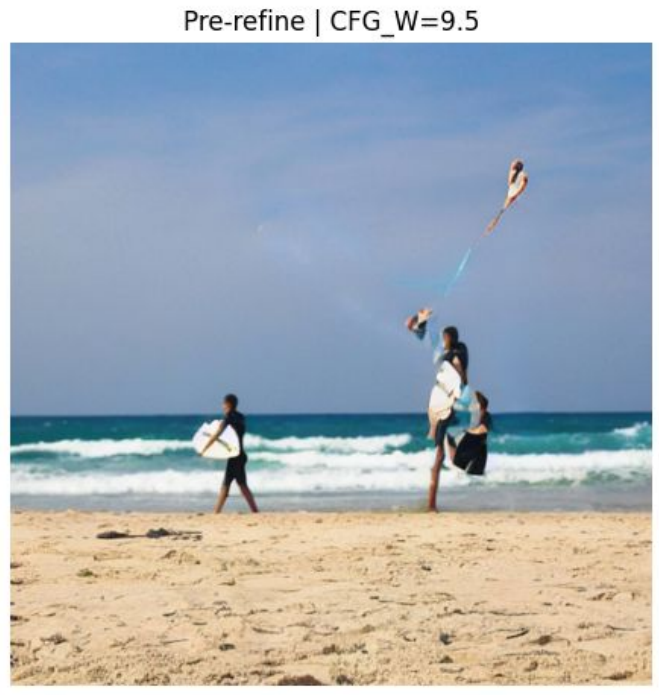

Refined | REFINE_ROUNDS=20 | TREF_FRAC=0.3 | CFG_REF=10.5

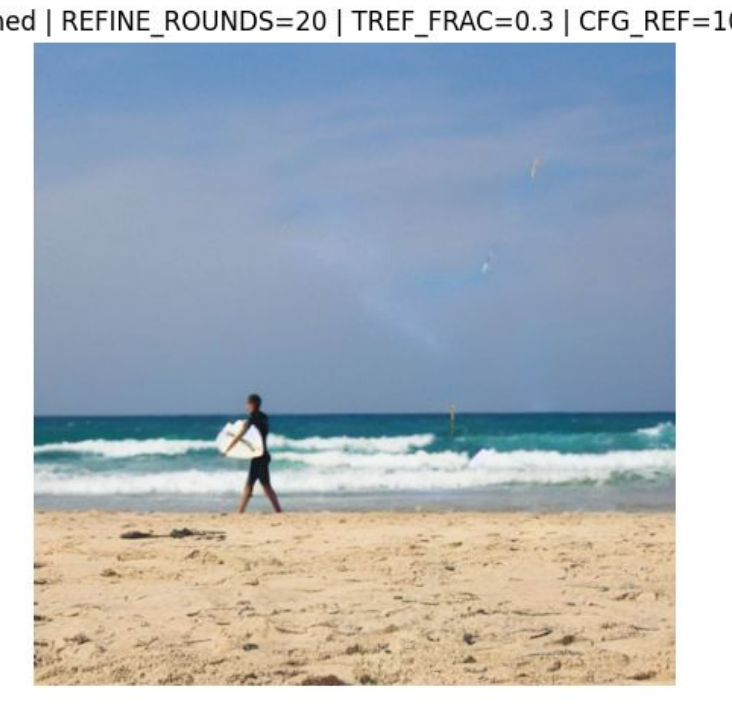

```
=================== FINAL OUTPUT EVAL (REFINED) ===================
FINAL OUTPUT EVAL (REFINED)
mask: area=0.0971 | band_px=12 | ring_px=20 | ring_seam_px=8

Boundary seam (grad): in=0.1173 out=0.1067
abs_mismatch=0.0106 | rel_mismatch=0.0996  (lower better)
Boundary seam (color jump L2): 0.0274           (lower better)
Neighborhood consistency (outside rings): rgb_l2=0.0162 | grad_abs=0.0004 (lower better)
BG alignment (patch vs context): sim=0.7978 (CLIP) (higher better)
Local feature distance (patch vs context): d=1.0705 (ResNet50) (lower better)
SSIM_ring (orig vs ring_mix): 0.9494         (higher better)
LPIPS_ring (orig vs ring_mix): 0.0481        (lower better)

==================================================================
```

The quantitative results are mixed but partly support the qualitative advantage of masked NTI. The masked-NTI result achieves substantially lower absolute and relative boundary-gradient mismatch, better outside-ring gradient consistency, and a marginally higher ring-focused SSIM. In contrast, the no-NTI variant obtains lower boundary color jump, slightly better outside-ring RGB consistency, higher CLIP-based background alignment, and marginally better local feature distance and ring-focused LPIPS. These results suggest that masked NTI mainly improves boundary-gradient consistency, while the remaining qualitative differences are not uniformly captured by the other metrics.

Target x0 (inv_xt[-1])

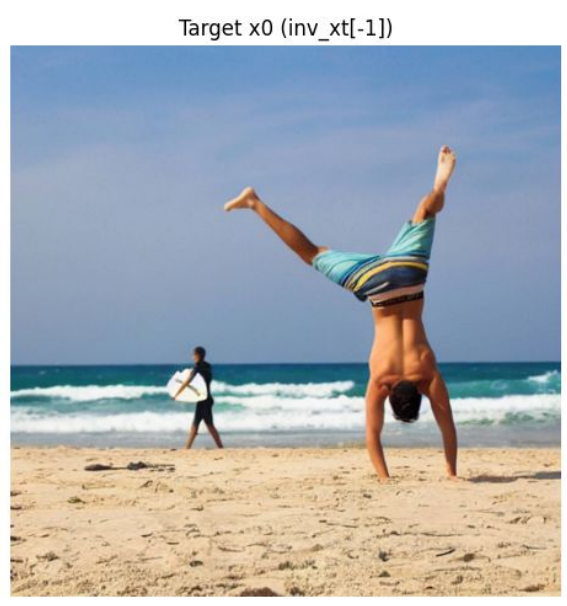

Recon FREE x0 (NO reinjection)

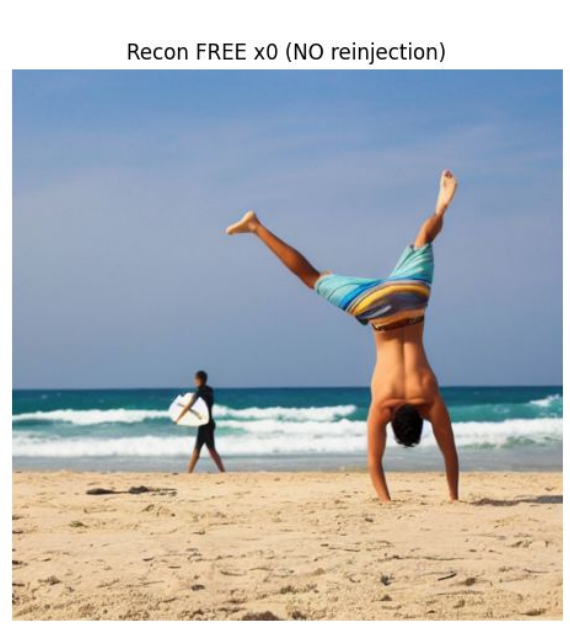

reconstruction with masked NTI

$\lambda_{in} \ll 1$